\documentclass[journal]{IEEEtran}
\usepackage{cite}
\usepackage{amsmath,amssymb,amsfonts}
\usepackage{algorithmic}
\usepackage{graphicx}
\usepackage{textcomp}

\usepackage{bm}
\usepackage[T1]{fontenc}
\usepackage{hyperref, times, cite}
\usepackage{url}
\usepackage{enumitem}
\usepackage{mathtools}
\usepackage{bm, bbm}
\usepackage{cool}
\usepackage{amssymb,amsfonts,amsmath,amsthm,amscd,dsfont,mathrsfs}
\usepackage{graphicx,float,psfrag,epsfig,amssymb}
\usepackage{wrapfig}
\usepackage{relsize}
\usepackage{color}
\usepackage{xcolor}
\usepackage{pict2e}
\usepackage{algorithm}
\usepackage{caption}
\usepackage{nameref}
\usepackage{enumerate}
\usepackage{stackrel}

\usepackage{caption}
\usepackage{subcaption}

\usepackage{amsthm}

\begin{document}
\thispagestyle{empty}
\title{Conditional Independence Graph Learning from Dependent Data Via Time-Delay Embedding}
\author{ Jitendra K.\ Tugnait    
\thanks{J.K.\ Tugnait is with the Department of 
Electrical \& Computer Engineering,
200 Broun Hall, Auburn University, Auburn, AL 36849, USA. 
Email: tugnajk@auburn.edu . }

\thanks{This work was supported by the National Science Foundation Grant CCF-2308473.}}

\maketitle

\renewcommand{\algorithmicrequire}{\textbf{Input:}}
\renewcommand{\algorithmicensure}{\textbf{Output:}}

\begin{abstract}
We consider the problem of inferring the conditional independence graph (CIG) of a high-dimensional stationary multivariate Gaussian time series which could be short-range dependent (SRD) or long-range dependent (LRD). In a time series graph, each component of the vector series is represented by a distinct node, and associations between components are represented by edges between the corresponding nodes. We consider a time-domain approach based on time-delay embedding, casting the problem as one of multi-attribute graph estimation for random vectors where a vector is associated with each node of the graph.  For an embedding dimension $d$, at each node the associated $(d+1)$-dimensional random vector consists of a time series component and its $d$ delayed copies, with integer delays up to $d \ge 1$.  Graph estimation from dependent data has been investigated in the past using frequency-domain methods whose analysis is restricted to SRD data. We consider a penalized negative pseudo log-likelihood function with convex lasso or non-convex regularizations to estimate the CIG. Non-convex regularization functions are log-sum, smoothly clipped absolute deviation (SCAD) and minimax concave penalty (MCP). We exploit an existing alternating direction method of multipliers (ADMM) approach to optimize the objective function. We establish  sufficient conditions in a high-dimensional setting for consistency, local convexity when using non-convex penalties, and graph recovery, for both SRD and LRD series. We illustrate our approach with both synthetic and real-data examples. For a $p$-dimensional time series, the computational complexity of the ADMM algorithm is ${\cal O}((d+1)^3 p^3)$. Based on the numerical examples, a heuristic is suggested and illustrated for selection of $d$ which, in theory, should tend to infinity to infer the time series graph.
\end{abstract}


\section{Introduction} \label{intro}
\IEEEPARstart{G}{raphical} models are used to display and explore conditional independence structure in the analysis of multivariate data  \cite{Whittaker1990, Lauritzen1996, Buhlmann2011, Wainwright2019}. In a time series Gaussian graphical model (TSGGM), the conditional statistical dependency structure among $p$ component series $\{ x_i(t) \}$, $i \in [p]$, of stationary, zero-mean,  $p-$dimensional Gaussian time series $\{{\bm x}(t), \, t \in \mathcal{Z} \}$, is represented using an undirected graph ${\cal G} = \left( V, {\cal E} \right)$ with a set of $p$ vertices (nodes) $V = \{1,2, \cdots , p\} =[p]$, and a corresponding set of (undirected) edges ${\cal E} \subseteq [p] \times [p]$. There is no edge between nodes $i$ and $j$ of the conditional independence graph (CIG) ${\cal G}$, (i.e., $\{ i,j \} \not\in {\cal E}$), if and only if (iff) $x_i(t)$ and $x_j(t)$ are conditionally independent given the remaining $p$-$2$ scalar series $x_\ell(t)$, $\ell \in V \textbackslash \{i, j \}$. This is a generalization of the CIG for random vectors where $\{ i,j \} \not\in {\cal E}$ iff $[{\bm \Omega}]_{ij} = 0$ where $[{\bm \Omega}]_{ij}$ denotes the $(i,j)$th component of ${\bm \Omega} = (E\{ {\bm x}(t) {\bm x}^\top(t) \})^{-1}$ \cite{Buhlmann2011, Dahlhaus2000, Wainwright2019}.

Graphical models were originally developed for random vectors \cite[p.\ 234]{Eichler2012} and are useful in a wide variety of applications \cite{Buhlmann2011, Eichler2012, Danaher2014}. Graphical modeling of stationary time series originated with \cite{Brillinger1996}, followed by \cite{Dahlhaus2000}. Denote the power spectral density (PSD) matrix of $\{ {\bm x}(t) \}$ by ${\bm S}_x(f)$, where ${\bm S}_x(f) = \sum_{\tau = -\infty}^{\infty}  {\bm R}_{x}( \tau ) e^{-\iota 2 \pi f \tau}$, ${\bm R}_{x}( \tau ) = {E} \{ {\bm x}(t + \tau) {\bm x}^T(t ) \}$ and $\iota = \sqrt{-1}$.  In \cite{Dahlhaus2000} it was shown that conditional independence of two time series components given all other components of the time series, is encoded by zeros in the inverse PSD, that is, $\{ i,j \} \not\in {\cal E}$ iff the $(i,j)$th element of ${\bm S}_x^{-1}(f)$,  $[{\bm S}_x^{-1}(f)]_{ij} = 0$ for every $f$; this requires ${\bm S}_x^{-1}(f)$ to exist for all $f$. Hence one can use estimated inverse PSD of observed time series to infer the associated graph. In \cite{Dahlhaus2000} the low-dimensional case is addressed. Nonparametric approaches for graphical modeling of real-valued time series in high-dimensional settings ($p$ is large and/or sample size $n$ is of the order of $p$) have been formulated in the form of penalized log-likelihood in frequency-domain in \cite{Jung2015a, Tugnait2022a, Tugnait2022d, Tugnait2025c}. 

In this paper we consider a time-domain approach to graph structure estimation for stationary Gaussian multivariate time series, using a time-delay embedding approach. The name delay embedding originates from \cite{Takens1981} in the context of chaos detection where such embeddings reveal the dynamic structure of the underlying system, and such embeddings have been employed in other contexts, e.g., \cite{Frank2010, Hirsch2021}. We cast the problem as one of multi-attribute graph estimation for random vectors where a vector is associated with each node of the graph. At each node, the associated random vector consists of a time series component and its time-delayed copies. In particular, given the $i$th component $x_i(t)$ of series $\{ {\bm x}(t) \}$, we associate vector series ${\bm z}_i(t) = [ x_i(t) \; x_i(t-1) \; \cdots \; x_i(t-d) ]^\top \in \mathbb{R}^{d+1}$ with node $i$ of the graph. Let ${\bm y}(t) = [ {\bm z}_1(t) \;{\bm z}_2(t) \; \cdots \; {\bm z}_p(t) ]^\top \in \mathbb{R}^{p(d+1)}$. We show that for ``large'' $d$, the precision matrix $(E\{ {\bm y}(t) {\bm y}^\top(t) \})^{-1}$ of the augmented sequence ${\bm y}(t)$ captures the CIG of time-dependent $\{ {\bm x}(t) \}$ (in the sense of \cite{Dahlhaus2000}). 

The delay embedding approach also allows us to analyze time series with long-range dependence, not analyzed in the frequency-domain approaches of \cite{Jung2015a, Tugnait2022a, Tugnait2025c}. A (zero-mean) univariate stationary time series $\{ y(t) \}$ is said to have long-range dependence (LRD), or long-memory, if its autocorrelation function $r_y(\tau) = E\{ y(t+\tau)y(t)\}$ satisfies $\sum_{\tau = - \infty}^\infty | r_y(\tau) | = \infty$, and short-range dependence (SRD), or short-memory, otherwise \cite{Hosking1981, Lobato1997, Taqqu2003, Shimotsu2007, GS2007, Kechagias2015, Pipiras}. A (zero-mean) multivariate stationary time series $\{ {\bm x}(t) \}$ ($p$-dimensional) is said to have LRD if its autocorrelation function ${\bm R}_x(\tau) = E\{ {\bm x}(t+\tau){\bm x}^\top(t)\}$ has the structure $[{\bm R}_x(\tau)]_{jk} = [\tilde{\bm R}(\tau)]_{jk} \tau^{d_j+d_k-1}$ where $[{\bm A}]_{jk}$ denotes $(j,k)$th component of ${\bm A}$, $d_j \in (0,0.5)$, $j=1,2, \cdots , p$, and $\tilde{\bm R}(\tau) \in \mathbb{R}^{p \times p}$ satisfies $\lim_{\tau \rightarrow \infty} \tilde{\bm R}(\tau) = \bar{\bm R}$ \cite{Kechagias2015}. Otherwise, $\{ {\bm x}(t) \}$ is said to have SRD. Note that in the multivariate case, there is no universal characterization of LRD properties \cite{Kechagias2015, Pipiras}.

\subsection{Related Work} \label{RW}
There appears to be no prior reported work on any nonparametric time-domain approach to TSGGM estimation in high-dimensional settings from dependent data. Past work on TSGGM estimation from dependent data has focused on frequency-domain approaches. Prior work on graphical modeling for single-attribute dependent time series in low-dimensional settings includes \cite{Dahlhaus2000, Matsuda2006, Wolstenholme2015, Luftman2016, Tugnait19d} where some nonparametric frequency-domain test statistics are used to test whether $\{ i,j \} \in {\cal E}$. These approaches do not scale to high dimensions where $p$ is comparable to or larger than the sample size $n$. 

Nonparametric approaches in high-dimensional settings have been formulated in the form of lasso-penalized log-likelihood in frequency-domain in \cite{Jung2015a, Tugnait2022a}. A sparse-group non-convex log-sum penalty is investigated in \cite{Tugnait2022d, Tugnait2025c} to regularize the problem considered in \cite{Tugnait2022a}, motivated by \cite{Candes2008}. The analysis of the approaches of \cite{Jung2015a, Tugnait2022a, Tugnait2022d, Tugnait2025c} is restricted to SRD processes. As an alternative to nonparametric modeling of time series, parametric graphical models utilizing  (Gaussian) vector autoregressive (AR) process models of ${\bm x}(t)$ have been proposed in \cite{Eichler2006, Eichler2012, Songsiri2009, Songsiri2010} and autoregressive moving average (ARMA) process (and related) models may be found in \cite{Avventi2013, Alpago2018, You2022}, but these approaches are suitable only for low-dimensional settings.

There has been extensive prior work on both single attribute and multi-attribute graph estimation for random vectors based on estimation of the underlying precision matrix, and their analysis is based on the assumption of i.i.d.\ data \cite{Buhlmann2011, Wainwright2019, Danaher2014, Lam2009, Zhang2010, Kolar2014, Ravikumar2011, Rothman2008, Friedman2008, Cai2011, Zhao2012, Tugnait21a, Tugnait2025b, Friedman2010b, Loh2017}. Analysis of estimation of the precision matrix and the corresponding graph under dependent data has been investigated in \cite{Shu19} using tail bounds derived using Hanson-Wright inequality \cite{Rudelson13}. Ref.\ \cite{Shu19} considers both SRD and LRD stationary processes. 

In this paper we consider estimation of TSGGM for both SRD and LRD data whereas \cite{Shu19} considers the more restrictive precision matrix based graph estimation. That is, given dependent data $\{ {\bm x}(t) \}$, in \cite{Shu19}, the graph estimation is based on ${\bm \Omega} = (E\{ {\bm x}(t) {\bm x}^\top(t) \})^{-1}$, not the inverse PSD ${\bm S}_x^{-1}(f)$ or an equivalent statistic, but analysis of the precision matrix estimator $\hat{\bm \Omega}$ accounts for temporal dependence of data. In this paper we account for temporal dependence of data both in graph estimation and in analysis. For some tail bounds in this paper we follow the time series dependence model of \cite{Chen16} which is formulated in terms of an infinite-order causal moving average model, unlike that of \cite{Shu19} which is stated in terms of the data covariance function.

In several applications such as functional magnetic resonance imaging (fMRI) signal analysis or financial time series analysis, the underlying temporal data is not i.i.d. In \cite[Sec.\ 5.2]{Songsiri2010} AR models are fitted to a financial time series (international stock market data) to infer the underlying time series graphical model. Analysis of some resting state fMRI data in \cite{Shu19} shows significant time-dependence. The data set analyzed in \cite[Sec.\ 2.2]{Shu19} consists of a single subject 1190 temporal brain images, each image with 907 functional brain nodes. This data passed stationarity, Gaussianity and linearity tests \cite[Sec.\ 2.2]{Shu19}, implying that stationary Gaussian time series assumption used in this paper would be appropriate. Furthermore, analysis reported in \cite[Sec.\ 2.2]{Shu19} determined that some fMRI series components are LRD, where, for large $\tau$ and some $C >0$, $|E\{ x_i(t+\tau)x_i(t)\}| = C |\tau|^{-0.3} $  (which corresponds to exponents $\gamma =0.65$ and $d_i=0.35$ in Secs.\ \ref{TSM} and \ref{ARF}, with $\gamma \in (0.5,1]$ and $d_i \in (0,0.5)$ implying long-range dependence). Therefore, consideration of long-range dependence in this paper is well-motivated.

Multi-attribute graph learning problem has been addressed in \cite{Kolar2014, Tugnait21a} with lasso regularization and in \cite{Tugnait2025b} with non-convex log-sum and SCAD regularizations, except that \cite{Kolar2014, Tugnait21a, Tugnait2025b} address estimation from i.i.d.\ data whereas in this paper we have dependent data. We use the log-likelihood of \cite{Kolar2014, Tugnait21a, Tugnait2025b} as our pseudo log-likelihood for dependent data. This allows us to optimize it as in \cite{Tugnait2025b} but analysis of its properties is different since now we have dependent data which could be SRD or LRD.

\subsection{Our Contributions}  \label{OC}
In this paper we propose a nonparametric time-domain approach to TSGGM learning from a high-dimensional stationary multivariate Gaussian time series which could be short-range dependent (SRD) or long-range dependent (LRD). We consider a time-domain approach based on time-delay embedding, casting the problem as one of multi-attribute graph estimation for random vectors where a vector is associated with each node of the graph. We consider a penalized negative pseudo log-likelihood function with convex lasso or non-convex regularizations to estimate the CIG. Non-convex regularization functions are log-sum \cite{Candes2008}, smoothly clipped absolute deviation (SCAD) \cite{Fan2001, Lam2009}, and minimax concave penalty (MCP) \cite{Zhang2010}. It is well-known that use of non-convex penalties can yield more accurate results compared to the convex lasso penalty. We exploit an existing alternating direction method of multipliers (ADMM) approach  from \cite{Tugnait2025b} to optimize the objective function. We establish  sufficient conditions in a high-dimensional setting for consistency, local convexity when using non-convex penalties, and graph recovery, for both SRD and LRD series. We illustrate our approach with both synthetic and real-data examples where comparisons with some existing approaches are provided.

Preliminary versions of this paper appear in the conference papers \cite{Tugnait2022b, Tugnait2022c}. The main idea of multi-attribute formulation appears in \cite{Tugnait2022b} with lasso penalty, and a consistency result is stated therein for SRD processes with no proof. In \cite{Tugnait2022c} a proof of consistency is given for LRD processes for lasso-penalized loss function and there are no simulation or real data results. In \cite{Tugnait2022b, Tugnait2022c} non-convex penalties are not addressed. Furthermore, in \cite{Tugnait2022b, Tugnait2022c} the sufficient conditions for consistency  for lasso penalty are stated in terms of the covariance of the multi-attribute process $\{ {\bm y}(t)\}$ whereas in this paper we also formulate them in terms of the original process $\{ {\bm x}(t) \}$; see Remarks 4 and 5 in Sec.\ \ref{TA}. There are no real data results in \cite{Tugnait2022b, Tugnait2022c}, and the synthetic data results in this paper are more comprehensive compared to \cite{Tugnait2022b}.

\subsection{Notation} \label{notat}
For a set $V$, $|V|$ denotes its cardinality. Given a real $a$, $(a)_+ = \max(0,a)$. Given ${\bm A} \in \mathbb{C}^{p \times p}$, we use $\phi_{\min }({\bm A})$, $\phi_{\max }({\bm A})$, $|{\bm A}|$ and $\mbox{tr}({\bm A})$ to denote the minimum eigenvalue, maximum eigenvalue, determinant and  trace of ${\bm A}$, respectively, and ${\bm A} \succ {\bm 0}$ and ${\bm A} \succeq {\bm 0}$ denote that ${\bm A}$ is positive-definite and positive semi-definite, respectively. Given ${\bm B} \in \mathbb{C}^{p \times m}$, $[{\bm B}]_{ij}$ denotes the $(i,j)$-th element of ${\bm B}$, and so does $B_{ij}$, and  ${\bm I}_q$ denotes the $q \times q$ identity matrix. The symbol $\otimes$ denotes the matrix Kronecker product. The superscripts $\ast$ and $H$ denote the complex conjugate and conjugate transpose operations, respectively. 

For ${\bm B} \in \mathbb{C}^{p \times q}$, we define the operator norm, the Frobenius norm, the vectorized $\ell_1$ norm and element-wise maximum, respectively, as $\|{\bm B}\| = \sqrt{\phi_{\max }({\bm B}^H  {\bm B})}$, $\|{\bm B}\|_F = \sqrt{\mbox{tr}({\bm B}^H  {\bm B})}$, $\|{\bm B}\|_1 = \sum_{i,j} |B_{ij}|$ and $\|{\bm B}\|_\infty = \max_{i,j} |B_{ij}|$. Given ${\bm A} \in \mathbb{C}^{n \times p}$, column vector $\mbox{vec}({\bm A}) \in \mathbb{C}^{np}$ denotes the vectorization of ${\bm A}$ which stacks the columns of the matrix ${\bm A}$. 

Given a variable vector ${\bm x}$ or matrix ${\bm X}$, we use ${\bm x}^\diamond$ or ${\bm X}^\diamond$, respectively, to denote their true values. Given a diagonal matrix ${\bm D} = \mbox{diag}\{d_1, d_2, \cdots , d_p \} \in \mathbb{C}^p$ and a scalar $a$, $a^D$ denotes $\mbox{diag}\{a^{d_1}, a^{d_2}, \cdots , a^{d_p} \}$, similar to \cite{Kechagias2015}. For ${\bm y}_n \in \mathbb{R}^p$ and ${\bm x}_n \in \mathbb{R}^q$, ${\bm x}_n = {\cal O}({\bm y}_n)$ means there exists $0 < M < \infty$ such that $\|{\bm x}_n\| \le M \|{\bm y}_n\|$ $\forall n \ge 1$, and ${\bm x}_n = o({\bm y}_n)$ means $\lim_{n \rightarrow \infty} \|{\bm x}_n\| /\|{\bm y}_n\| = 0$. The notation ${\bm y}_n = {\cal O}_P({\bm x}_n)$ for random vectors ${\bm y}_n \in \mathbb{R}^p$ and ${\bm x}_n \in \mathbb{R}^q$ means that for any $\varepsilon > 0$, there exists $0 < M < \infty$ such that $P ( \|{\bm y}_n\| \le M \|{\bm x}_n\|) \ge 1 - \varepsilon$ $\forall n \ge 1$.

\subsection{Outline} \label{outline}
The rest of the paper is organized as follows.  Some background material for graphical models for random vectors and for multivariate time series is presented in Sec.\ \ref{GM}. Time-delay embedding and the resulting multi-attribute graph formulation is discussed in Sec.\ \ref{TDE}. The resulting log-likelihood formulation of the problem is presented in Sec.\ \ref{PNLL} and its optimization via ADMM is presented in Sec.\ \ref{OPT}. A computational complexity analysis of the ADMM algorithm is in Sec.\ \ref{CC}, and model selection aspect for choosing the regularization parameter in the penalized pseudo log-likelihood objective function is discussed in Sec.\ \ref{modsel}.  In Sec.\ \ref{TA} we provide a theoretical analysis of the proposed approach, resulting in Theorems 1-3. Synthetic data numerical results are presented in Sec.\ \ref{NE} and real data results are in Sec.\ \ref{NEreal}. The proofs of the technical results in Sec.\ \ref{TA} are given in two appendices.

\section{Graphical Models} \label{GM}
Here we provide some background material for graphical models for random vectors and for multivariate time series which then forms the basis for the proposed time-delay embedding discussed in Sec.\ \ref{TDE}.

\subsection{Random Vectors} \label{MRV}
Consider a graph ${\cal G} = \left( V, {\cal E} \right)$ with $V = [p]$ and ${\cal E} \subseteq V \times V$. Let ${\bm x} = [x_1 \; x_2 \; \cdots \; x_p]^\top \in \mathbb{R}^p$ denote a Gaussian random vector that is zero-mean with covariance ${\bm \Sigma} = E\{ {\bm x} {\bm x}^\top \} \succ {\bm 0}$. The conditional independence relationships among $x_i$'s are encoded in ${\cal E}$ where edge $\{ i,j \} \not\in {\cal E}$ iff $x_i$ and $x_j$ are conditionally independent given $x_\ell$, $\ell \in V \textbackslash \{i, j \}$. Let 
\begin{equation} \label{neweq1}
   {\bm x}_{-ij} = \{ x_k \, : \, k \in V \textbackslash \{i, j \} \}  \in \mathbb{R}^{p-2}
\end{equation}
denote the vector ${\bm x}$ after deleting $x_i$ and $x_j$ from it. Let $\bm{\Omega} = \bm{\Sigma}^{-1}$ denote the precision matrix.  Define
\begin{equation}  \label{neweq2}
   e_{i|-ij} = x_i - E \{ x_i | {\bm x}_{-ij} \} \, , \;\; 
	 e_{j|-ij} = x_j - E \{ x_j | {\bm x}_{-ij} \} \, .
\end{equation}
Then we have the following equivalence \cite{Lauritzen1996}
\begin{equation}  \label{neweq3}
  \{ i,j \} \not\in {\cal E} \; \Leftrightarrow \; \Omega_{ij} = 0 
  \;  \Leftrightarrow \; E\{ e_{i|-ij} e_{j|-ij} \} = 0 \, .
\end{equation}

\subsection{Multivariate Time Series Graphical Models} \label{MTS} 
Consider zero-mean stationary Gaussian time series ${\bm x}(t) \in \mathbb{R}^p$, $t \in \mathbb{Z}$, with PSD ${\bm S}_{x}(f)$. The conditional independence relationships among time series components $\{ x_i(t) \}$'s are encoded in edge set ${\cal E}$ of ${\cal G} = \left( V, {\cal E} \right)$ where edge $\{ i,j \} \not\in {\cal E}$ iff $\{ x_i(t), \, t \in \mathbb{Z} \}$ and $\{ x_j(t), \, t \in \mathbb{Z} \}$ are conditionally independent given the remaining $p$-$2$ components 
\begin{equation} \label{neweq10}
   {\bm x}_{-ij,\mathbb{Z}} = \{ x_k(t) \, : \, k \in V \textbackslash \{i, j \}, \, 
	    t \in \mathbb{Z}  \}   \, .
\end{equation}
Define
\begin{align}  
   e_{i|-ij}(t) = & x_i(t) - E \{ x_i(t) | {\bm x}_{-ij,\mathbb{Z}} \}  \label{neweq11} \\
	 e_{j|-ij}(t) = & x_j(t) - E \{ x_j(t) | {\bm x}_{-ij,\mathbb{Z}} \} \, . \label{neweq12}
\end{align}
Then we have the following equivalence \cite{Dahlhaus2000}
\begin{align}  
  \{ i,j \} \not\in {\cal E} & \; \Leftrightarrow \; [{\bm S}_x^{-1}(f)]_{ij} = 0 \; 
	              \forall f \in [0,1]  \nonumber \\
  & \;  \Leftrightarrow \; E\{ e_{i|-ij}(t+\tau) e_{j|-ij}(t) \} = 0 \; \forall \tau \in \mathbb{Z} \, .
	       \label{neweq14}
\end{align}

\subsection{Multi-Attribute Graphical Models for Random Vectors} \label{MAMRV} 
Now consider $p$ jointly Gaussian vectors ${\bm z}_i \in \mathbb{R}^m$, $i \in [p]$. We associate ${\bm z}_i$ with the $i$th node of graph ${\cal G} = \left( V, {\cal E} \right)$,  $V = [p]$, ${\cal E} \subseteq V \times V$. We now have $m$ attributes per node. Now $\{ i,j \} \not\in {\cal E}$ iff vectors ${\bm z}_i$ and ${\bm z}_j$ are conditionally independent given the remaining $p$-$2$ vectors $\{ {\bm z}_\ell \, , \ell \in V \textbackslash \{i, j \} \}$. \begin{equation}  \label{neweq20}
   {\bm x} = [ {\bm z}_1^\top \; {\bm z}_2^\top \; \cdots \; {\bm z}_p^\top ]^\top 
	              \in \mathbb{R}^{mp} \, .
\end{equation}
Let ${\bm \Omega} = (E \{ {\bm x} {\bm x}^\top \} )^{-1}$ assuming $E \{ {\bm x} {\bm x}^\top \} \succ {\bm 0}$. Define the $m \times m$ subblock ${\bm \Omega}^{(ij)}$ of ${\bm \Omega}$ as
\begin{equation}  \label{neweq21}
   [{\bm \Omega}^{(ij)}]_{rs} = [{\bm \Omega}]_{(i-1)m+r, (j-1)m+s} \, , \; r,s \in [m] \, .
\end{equation}
Let
\begin{equation} \label{neweq22}
   {\bm z}_{-ij} = \{ {\bm z}_k \, : \, k \in V \textbackslash \{i, j \} \}  \in \mathbb{R}^{m(p-2)}
\end{equation}
denote the vector ${\bm x}$ in (\ref{neweq20}) after deleting vectors ${\bm z}_i$ and ${\bm z}_j$ from it. Define
\begin{equation}  \label{neweq23}
   {\bm e}_{i|-ij} = {\bm z}_i - E \{ {\bm z}_i | {\bm z}_{-ij} \} \, , \;\; 
	 {\bm e}_{j|-ij} = {\bm z}_j - E \{ {\bm z}_j | {\bm z}_{-ij} \} \, .
\end{equation}
Then we have the following equivalence \cite{Kolar2014}
\begin{equation}  \label{neweq24}
  \{ i,j \} \not\in {\cal E} \; \Leftrightarrow \; {\bm \Omega}^{(ij)} = {\bm 0}
  \;  \Leftrightarrow \; E\{ {\bm e}_{i|-ij} {\bm e}_{j|-ij}^\top \} = {\bm 0} \, ,
\end{equation}
where the first equivalence in (\ref{neweq24}) is given in \cite[Sec.\ 2.1]{Kolar2014} and the second equivalence is given in \cite[Appendix B.3]{Kolar2014}.

\section{Time-Delay Embedding}  \label{TDE}
Consider stationary, zero-mean,  $p-$dimensional multivariate Gaussian time series ${\bm x}(t)$ with PSD ${\bm S}_x(f) \succ {\bm 0}$. We wish to estimate the associated CIG ${\cal G} = \left( V, {\cal E} \right)$ as discussed in Sec.\ \ref{MTS}, but using a time-domain formulation which is explained in this section. This approach results in a multi-attribute graph estimation problem discussed in \cite{Tugnait2025b}, except that \cite{Tugnait2025b} deals with i.i.d.\ data whereas here we have dependent data. In Sec.\ \ref{TSM} we quantify the dependence structure of $\{ {\bm x}(t) \}$.

\subsection{Multi-Attribute Formulation for Time Series Graphical Modeling: Time-Delay Embedding} \label{MAMTS}
With $\{\bm x(t) \}$ as in Sec.\ \ref{MTS}, let ($d \ge 1$)
\begin{align}  
   {\bm z}_i(t) = & [ x_i(t) \; x_i(t-1) \; \cdots \; x_i(t-d) ]^\top \in \mathbb{R}^{d+1}
	               \label{neweq30} \\
	{\bm y}(t) = & [ {\bm z}_1^\top (t) \; {\bm z}_2^\top (t) \; \cdots \; 
	         {\bm z}_p^\top(t) ]^\top  \in \mathbb{R}^{(d+1)p}
	               \, .\label{neweq31}
\end{align}
Let ${\bm \Omega}_y = (E\{ {\bm y}(t) {\bm y}^\top(t) \})^{-1}$. With $m=d+1$, define the $m \times m$ subblock ${\bm \Omega}_y^{(ij)}$ of ${\bm \Omega}_y$ as
\begin{equation}  \label{neweq32}
   [{\bm \Omega}_y^{(ij)}]_{rs} = [{\bm \Omega}_y]_{(i-1)m+r, (j-1)m+s} \, , \; r,s \in [m] \, .
\end{equation}

{\it Lemma 1}. For a stationary zero-mean time series $\{\bm x(t) \}$ as in Sec.\ \ref{MTS}, $\{ x_i(u), \; t-d \le u \le t \}$ and $\{ x_j(u), \; t-d \le u \le t \}$, $i \ne j$,  are conditionally independent given 
$\{ x_{\ell}(u), \; t-d \le u \le t, \; \ell \in V \textbackslash \{i, j \} \}$ iff ${\bm \Omega}_y^{(ij)} = {\bm 0}$.
 $\quad \bullet$ \\
{\it Proof}. 
Let
\begin{align}  
   {\bm z}_{-ij}(t) = & \{ {\bm z}_k(t) \, : \, k \in V \textbackslash \{i, j \} \} \, , \label{neweq33} \\
   {\bm e}_{i|-ij}(t) = & {\bm z}_i(t) - E \{ {\bm z}_i(t) | {\bm z}_{-ij}(t) \} \, , 
	    \label{neweq34} \\
	 {\bm e}_{j|-ij}(t) = & {\bm z}_j(t) - E \{ {\bm z}_j(t) | {\bm z}_{-ij}(t) \} \, . \label{neweq35}
\end{align} 
Then by multi-attribute graphical modeling (Sec.\ \ref{MAMRV}),
\begin{equation}  \label{neweq36}
    {\bm \Omega}_y^{(ij)} = {\bm 0} \; \Leftrightarrow \; E\{ {\bm e}_{i|-ij}(t) {\bm e}_{j|-ij}^\top(t) \} = {\bm 0} \, .
\end{equation}
Define $m(p-2)$-vector
\begin{align}
   \tilde{\bm x}_{-ij;t,d} = & \{ x_k(s) \, : \, k \in V \textbackslash \{i, j \} \, ,
	     t-d \le s \le t \} \, , \label{neweq37} \\  
   e_{xi|-ij}(t') = & x_i(t') - E \{ x_i(t') | \tilde{\bm x}_{-ij;t,d} \} \, ,
	    \label{neweq38} \\
	 e_{xj|-ij}(t') = & x_j(t') - E \{ x_j(t') | \tilde{\bm x}_{-ij;t,d} \} \, . \label{neweq39}
\end{align} 
Notice that $e_{xi|-ij}(t')$ above is an element of ${\bm e}_{i|-ij}(t)$ defined in (\ref{neweq34}) for any $t-d \le t' \le t$. Then by (\ref{neweq24}) and (\ref{neweq36}), we have
\begin{align}  
  {\bm \Omega}_y^{(ij)} & = {\bm 0} \;  \Leftrightarrow \; 
	 E\{ e_{xi|-ij}(t_1) e_{xj|-ij}(t_2) \} =0, \nonumber \\
	  & \quad\quad \mbox{for } \;\; t-d \le t_1, t_2 \le t.  \label{neweq40}
\end{align}
This proves the desired result. $\quad \blacksquare$

{\it Remark 1}. It follows from (\ref{neweq40}) that if we let $ d \uparrow \infty$, then checking if ${\bm \Omega}_y^{(ij)} = {\bm 0}$ to ascertain (\ref{neweq36}) becomes a surrogate for checking if the last equivalence in (\ref{neweq14}) holds true for TSGGM estimation without using frequency-domain methods followed e.g., in \cite{Jung2015a, Tugnait2022a, Tugnait2025c}. For ``large'' $d$, with ${\bm \Omega}_y^{(ij)} = {\bm 0}$ as surrogate for $[{\bm S}_x^{-1}(f)]_{ij} = 0 \; \forall f \in [0,1]$, in this paper we wish to learn the graph ${\cal G} = \left( V, {\cal E} \right)$ where we take 
\[ 
    \{i,j\} \not\in {\cal E} \Leftrightarrow {\bm \Omega}_y^{(ij)} = {\bm 0}
\] 
instead of $\{i,j\} \not\in {\cal E} \Leftrightarrow [{\bm S}_x^{-1}(f)]_{ij} = 0 \; \forall f \in [0,1]$ (as e.g., in \cite{Tugnait2022a}). Time-domain methods allow us to address inference of CIGs for both SRD and LRD sequences quantitatively as we show in this paper, unlike frequency-domain methods of \cite{Jung2015a, Tugnait2022a, Tugnait2025c} whose analysis is restricted to SRD sequences. $\quad \Box$

{\it Remark 2}. By an abuse of notation, let  ${\bm \Omega}_{y,d}$ denote ${\bm \Omega}_y$ with the dependence on $d$ made explicit. If $d_1 < d_2$, then ${\bm \Omega}_{y,d_1}^{(ij)}$ is a submatrix of ${\bm \Omega}_{y,d_2}^{(ij)}$. By \cite[Remark 2.3]{Dahlhaus2000}, $[{\bm S}_x^{-1}(f)]_{ij} \equiv 0 \; \Leftrightarrow \; \lim_{d \rightarrow \infty} \|{\bm \Omega}_{y,d}^{(ij)} \|_F =0$, implying that $\|{\bm \Omega}_{y,d}^{(ij)} \|_F =0$ for any $d \ge 0$ and therefore  $\{i,j\} \not\in {\cal E}$.  But if  $\{i,j\} \in {\cal E}$, then $[{\bm S}_x^{-1}(f)]_{ij} \not\equiv 0$, but $\|{\bm \Omega}_{y,d}^{(ij)} \|_F$ could be zero. How large should $d$ be so that the set of edges for which ${\bm \Omega}_{y,d}^{(ij)} = {\bm 0}$ yields the set of edges for which $[{\bm S}_x^{-1}(f)]_{ij} = 0 \; \forall f \in [0,1]$? This question is yet unresolved. We show in Sec.\ \ref{CC} that the computational complexity of the ADMM optimization algorithm used to estimate ${\bm \Omega}_y$ from data is ${\cal O}((d+1)^3 p^3)$. In Sec.\ \ref{TA}, under some technical conditions, we provide an upperbound on the Frobenius norm of the error in estimating ${\bm \Omega}_y$, and it is ${\cal O}((d+1)^2/\sqrt{n-d})$ for SRD and a class of LRD sequences (see Theorem 1 and Remark 6 in Sec.\ \ref{TA}). Note that the number of unknown parameters in ${\bm \Omega}_y$ increase as ${\cal O}((d+1)^2)$. As $d$ increases, both computational complexity and the Frobenius error norm in estimating ${\bm \Omega}_y$ increase. But from a graph recovery viewpoint, the empirical results based on the synthetic data examples in Sec.\ \ref{SRDM2} and Table \ref{table2} show that as one increases $d$, the graph recovery performance metrics (the $F_1$ score and the Hamming distance between the true and estimated edge-sets) improve significantly at first, but then they plateau out with increasing $d$. This observation suggests the heuristic that the user should try several values of $d$ and pick the smallest value, say $d_0$, beyond which (e.g., for $d=d_0+1$) the estimated graph is only marginally different from the estimated graph for $d=d_0$. In practice, since the true graph is unknown, one cannot compute the $F_1$ score or the Hamming distance. For real data, what constitutes marginal difference would be problem and user dependent. A financial time series based real data results are presented in Sec.\ \ref{NEreal} where Fig.\ \ref{figreal2} shows the estimated graphs ($p=$97) for $d=3$, 4 and 5, and where the strengths of the edge connections, based on normalized $\| \hat{\bm \Omega}_{y \lambda}^{(k \ell)} \|_F/(\max_{k, \ell: ~k \ne \ell} \| \hat{\bm \Omega}_{y \lambda}^{(k \ell)} \|_F)$ for the edge $\{ k, \ell \}$,  are color-coded ($\hat{\bm \Omega}_{y \lambda}$ denotes the estimate of ${\bm \Omega}_y$ obtained using a regularization parameter $\lambda$ as discussed in Sec.\ \ref{PNLL1}). A visual examination shows very minor differences in the three graphs. The number of distinct edges are 347, 307 and 310 for $d=3$, 4 and 5, respectively, with differences being in weaker edges (smaller $\| \hat{\bm \Omega}^{(k \ell)} \|_F$ values).   $\quad \Box$

\subsection{Time Series Model} \label{TSM}
To quantify the dependence structure of $\{ {\bm x}(t) \}$, we will follow \cite{Chen16}; other possibilities include \cite{Shu19}. Assume
\begin{itemize}
\setlength{\itemindent}{0.1in}
\item[(A1)] Assume that $\{ {\bm x}(t) \}$ of Sec.\ \ref{MTS} obeys 
\begin{equation}  \label{neweq100}
  {\bm x}(t) = \sum_{i=0}^\infty {\bm A}_i {\bm e}(t-i) \, , 
\end{equation}
where $\{ {\bm e}(t) \}$ is i.i.d., Gaussian, zero-mean with identity covariance, ${\bm e}(t) \in \mathbb{R}^p$, $\, {\bm A}_i \in \mathbb{R}^{p \times p}$, and 
\begin{equation}  \label{neweq101}
  \max_{1 \le q \le p} \sqrt{ \sum_{k=1}^p ([{\bm A}_i]_{qk})^2 } \le \frac{c_a(p)}{(\max(1,i))^\gamma}
\end{equation}
for all $i \ge 0$, some $c_a(p) \in (0, \infty)$, and $\gamma > \frac{1}{2}$.
\end{itemize}

By assumption (A1), it follows that with ${\bm y}(t)$ as in (\ref{neweq31}), ${\bm y}(t) = \sum_{i=0}^\infty {\bm B}_i {\bm e}(t-i)$, ${\bm e}(t) \in \mathbb{R}^{p}$ is as in assumption (A1), $m=d+1$, $\, {\bm B}_i \in \mathbb{R}^{(mp) \times p}$, for some ${\bm B}_i$'s such that
\begin{equation}  \label{neweq102}
  \max_{1 \le q \le mp} \sqrt{ \sum_{k=1}^{p} ([{\bm B}_i]_{qk})^2 } \le \frac{{c}_a(p)}{(\max(1,i))^\gamma}
\end{equation}
for all $i \ge 0$, with ${c}_a(p)$ and $\gamma$ as in assumption (A1).

By assumption (A1), $|[{\bm A}_i]_{qk}| = {\cal O}(i^{-\gamma})$, $ i \ge 1$. In assumption (A1), the constant $c_a(p)$ may be a function of $p$. For instance, suppose that $|[{\bm A}_i]_{qk}| \le \bar{a} i^{-\gamma}$ $\forall \, i,q,k \ge 1$, for some $0 < \bar{a} < \infty$. Then $c_a(p) = {\cal O}(\sqrt{p})$. For the specific models considered in Secs.\ \ref{AR} and \ref{ARF} and used in our numerical examples, the condition $|[{\bm A}_i]_{qk}| \le \bar{a} i^{-\gamma}$ holds true. If ${\bm A}_i$ has  a block-structure (as e.g., resulting from the model (\ref{VAR1}) considered in Sec.\ \ref{SRDM1}) where the block-size is fixed independent of $p$, then $c_a(p)$ is a constant.

We have ${\bm R}_{x}( \tau ) = \mathbb{E} \{ {\bm x}(t + \tau) {\bm x}^T(t ) \} = \sum_{i=0}^\infty {\bm A}_{i+\tau} {\bm A}_i^\top$. Then by \cite[Lemma VI.1, Supplement]{Chen16}, for every $i,j \in [p]$, $|[{\bm R}(\tau)]_{ij}| = {\cal O}(\tau^{-\gamma})$, ${\cal O}(\tau^{-\gamma} \ln (\tau))$ and ${\cal O}(\tau^{1-2\gamma})$ for $\gamma > 1$, $\gamma =1$ and $\frac {1}{2} < \gamma < 1$, respectively, for $\tau \ge 1$.  For $\gamma > 1$, $[{\bm R}(\tau)]_{ij}$ is (absolutely) summable $\forall \, i,j \in [p]$, hence, $\{ {\bm x}(t) \}$ is SRD. For $\gamma \in (\frac{1}{2},1 ]$, $\sum_{\tau=-\infty}^\infty |[{\bm R}(\tau)]_{ij}|$ may not be finite for some $i,j \in [p]$, implying $\{ {\bm x}(t) \}$ is LRD \cite{Chen16, Kechagias2015, Pipiras}.

We now consider specific models that result in (\ref{neweq100}) satisfying assumption (A1) for different values of $\gamma$.
\subsubsection{Vector ARMA models} \label{AR}
In the SRD case ($\gamma >1$), assumption (A1) is satisfied if ${\bm x}(t)$ is generated by an asymptotically stable vector autoregressive moving average (ARMA) model satisfying 
\begin{equation}  \label{neweq1001}
   {\bm x}(t) = \sum_{i=1}^{q} \tilde{\bm A}_i {\bm x}(t-i) +\sum_{i=0}^{r} \tilde{\bm B}_i {\bm e}(t-i) 
\end{equation}
where $\{ {\bm e}(t) \}$ is as in (\ref{neweq100}) and $\tilde{\bm B}_0$ is full rank. The model in (\ref{neweq1001}) is called the ARMA($q,r$) model. Let ($z \in \mathbb{C}$)
\begin{equation}  \label{neweq1003}
   {\cal A}(z) := {\bm I}_p - \sum_{i=1}^{q} \tilde{\bm A}_i z^{-i} \, , \quad 
	{\cal B}(z) := \sum_{i=0}^{r} \tilde{\bm B}_i z^{-i} \, .
\end{equation}
We assume that $\mbox{det}({\cal B}(z)) \ne 0$ for any $|z|=1$ and $\mbox{det}({\cal A}(z)) \ne 0$ for any $|z| \ge 1$.
Consider the roots of $\mbox{det}({\cal A}(z)) = 0$. For stationarity of ${\bm x}(t)$ we need these roots $\bar{z}_\ell$'s ($\ell \in [pq]$) to satisfy $|\bar{z}_\ell| < 1$. Suppose that the maximum magnitude root is $\bar{z}_{\max}$ with multiplicity $\bar{m}$. Then by \cite[p.\ 186]{Ant1997}, the impulse response sequence $\{{\bm A}_i\}_{i=0}^\infty$ of (\ref{neweq1001}) (see also  (\ref{neweq100})) satisfies 
\begin{equation}  \label{neweq10010}
   \|{\bm A}_i\|_F \le a ~ i^{ \bar{m}-1} ~ |\bar{z}_{\max}|^i \, , \quad i \ge 1, \quad 1 \le \bar{m} \le pq \, , 
\end{equation}
for some positive $a < \infty$. Then there exists $0 < b < \infty$ and $1 < \gamma < \infty$ such that $ a i^{ \bar{m}-1} |\bar{z}_{\max}|^i \le b \, i^{-\gamma}$ for $i \ge 1$, thereby satisfying assumption (A1). [Proof: Pick $\gamma > 1$ and then pick $i_0$ such that for any $i \ge i_0$, $i^{ \bar{m}-1} |\bar{z}_{\max}|^i \le i^{-\gamma}$. Such a solution for $i_0$ exists. The desired inequality is equivalent to $\gamma \le -(\bar{m}-1) + \frac{i}{\ln(i)} (-\ln(|\bar{z}_{\max}|))$ for $i \ge i_0$. Now $-\ln(|\bar{z}_{\max}|)> 0$, $\frac{i}{\ln(i)}$ is increasing in $i$ for $\ln(i) > 1$ and $\lim_{i \rightarrow \infty} \frac{i}{\ln(i)} = \infty$. Hence, for a fixed $\gamma > 1$, there exists an $i_0$ such that $i^{ \bar{m}-1} |\bar{z}_{\max}|^i \le i^{-\gamma}$ for any $i \ge i_0$. Now set $b = \max \left\{ \max_{1 \le i \le i_0} \left\{ a |\bar{z}_{\max}|^i \, i^{\bar{m}-1+\gamma} \right\}, 1 \right\}$,  leading to $\|{\bm A}_i\|_F \le b \, i^{-\gamma}$ for $i \ge 1$, proving the claim.]

The PSD of the ARMA($q,r$) model (\ref{neweq1001}) is given by  
\begin{align}
  {\bm S}_x(f) & = {\cal A}^{-1}(e^{\iota 2 \pi f}) {\cal B}(e^{\iota 2 \pi f}) {\cal B}^H(e^{\iota 2 \pi f})
	                         ({\cal A}^{-1}(e^{\iota 2 \pi f}))^H \, . \label{neweq10013}
\end{align}
Since $\mbox{det}({\cal B}(z)) \ne 0$ for any $|z|=1$, ${\bm S}_x^{-1}(f)$ exists. By \cite[Sec.\ 2.2]{Basu15}, ${\bm S}_x(f)$ is bounded and continuous in $f$ with ${\bm S}_x(f) \succ {\bm 0}$ and ${\bm S}_x^{-1}(f) \succ {\bm 0}$ for $f \in [-0,5, 0.5]$. By \cite[(2.6)]{Basu15}, we have 
\begin{align}
  \min_f \phi_{\min}({\bm S}_x(f)) & 
	\ge \frac{ \min_f \phi_{\min} \Big( {\cal B}^H(e^{\iota 2 \pi f}) {\cal B}(e^{\iota 2 \pi f}) \Big) }
	        { \max_f \phi_{\max} \Big( {\cal A}^H(e^{\iota 2 \pi f}) {\cal A}(e^{\iota 2 \pi f}) \Big) } \label{neweq10015}  \\
 \max_f \phi_{\max}({\bm S}_x(f)) & 
	\le \frac{ \max_f \phi_{\max} \Big( {\cal B}^H(e^{\iota 2 \pi f}) {\cal B}(e^{\iota 2 \pi f}) \Big) }
	        { \min_f \phi_{\min} \Big( {\cal A}^H(e^{\iota 2 \pi f}) {\cal A}(e^{\iota 2 \pi f}) \Big) } \label{neweq10016}
\end{align}
where the $\max$ and $\min$ operations are over $f \in [-0.5,0.5]$ since ${\bm S}_x(f)$, ${\cal A}(e^{\iota 2 \pi f})$ and ${\cal B}(e^{\iota 2 \pi f})$ are periodic in $f$ with period $1$. Since $\mbox{det}({\cal B}(z)) \ne 0$ for any $|z|=1$, 
\begin{align}
  \mbox{det}({\cal B}^H(e^{\iota 2 \pi f}) {\cal B}(e^{\iota 2 \pi f}))  
   & = | \mbox{det}({\cal B}(e^{\iota 2 \pi f}) |^2 > 0
	                          \label{neweq10020}
\end{align} 
and therefore, all eigenvalues of Hermitian ${\cal B}^H(e^{\iota 2 \pi f}) {\cal B}(e^{\iota 2 \pi f})$ are positive for $f \in [-0,5, 0.5]$. By continuity of the eigenvalues of ${\cal B}^H(e^{\iota 2 \pi f}) {\cal B}(e^{\iota 2 \pi f})$ in $f$ and compactness of the interval $[-0,5, 0.5]$, the maximum and the minimum of $\phi_{\max} \Big( {\cal B}^H(e^{\iota 2 \pi f}) {\cal B}(e^{\iota 2 \pi f}) \Big)$ are achieved for some $\bar{f}$ and $\underbar{f}$ in $[-0,5, 0.5]$, respectively. Similar considerations apply to ${\cal A}^H(e^{\iota 2 \pi f}) {\cal A}(e^{\iota 2 \pi f})$. Thus, ${\bm S}_x(f) \succ {\bm 0}$ and ${\bm S}_x^{-1}(f) \succ {\bm 0}$.

\subsubsection{Vector ARFIMA models} \label{ARF} There are several variations of vector autoregressive fractionally integrated moving average (ARFIMA) models which are all equivalent in the scalar case \cite{Lobato1997, Taqqu2003,  Shimotsu2007, Pipiras}. They are also known as FARIMA (fractional autoregressive  integrated moving average) models \cite{GS2007, Pipiras}.  They generate both short range and long range dependencies through ARMA and fractional integration modeling, respectively \cite{Hosking1981,  Shimotsu2007, GS2007, Kechagias2015, Pipiras}. We consider the following version from \cite{Shimotsu2007}.  Let $\{{\bm u}(t)\}$ be an ARMA($q,r$) sequence as in (\ref{neweq1001}) (with a notation change), given by
\begin{equation}  \label{neweq10030}
   {\bm u}(t) = \sum_{i=1}^{q} \tilde{\bm A}_i {\bm u}(t-i) +\sum_{i=0}^{r} \tilde{\bm B}_i {\bm e}(t-i) 
\end{equation}
where $\tilde{\bm A}_i$, $\tilde{\bm B}_i$ and $\{ {\bm e}(t) \}$ are as in Sec.\ \ref{AR}. We generate the $i$th component $x_i(t)$ of ${\bm x}(t)$ by filtering the $i$th component $u_i(t)$ of ${\bm u}(t)$, $i \in [p]$, as
\begin{align}
  x_i(t) & = \sum_{\ell =0}^\infty c_{i \ell} u_i(t-\ell) \,,  \label{neweq10031} \\
	  c_{i \ell}&  := \frac{\Gamma(\ell + d_i)}{ \Gamma(\ell + 1) \Gamma(d_i) } \,, \quad d_i \in (0,0.5) \, ,
	                          \label{neweq10032}
\end{align}
where $\Gamma(z)$ is the Gamma function $\Gamma(z) := \int_{0}^\infty t^{z-1} e^{-t} \, dt$, $z \in \mathbb{C}$, $\mbox{Re}(z) > 0$. With $L$ denoting the backward-shift operator $L {\bm u}(t) := {\bm u}(t-1)$, we have \cite{Hosking1981, Shimotsu2007, GS2007, Pipiras}
\begin{align}
  x_i(t) = (1-L)^{-d_i} u_i(t) \, , \quad (1-L)^{d_i} x_i(t) = u_i(t)
	                          \label{neweq10034}
\end{align}
and if $\{ u_i(t) \}$ is stationary then so is $\{ x_i(t) \}$ for $0 < d_i < 0.5$ \cite{Hosking1981, Shimotsu2007, GS2007, Pipiras}. Note that $\lim_{\ell \rightarrow \infty} c_{i \ell} = {\cal O}(\ell^{d_i-1})$ \cite{Taqqu2003, GS2007}. This ARFIMA model is called the ARFIMA($q,d_i,r$) model.

The PSD of $\{ x_i(t) \}$ is given by \cite{Hosking1981, GS2007, Pipiras}
\begin{align}
  S_{x_i}(f) & = (1-e^{-\iota 2 \pi f})^{-d_i} S_{u_i}(f) (1-e^{\iota 2 \pi f})^{-d_i} \, .
	                          \label{neweq10035}
\end{align}
Define a diagonal matrix ${\bm D} = \mbox{diag}\{d_1, d_2, \cdots , d_p \} \in \mathbb{R}^p$ and for a scalar $a$, let $a^D$ denote $\mbox{diag}\{a^{d_1}, a^{d_2}, \cdots , a^{d_p} \}$. The PSD of the ARFIMA process (\ref{neweq10030})-(\ref{neweq10034}) is given by \cite{Shimotsu2007}
\begin{align}
  {\bm S}_x(f) & = (1-e^{-\iota 2 \pi f})^{-{\bm D}} {\bm S}_u(f) (1-e^{\iota 2 \pi f})^{-{\bm D}} 
	                          \label{neweq10037}
\end{align}
where 
\begin{align}
  {\bm S}_u(f) & = {\cal A}^{-1}(e^{\iota 2 \pi f}) {\cal B}(e^{\iota 2 \pi f}) {\cal B}^H(e^{\iota 2 \pi f})
	                         ({\cal A}^{-1}(e^{\iota 2 \pi f}))^H \, . \label{neweq10039}
\end{align} 
It follows from Sec.\ \ref{AR}, (\ref{neweq10037}) and (\ref{neweq10039}) that ${\bm S}_x(f) \succ {\bm 0}$ and ${\bm S}_x^{-1}(f) \succ {\bm 0}$ for $f \in [-0.5,0.5] \setminus \{0\}$, and some or all elements of ${\bm S}_x^{-1}(0)$ are zero and some or all elements of $\lim_{f \rightarrow 0} {\bm S}_x(f)$ tend to $\infty$. As $f \rightarrow 0^+$, $(1-e^{-\iota 2 \pi f})^{-{\bm D}} \rightarrow (\iota 2 \pi f)^{-{\bm D}}$, therefore,
\begin{align}
  \lim_{f \rightarrow 0^+} {\bm S}_x(f) & \propto  
	  f^{-{\bm D}} {\cal A}^{-1}(1) {\cal B}(1) {\cal B}^H(1) ({\cal A}^{-1}(1))^H f^{-{\bm D}} \, .
	                          \label{neweq10040}
\end{align}
The expression (\ref{neweq10040}), together with (\ref{neweq10037}) and (\ref{neweq10039}), satisfies the spectral domain definition of a multivariate stationary LRD process given in \cite[Definition 2.2]{Kechagias2015}. It then follows from \cite[Prop.\ 3.1]{Kechagias2015} that for large $|\tau|$, the autocorrelation function ${\bm R}_x(\tau)$ of $\{{\bm x}(t) \}$ satisfies
\begin{align}
  [{\bm R}_x(\tau)]_{\ell k} & \propto [\bar{\bm R}]_{\ell k} \tau^{d_\ell + d_k -1} \, , \quad \ell ,k \in [p],
	                          \label{neweq10042}
\end{align}
for some $\bar{\bm R} \in \mathbb{R}^{p \times p}$ such that $[\bar{\bm R}]_{ii} \ne 0$, $i \in [p]$. Thus,  $[{\bm R}_x(\tau)]_{\ell k} = {\cal O}(\tau^{d_\ell+d_k-1})$ as $|\tau| \rightarrow \infty$. Setting $\gamma = 1- \max_{\ell \in [p]} d_\ell$, it then follows from \cite[Prop.\ 3.1]{Kechagias2015} that  $[{\bm R}_x(\tau)]_{\ell k} = {\cal O}(\tau^{d_\ell +d_k-1}) = {\cal O}(\tau^{1-2 \gamma})$, $|\tau| > 0$. For $d_\ell \in (0,0.5)$, $\ell \in [p]$, we have $\gamma \in (0.5,1)$ and therefore, ${\bm R}_x(\tau)$ is not absolutely summable, implying an LRD process \cite{ Chen16, Basu15, Kechagias2015, Pipiras}.

{\it Remark 3}. For the ARFIMA model we have ${\bm S}_x^{-1}(f) = (1-e^{\iota 2 \pi f})^{{\bm D}} {\bm S}_u^{-1}(f) (1-e^{-\iota 2 \pi f})^{{\bm D}}$ with the ARMA inverse PSD ${\bm S}_u^{-1}(f) = {\cal A}^H(e^{\iota 2 \pi f}) ({\cal B}^{-1}(e^{\iota 2 \pi f}))^H {\cal B}^{-1}(e^{\iota 2 \pi f}) {\cal A}(e^{\iota 2 \pi f})$. Therefore, 
\begin{align}
 [{\bm S}_x^{-1}(f)]_{ij} = &  [{\bm S}_u^{-1}(f)]_{ij} (1-e^{\iota 2 \pi f})^{d_i} (1-e^{-\iota 2 \pi f})^{d_j} 
	                          \label{neweq10045}
\end{align}
for $i,j \in [p]$, implying $[{\bm S}_x^{-1}(f)]_{ij} = 0 ~ \Leftrightarrow ~ [{\bm S}_u^{-1}(f)]_{ij} = 0$ for any $f \in [-0.5,0.5] \setminus \{0\}$. Therefore, the conditional independence structure of the ARFIMA process $\{{\bm x}(t)\}$ coincides with that of the underlying ARMA process $\{{\bm u}(t)\}$.
 $\quad \Box$

\subsection{Summary and Discussion of Model Assumptions} \label{STSM}
To summarize, we require $\{ {\bm x}(t) \}$ to be a stationary, zero-mean, $p-$dimensional multivariate Gaussian time series of dimension $p$ such that assumption (A1) holds and its PSD ${\bm S}_x(f) \succ {\bm 0}$. The vector ARMA and ARFIMA models discussed in Secs.\ \ref{AR} and \ref{ARF}, respectively, generate such sequences. Assumption (A1) quantifies the temporal dependence structure of $\{ {\bm x}(t) \}$. If $d_i \ge 0.5$ in (\ref{neweq10032}), then $x_i(t)$ in (\ref{neweq10031}) is non-stationary \cite{Hosking1981, Shimotsu2007, GS2007, Pipiras} and $\gamma \le 0.5$ in (\ref{neweq101}). Later, to establish some theoretical results in Sec.\ \ref{TA}, we requires additional assumptions (A2) and (A3) whose implications are discussed therein after their statements. 

The assumptions that $\{ {\bm x}(t) \}$ is stationary and linear (i.e., generated via (\ref{neweq100})) are critical to the results of this paper. 

In this paper we are interested in CIGs for time series, as defined in \cite{Dahlhaus2000}. As noted in  \cite{Dahlhaus2000}, if $\{ {\bm x}(t) \}$  is not necessarily Gaussian, then the zeros of ${\bm S}_x^{-1}(f)$ encode a partial correlation graph which is not necessarily a CIG since uncorrelation does not necessarily imply independence for non-Gaussian processes. For non-Gaussian sequences, our time-delay embedding approach yields an approximation to the underlying time series partial correlation graph. In Lemma 1, one replaces the conditional expectations with the corresponding optimum linear estimators, e.g., $E \{ {\bm z}_i(t) | {\bm z}_{-ij}(t) \}$ in (\ref{neweq34}) is replaced with the optimum linear estimator of ${\bm z}_i(t)$ based on a linear weighted sum of elements in ${\bm z}_{-ij}(t)$ (see \cite[Sec.\ 2]{Dahlhaus2000}), and similarly, in (\ref{neweq38}), $E \{ x_i(t') | \tilde{\bm x}_{-ij;t,d} \}$ is replaced with the optimum linear estimator of $x_i(t')$ based on a linear weighted sum of elements in $\tilde{\bm x}_{-ij;t,d}$. For Gaussian processes the conditional expectations equal the optimum linear estimators. In Sec.\ \ref{SRDM5} we present some simulation results for non-Gaussian sequences (using either exponential or Laplacian ${\bm e}(t)$ in  (\ref{neweq100}), instead of Gaussian ${\bm e}(t)$), with all sequences, Gaussian or non-Gaussian, having identical PSDs and therefore, identical partial correlation graphs. These results show comparable performances for Gaussian and non-Gaussian sequences in graph recovery.

\section{Objective Function and Its Optimization} \label{PNLL}
As noted in Remark 1, under delay embedding, we have a multi-attribute graph learning problem which has been addressed in \cite{Kolar2014, Tugnait21a} with lasso regularization and in \cite{Tugnait2025b} with non-convex log-sum and SCAD regularizations, except that \cite{Kolar2014, Tugnait21a, Tugnait2025b} address estimation from i.i.d.\ data whereas in this paper we have dependent data. Nevertheless we will use the log-likelihood of \cite{Kolar2014, Tugnait21a, Tugnait2025b} as our pseudo log-likelihood for dependent data. We show in Sec.\ \ref{TA} (see Theorems 1-3 therein) that when one chooses ${\bm \Omega}_y$ to minimize the resulting objective function, the estimator of ${\bm \Omega}_y$ has several desirable properties resulting in estimator consistency (Theorem 1), uniqueness (Theorem 2) and graph recovery (Theorem 3). Also, the resulting objective function and its optimization is just as in \cite{Tugnait2025b} to which one should refer to for details. In this section we introduce the objective and the regularization functions to introduce notation (and MCP, not discussed in \cite{Tugnait2025b}), and to  provide some details regarding optimization that are used later for theoretical analysis in Sec.\ \ref{TA} and for numerical results in Secs.\ \ref{NE} and \ref{NEreal}

\subsection{Penalized Negative Pseudo Log-Likelihood} \label{PNLL1}
Consider a finite set of data comprised of $n$ zero-mean observations  ${\bm x}(t)$, $t= 1,2, \cdots , n$. Pick $d > 0$ and as in (\ref{neweq31}), construct ${\bm y}(t)$ for $t=d+1, d+2, \cdots , n$ with sample size $\bar{n} = n-d$. Define the sample covariance 
\begin{equation}  \label{moresm} 
	\hat{\bm{\Sigma}}_y = \frac{1}{\bar{n}} \sum_{t=d+1}^{n} {\bm y}(t) {\bm y}^\top(t) \, . 
\end{equation}
If the vector sequence $\{ {\bm y}(t) \}_{t=d+1}^{n}$ were i.i.d., the negative log-likelihood (up to some constants) would be given by \cite{Tugnait21a}
\begin{align} 
    {\cal L}({\bm \Omega}_y) 
					&	:= - \ln (|\bm{\Omega}_y|) 
		        + {\rm tr} \left( \hat{\bm \Sigma}_y \bm{\Omega}_y \right) \, .    	 \label{eqth2_10}
\end{align}
We will still use this expression as a pseudo log-likelihood and consider the penalized negative pseudo log-likelihood $\bar{\cal L}({\bm \Omega}_y)$ given by
\begin{align}  
   \bar{\cal L}({\bm \Omega}_y) & =  {\cal L}({\bm \Omega}_y) + \alpha P_e({\bm \Omega}_y) 
	   + (1-\alpha) P_g({\bm \Omega}_y) , \label{eqth2_20} \\
	   P_e({\bm \Omega}_y) &  =   \sum_{i \ne j}^{mp} 
		  \rho_\lambda \left( \Big| [ {\bm{\Omega}_y} ]_{ij} \Big| \right)  , \label{eqth2_20a} \\
		P_g({\bm \Omega}_y) &  = m  \, \sum_{ k \ne \ell}^p \; 
		    \rho_\lambda \left( \| {\bm \Omega}_y^{(k \ell)} \|_F \right) \label{eqth2_20b} 
\end{align}
where ${\bm \Omega}^{(k \ell )}_y \in \mathbb{R}^{m \times m}$ is defined as in (\ref{neweq32}), $\lambda > 0$, $\alpha \in [0,1]$, $m $ in (\ref{eqth2_20b}) reflects the number of group variables \cite{Yuan2007}, and for $u \in \mathbb{R}$, $\rho_\lambda(u)$ is a penalty function that is function of $|u|$. In (\ref{eqth2_20a}), the penalty term is applied to each off-diagonal element of $\bm{\Omega}$ and in (\ref{eqth2_20b}), the penalty term is applied to the off-block-diagonal group of $m^2$ terms via ${\bm \Omega}^{(k \ell)}$. The parameter $\alpha \in [0,1]$ ``balances'' element-wise and group-wise penalties \cite{Friedman2010a, Friedman2010b, Tugnait21a}, yielding a convex combination of the two.

\subsubsection{Penalty Functions} \label{PF}
We consider the following:
\begin{itemize}
\item {\it Lasso}. For some $\lambda > 0$,   $\rho_\lambda(u) = \lambda |u|$, $u \in \mathbb{R}$. It is a convex function that is widely used. 
\item {\it Log-sum} \cite{Candes2008}. For some $\lambda > 0$ and $1 \gg \epsilon > 0$, $\rho_\lambda(u) = \lambda \epsilon \, \ln \left( 1 + \frac{|u|}{\epsilon} \right)$. It is a non-convex function.
\item {\it Smoothly Clipped Absolute Deviation (SCAD)}. \cite{Fan2001, Lam2009} For some $\lambda > 0$ and $a > 2$, $\rho_\lambda(u) = \lambda | u |$ for $|u| \le \lambda$, $= (2 a \lambda | u |- | u |^2 - \lambda^2)/(2 (a-1))$ for $\lambda < |u| < a \lambda$, and $=\lambda^2 (a+1)/2$ for $|u| \ge a$. It is a non-convex function.
We have $\rho_\lambda^\prime(u) = \frac{d \rho_\lambda (u)}{du} = 0$ for $|u| \ge a \lambda$.
\item {\it Minimax Concave Penalty (MCP)} \cite{Zhang2010}. For some $\lambda > 0$ and $b > 0$, $\rho_\lambda(u) = \lambda | u | - \frac{u^2}{2b}$ for $|u| \le b \lambda$, and $=\frac{b \lambda^2 }{2}$ for $|u| \ge b \lambda $. 
In \cite{Breheny2015} $b >1$ is taken. We have $\rho_\lambda^\prime(u)  = 0$ for $|u| \ge b \lambda$.
\end{itemize}
The MCP penalty has not been considered in \cite{Kolar2014, Tugnait21a, Tugnait2025b}. 
As shown in \cite[Sec.\ 2.2]{Loh2017}, $\rho_\lambda(u) = \rho_\lambda(-u)$  and differentiable for $u \ne 0$ with $\lim_{u \rightarrow 0^+} \rho_\lambda^\prime(u) = \lambda$. Furthermore, $\rho_\lambda(u) +\frac{\mu}{2} u^2$ is convex, for some $\mu \ge 0$.  It is shown in \cite[Appendix A.1]{Loh2017}, that $\mu = 0$ for lasso, $\mu =1/(a-1)$ for SCAD and $\mu = 1/b$ for MCP. For the log-sum penalty $\mu = \frac{\lambda}{\epsilon}$ \cite{Tugnait2025b}.
The above penalties also satisfy:
\begin{itemize}
\item[(i)] For some $C_\lambda > 0$ and $\delta_\lambda > 0$, 
\begin{equation}  \label{prop7} 
        \rho_\lambda(u) \ge C_\lambda |u| \mbox{  for  } |u| \le \delta_\lambda \, .
\end{equation}
\item[(ii)] $\frac{d \rho_\lambda (u)}{d |u|} \le \lambda$ for $u \ne 0$.
\end{itemize}
Property (ii) is straightforward to verify. For lasso,  $C_\lambda = \lambda$ and $\delta_\lambda = \infty$. For SCAD, $C_\lambda = \lambda$ and $\delta_\lambda = \lambda$. For the log-sum penalty, $C_\lambda = \frac{\lambda}{2}$ and $\delta_\lambda =\epsilon$ \cite{Tugnait2025b}. We take $C_\lambda = \frac{\lambda}{2}$ for all penalties (in proving Theorem 1), in which case (\ref{prop7}) holds for MCP with $\delta_\lambda = b \lambda$.

\begin{algorithm} 
\caption{ADMM Algorithm}
\label{alg0}

\algorithmicrequire{\; Data $\{{\bm x}(t)\}_{t=1}^n$, delay $d \ge 0$, $m=d+1$, regularization parameters $\lambda$ and $\alpha$, ADMM penalty parameter $\rho=\bar{\rho}$, tolerances $\tau_{abs}$ and $\tau_{rel}$, variable penalty factor $\bar{\phi}$, maximum number of iterations $r_{\max}$. Initial guess $\bar{\bm{\Omega}}$}.  \\
\algorithmicensure{\;\ estimated inverse covariance $\hat{\bm \Omega}_{y \lambda}$ and edge-set $\hat{\cal E}$}

\begin{algorithmic}[1] 
\STATE Construct ${\bm y}(t)$ for $t=d+1, d+2, \cdots , n$ with sample size $\bar{n} = n-d$, as in (\ref{neweq31}). Calculate sample covariance $\hat{\bm{\Sigma}}_y = \frac{1}{\bar{n}} \sum_{t=d+1}^{n} {\bm y}(t) {\bm y}^\top(t)$. For lasso, $\lambda_{e,ij} = \lambda$ ($\forall i,j \in [mp]$), $\lambda_{g, k \ell} = \lambda$ ($k, \ell \in [p]$). For non-convex penalties, set $\lambda_{e,ij}$ ($i,j \in [mp]$) and $\lambda_{g, k \ell}$ ($k, \ell \in [p]$) as in (\ref{lsp}), (\ref{scad1})-(\ref{scad2}), or (\ref{mcp1})-(\ref{mcp2}), for LSP, SCAD or MCP, respectively, using the initial guess $\bar{\bm{\Omega}}$.
\STATE Initialize: ${\bm U}^{(0)} = {\bm V}^{(0)} = {\bm 0}$, where ${\bm U}, {\bm V} \in \mathbb{R}^{(mp) \times (mp)}$, ${\bm \Omega}_{y \lambda}^{(0)}= \bar{\bm{\Omega}}$, $\rho^{(0)} = \bar{\rho}$.
\STATE converged = \FALSE, $r=0$
\WHILE{converged = \FALSE $\;$ \AND $\;$ $r \le r_{max}$,}
\STATE Eigen-decompose $\hat{\bm \Sigma}_y - \rho^{(r)} \left( {\bm V}^{(r)} - {\bm U}^{(r)} \right)$ as $\hat{\bm \Sigma}_y - \rho^{(r)} \left( {\bm V}^{(r)} - {\bm U}^{(r)} \right) = {\bm P}{\bm D}{\bm P}^\top$ with diagonal matrix ${\bm D}$ consisting of eigenvalues.  Define diagonal matrix $\tilde{\bm D}$ with $\ell$th diagonal element
$\tilde{\bm D}_{\ell \ell} = ( -{\bm D}_{\ell \ell} + \sqrt{ {\bm D}_{\ell \ell}^2 + 4 \rho^{(r)}  } \, )/(2 \rho^{(r)})$. Set $\bm{\Omega}_{y \lambda}^{(r+1)} = {\bm P} \tilde{\bm D} {\bm P}^\top$.
\STATE Define soft thresholding scalar operator $T(a, \beta) := (1-\beta/|a|)_+ a$  where $(a)_+ := \max(0,a)$. Set ${\bm A}^{(k \ell)} = (\bm{\Omega}_{y \lambda}^{(r+1)})^{(k \ell)} + ({\bm U}^{(r)})^{(k \ell)}$. The diagonal $m \times m$ subblocks of ${\bm V}$ are updated as 
\begin{align*}  
    [({\bm V}^{(r+1)})^{(kk)}]_{u v} & = 
	 \left\{ \begin{array}{l}
			    [\bm{A}^{(kk)}]_{uu}  \quad \mbox{  if } u=v\\
					\hspace*{-0.1in} T([{\bm A}_k^{(kk)}]_{uv}, \frac{ \alpha \lambda_{e,ij}}{\rho^{(r)}}) 
					   \,  \mbox{ if } u \neq v \end{array} \right.
\end{align*}
$k \in [p]$, $\; u,v \in [m]$, $i=(k-1)m+u$, $j=(k-1)m+v$. The off-diagonal $m \times m$ subblocks of ${\bm V}$ are updated as 
\begin{align*}  
   ({\bm V}^{(r+1)})^{(k \ell)} & = 
	 {\bm B} \Big( 1 - \frac{(1-\alpha) m \lambda_{g,k \ell}}
					    {\rho^{(r)} \| {\bm B} \|_F }  \Big)_+ 
\end{align*}
where $k \ne \ell \in [p]$, $m \times m$ ${\bm B}$ has its $(u,v)$th element as $[{\bm B}]_{uv}=T([{\bm A}^{(k \ell)}]_{uv}, \alpha \lambda_{e,ij}/\rho^{(r)} )$, $i=(k-1)m+u$, $j=(\ell-1)m+v$.
\STATE Dual update ${\bm U}^{(r+1)} = {\bm U}^{(r)} + \left( \bm{\Omega}_{y \lambda}^{(r+1)} - {\bm V}^{(r+1)} \right)$. 
\STATE Check convergence. Set tolerances
\begin{align*}
   \tau_{pri} = & mp \, \tau_{abs} + \tau_{rel} \, \max ( \| {\bm \Omega}_{y \lambda}^{(r+1)} \|_F , \| {\bm V}^{(r+1)} \|_F ) \\
  \tau_{dual} = & mp \, \tau_{abs} + \tau_{rel} \,  \| {\bm U}^{(r+1)} \|_F / \rho^{(r)} \, .
\end{align*}
Define $d_p = \| {\bm \Omega}_{y \lambda}^{(r+1)} - {\bm V}^{(r+1)} \|_F$ and $d_d = \rho^{(r)} \| {\bm V}^{(r+1)} - {\bm V}^{(r)} \|_F$.
If $( d_p \le \tau_{pri}) \; \AND \; (d_d \le \tau_{dual})$, set converged = \TRUE .
\STATE Update penalty parameter $\rho$ : $\rho^{(r+1)} = 2 \rho^{(r)}$ if $d_p > \bar{\phi} d_d$, $=\rho^{(r)} /2 $ if $d_d > \bar{\phi} d_p$, $=\rho^{(r)} $ otherwise.
We also need to set ${\bm U}^{(r+1)} = {\bm U}^{(r+1)}/2$ for $d_p > \bar{\phi} d_d$ and ${\bm U}^{(r+1)} = 2 {\bm U}^{(r+1)}$ for $d_d > \bar{\phi} d_p$.
\STATE $r \leftarrow r+1$
\ENDWHILE
\STATE For $k \ne \ell$, if $\|{\bm V}^{(k \ell)}\|_F > 0$, assign edge $\{ k, \ell\} \in \hat{\cal E}$, else $\{ k, \ell\} \not\in \hat{\cal E}$. Inverse covariance estimate $\hat{\bm \Omega}_{y \lambda} = {\bm V}$.
\end{algorithmic}
\end{algorithm} 

\subsection{Optimization} \label{OPT}
For non-convex $\rho_\lambda(u)$, we use a local linear approximation (LLA) as in \cite{Tugnait2025b}, following\cite{Zou2008, Lam2009}, to yield 
\begin{equation}
  \rho_{\lambda}(u) \approx \rho_{\lambda}(|u_0|) 
	 + \rho_\lambda^\prime(|u_0|) (|u| - |u_0|)
	 \, \Rightarrow \, \rho_\lambda^\prime(|u_0|) |u| \, ,
\end{equation}
where $u_0$ is an initial guess, and $\rho_\lambda^\prime(|u_0|)$ is the penalty function derivative at $|u| = |u_0|$. Therefore, with $u_0$ fixed, we consider only the last term above for optimization w.r.t.\ $u$. By \cite[Theorem 1]{Zou2008}, the LLA provides a majorization of the non-convex penalty, thereby yielding a majorization-minimization approach. Suppose the initial guess is $\bar{\bm{\Omega}}_y$, then with LLA, the original objective function is transformed to its convex LLA approximation
\begin{align} 
   \tilde{\cal L}({\bm \Omega}_y) & =  {\cal L}({\bm \Omega}_y) + \alpha \tilde{P}_e({\bm \Omega}_y) 
	   + (1-\alpha) \tilde{P}_g({\bm \Omega}_y) , \label{admm} 
\end{align}
\begin{align}
	   \tilde{P}_e({\bm \Omega}_y) &  =  \sum_{i \ne j}^{mp} 
		   \lambda_{e,ij} \Big| [ \bm{\Omega}_{y} ]_{ij} \Big|  , \label{admm1} 
\end{align}
\begin{align}
		\tilde{P}_g({\bm \Omega}_y) &  = m  \, \sum_{ k \ne \ell}^p \; 
		    \lambda_{g,k \ell} \| {\bm \Omega}_y^{(k \ell )} \|_F  \label{admm2} 
\end{align}
where $\lambda_{e,ij}  = \rho_\lambda^\prime(|[ \bar{\bm{\Omega}}_y ]_{ij}|)$ and $\lambda_{g,k \ell} = \rho_\lambda^\prime(\| \bar{\bm{\Omega}}_y^{(k \ell )} \|_F )$. 
We have for the log-sum penalty (LSP), 
\begin{align} \label{lsp}
   & \lambda_{e,ij} :=
	 \frac{\lambda \epsilon }{| [\bar{\bm \Omega}_y]_{ij} |  +\epsilon} \, ,  \quad
	\lambda_{g,k \ell} :=
	 \frac{\lambda \epsilon }{\| \bar{\bm \Omega}_y^{(k \ell)} \|_F  +\epsilon} \, ,
\end{align}
for SCAD,  
\begin{align} 
\lambda_{e,ij} = &
			 \left\{ \begin{array}{ll} \lambda , & \mbox{if  } | [\bar{\bm \Omega}_y]_{ij} | \le \lambda \\
				    \frac{ a \lambda - | [\bar{\bm \Omega}_y]_{ij} |}{ a-1} , 
									& \mbox{if  } \lambda < | [\bar{\bm \Omega}_y]_{ij} | \le a \lambda \\ 
						0 , & \mbox{if  } a \lambda < | [\bar{\bm \Omega}_y]_{ij} |  \end{array} \right.  \, , \label{scad1} \\
  \lambda_{g,k \ell} = &
			 \left\{ \begin{array}{ll} \lambda , & \mbox{if  } \| \bar{\bm \Omega}_y^{(k \ell)} \|_F \le \lambda \\
				    \frac{ a \lambda - \| \bar{\bm \Omega}_y^{(k \ell)} \|_F}{ a-1} , 
									& \mbox{if  } \lambda < \| \bar{\bm \Omega}_y^{(k \ell)} \|_F \le a \lambda \\ 
						0 , & \mbox{if  } a \lambda < \| \bar{\bm \Omega}_y^{(k \ell)} \|_F  \end{array} \right.  \, ,  \label{scad2}
\end{align}
and for MCP,
\begin{align} 
\lambda_{e,ij} = &
			 \left\{ \begin{array}{ll} \lambda - \frac{| [\bar{\bm \Omega}_y]_{ij} |}{b}, 
			 & \mbox{if  } | [\bar{\bm \Omega}_y]_{ij} | \le b \lambda \\
						0 , & \mbox{if  } b \lambda < | [\bar{\bm \Omega}_y]_{ij} |  \end{array} \right.  \, , \label{mcp1} 
\end{align}
\begin{align}
  \lambda_{g,k \ell} = &
			 \left\{ \begin{array}{ll} \lambda - \frac{\| \bar{\bm \Omega}_y^{(k \ell)} \|_F |}{b}, 
			 & \mbox{if  } \| \bar{\bm \Omega}_y^{(k \ell)} \|_F \le b \lambda \\ 
						0 , & \mbox{if  } b \lambda < \| \bar{\bm \Omega}_y^{(k \ell)} \|_F  \end{array} \right.  \, .  \label{mcp2}
\end{align}
For lasso, we have $\lambda_{e,ij} = \lambda$ $\forall i,j$ and $\lambda_{g,k \ell} = \lambda$ $\forall k, \ell$.

We use the ADMM approach as described in detail in Algorithm 1  in \cite[Sec.\ IV]{Tugnait2025b} to minimize $\bar{\cal L}({\bm \Omega}_y)$ in (\ref{eqth2_20}) w.r.t.\ ${\bm \Omega}_y$ for both convex lasso and non-convex LSP/SCAD/MCP regularizations.  The initial guess $\bar{\bm{\Omega}}_y$ for LLA is taken to be the lasso penalized solution. 
Using variable splitting \cite{Boyd2010}, consider
\begin{align} 
 \min_{\bm{\Omega}_y \succ {\bm 0}, {\bm V} } & \Big\{  {\cal L}({\bm \Omega}_y) + \alpha \tilde{P}_e({\bm V}) 
	   + (1-\alpha) \tilde{P}_g({\bm V})   \Big\} \label{eqth2_4000} \\
			& \quad\quad		 \mbox{ subject to  }  \bm{\Omega}_y = {\bm V}  \, ,    \label{eqth2_4001}   
\end{align}
where ${\bm V} \in \mathbb{R}^{(mp) \times (mp)}$ results from variable splitting, and in the penalties we use  ${\bm V}$ instead of ${\bm \Omega}_y$, adding the equality constraint ${\bm V}={\bm \Omega}$.
The scaled augmented Lagrangian \cite{Boyd2010} for this problem after variable splitting, is given by 
\begin{align}
 \bar{\cal L}_\rho &(\bm{\Omega}_y , {\bm V} , {\bm U}  ) =   
   {\cal L}( \bm{\Omega}_y )	+ \alpha \tilde{P}_e({\bm V}) 
	    \nonumber \\
		&  + (1-\alpha) \tilde{P}_g({\bm V}) + \frac{\rho}{2}  \| \bm{\Omega}_y - {\bm V} + {\bm U} \|^2_F \, , \label{admm0} 
\end{align}
where $ {\bm U}$ is the dual variable  and $\rho >0$ is the ``penalty parameter'' \cite{Boyd2010}. Given the results $ \bm{\Omega}^{(r)}, {\bm V}^{(r)}, {\bm U}^{(r)}$ of the $r$th iteration, in the $(r+1)$st iteration, an ADMM algorithm executes the following three updates:
\begin{itemize}
\item[(a)] $\bm{\Omega}_{y \lambda}^{(r+1)} \leftarrow \arg \min_{\bm{\Omega}_y} \, L_a(\bm{\Omega}_y) ,\;\;
           L_a(\bm{\Omega}_y) := {\cal L}( \bm{\Omega}_y ) + \frac{\rho}{2}  \| \bm{\Omega}_y - {\bm V}^{(r)} + {\bm U}^{(r)}\|^2_F $
\item[(b)] ${\bm V}^{(r+1)}  \leftarrow \arg \min _{ {\bm V} } L_b({\bm V}) , \;\;
          L_b({\bm V}) :=  \alpha \tilde{P}_e({\bm V}) 
	   + (1-\alpha) \tilde{P}_g({\bm V})   
					+ \frac{\rho}{2}  \| \bm{\Omega}_{y \lambda}^{(r+1)} - {\bm V} + {\bm U}^{(r)} \|^2_F$
\item[(c)] ${\bm U}^{(r+1)} \leftarrow {\bm U}^{(r)}  +
   \left( \bm{\Omega}_{y \lambda}^{(r+1)} - {\bm V}^{(r+1)} \right)$
\end{itemize}
A pseudo code for the ADMM algorithm used in this paper is given in Algorithm \ref{alg0} where we use the stopping (convergence) criterion following \cite[Sec.\ 3.3.1]{Boyd2010} and varying penalty parameter $\rho$ following \cite[Sec.\ 3.4.1]{Boyd2010}. See \cite{Tugnait21a} for further details; note that by construction, $\bm{\Omega}_{y \lambda}^{(r+1)}$ in step 5 of Algorithm \ref{alg0} is positive definite.

For non-convex penalties, we have an iterative solution: first solve with sparse group-lasso penalty, then use the LLA formulation and solve the resulting adaptive lasso type convex problem. In practice, just two iterations seem to be enough.

\subsubsection{Optimization Algorithm Summary} Our ADMM-based optimization algorithm is as follows.
\begin{itemize}
\item[1.] Given data $\{{\bm x}(t)\}_{t=1}^n$ and chosen delay $d \ge 0$, $m=d+1$, construct ${\bm y}(t)$ for $t=d+1, d+2, \cdots , n$ with sample size $\bar{n} = n-d$, as in (\ref{neweq31}). Calculate sample covariance $\hat{\bm{\Sigma}}_y = \frac{1}{\bar{n}} \sum_{t=d+1}^{n} {\bm y}(t) {\bm y}^\top(t)$. Initialize iteration $\tilde{m}=1$,  ${\bm \Omega}_{y \lambda}^{(0)} = (\mbox{diag}({\hat{\bm \Sigma}}_y))^{-1}$, $\bar{\bm{\Omega}}=  {\bm \Omega}_{y \lambda}^{(0)}$ and use $\bar{\bm{\Omega}}$ to compute $\lambda_{e,ij}$'s and $\lambda_{g,k \ell}$'s.
\item[2.] Execute Algorithm \ref{alg0} with initial guess $\bar{\bm{\Omega}}$ to obtain the estimate $\hat{\bm \Omega}_{y \lambda}$. 
\item[3.] Quit if using sparse-group lasso, else set ${\bm \Omega}_{y \lambda}^{(\tilde{m})} = \hat{\bm \Omega}_{y \lambda}$ and $\bar{\bm{\Omega}} = {\bm \Omega}_{y \lambda}^{(\tilde{m})}$ to re-compute $\lambda_{e,ij}$'s and $\lambda_{g,k \ell}$'s via the LLA. Set $\tilde{m} \leftarrow \tilde{m}+1$.
\item[4.] Repeat steps 2 and 3 until convergence. The converged $\hat{\bm \Omega}_{y \lambda}$ is the final estimate of the inverse covariance. (For the numerical results shown in Sec.\ \ref{NE}, we terminated after two iterations of steps 2 and 3, similar to \cite{Zou2008, Lam2009}.)
\item[5.] Given the estimate $\hat{\bm{\Omega}}_{y \lambda}$ resulting from the ADMM algorithm, we estimate the graph edge-set as
\begin{equation} \label{edgesel}
  \hat{\cal E} = \Big\{ \{k, \ell \} \, : \, \|\hat{\bm{\Omega}}_{y \lambda}^{(k \ell)} \|_F > 0, 
	   ~ k, \ell \in [p] \Big\} \, .
\end{equation}
\end{itemize} 
A pseudocode for the above ADMM algorithm is given in Algorithm \ref{alg1}. It utilizes Algorithm \ref{alg0} in each LLA step.

For the numerical results presented in the paper, we picked $r_{\max} = 200$, $\bar{\rho} =2$, $\bar{\phi} =10$, $\tau_{abs}=\tau_{rel} =10^{-4}$, and $\tilde{m}_{\max} =1$ for lasso and $=2$ for LSP, SCAD and MCP penalties in Algorithms \ref{alg0} and \ref{alg1}. For the SCAD penalty $a=3.7$ (as in \cite{Lam2009}), for MCP $b=3.7$, and for the log-sum penalty $\epsilon = 0.0001$.

\begin{algorithm} [htbp]
\caption{LLA-based ADMM Algorithm for Optimizing (\ref{eqth2_20})}
\label{alg1}
\algorithmicrequire{\; Data $\{{\bm x}(t)\}_{t=1}^n$, delay $d \ge 0$, $m=d+1$, regularization parameters $\lambda$ and $\alpha$, ADMM penalty parameter $\rho=\bar{\rho}$, tolerances $\tau_{abs}$ and $\tau_{rel}$, variable penalty factor $\bar{\phi}$, maximum number of iterations $r_{\max}$ and $\tilde{m}_{\max}$. For lasso penalty, $\tilde{m}_{\max} =1$. } \\
\algorithmicensure{\;\ estimated inverse covariance $\hat{\bm \Omega}_{y \lambda}$ and edge-set $\hat{\cal E}$}
\begin{algorithmic}[1] 
\STATE Construct ${\bm y}(t)$ for $t=d+1, d+2, \cdots , n$ with sample size $\bar{n} = n-d$, as in (\ref{neweq31}). Calculate sample covariance $\hat{\bm{\Sigma}}_y = \frac{1}{\bar{n}} \sum_{t=d+1}^{n} {\bm y}(t) {\bm y}^\top(t)$. 
\STATE Initialize ${\bm \Omega}_{y \lambda}^{(0)} = (\mbox{diag}({\hat{\bm \Sigma}}_y))^{-1}$, and set $\lambda_{e,ij} = \lambda$, $\forall i,j \in [mp]$, and $\lambda_{g,k \ell} = \lambda$, $\forall k, \ell \in [p]$. 
\STATE $\tilde{m}=1$
\WHILE{$\tilde{m} \le \tilde{m}_{\max}$,}
\STATE Execute steps 2-12 of Algorithm \ref{alg0}, resulting in output ${\bm \Omega}_{y \lambda}$. Set ${\bm \Omega}_{y \lambda}^{(\tilde{m})} = \hat{\bm \Omega}_{y \lambda}$.
\IF{LSP/SCAD/MCP}
\STATE Set $\bar{\bm{\Omega}} = {\bm \Omega}_{y \lambda}^{(\tilde{m})}$ and re-compute $\lambda_{e,ij}$'s and $\lambda_{g,k \ell}$'s via the LLA.
\ENDIF
\STATE $\tilde{m} \leftarrow \tilde{m}+1$
\ENDWHILE
\STATE Denote the converged estimate by $\hat{\bm{\Omega}}_{y \lambda}$. For $k \ne \ell$, if $\|\hat{\bm{\Omega}}_{y \lambda}^{(k \ell)}\|_F > 0$, assign edge $\{ k, \ell\} \in \hat{\cal E}$, else $\{ k, \ell\} \not\in \hat{\cal E}$.  
\end{algorithmic}
\end{algorithm}

\subsection{Computational Complexity}  \label{CC}
Each iteration of the ADMM algorithm \ref{alg0} has three updates (a), (b) and (c), implemented in steps 5, 6 and 7, respectively, of Algorithm \ref{alg0}. Step 5 requires an eigen-decomposition of an $(mp)\times(mp)$ matrix, resulting in ${\cal O}((mp)^3) =  {\cal O}(((d+1)p)^3)   
$ computational complexity. In step 6 of Algorithm \ref{alg0} one first performs elementwise soft thresholding of an $(mp)\times(mp)$ matrix (${\cal O}((mp)^2)$ complexity), followed by group soft thresholding of $p^2$ $m \times m$ submatrices (${\cal O}((mp)^2)$ complexity). Step 7 involves additions/subtractions among three $(mp)\times(mp)$ matrices, resulting in  ${\cal O}((mp)^2)$ complexity. Step 8 of Algorithm \ref{alg0} requires computation of the Frobenius norms of three $(mp)\times(mp)$ matrices, resulting in  ${\cal O}((mp)^2)$ complexity. The computational complexity in step 9 is ${\cal O}((mp)^2)$ as it involves scaling of an $(mp)\times(mp)$ matrix. Combining the results of steps 5-9, the computational complexity in each iteration of the ADMM algorithm is ${\cal O}(((d+1)p)^3)$. The sample covariance is computed just once, with a computational complexity of ${\cal O}(n(mp)^2)$. The number of iterations to convergence is problem dependent. If the number of iterations to convergence is $N_c$, the computational complexity would be ${\cal O}(n((d+1)p)^2+N_c ((d+1)p)^3)$.

We now turn to memory requirements in each iteration of the ADMM algorithm \ref{alg0}. The data $\{{\bm x}(t)\}_{t=1}^n$ needs storage for $np$ real numbers while ${\bm y}(t)$ requires $(n-d)(d+1)p$ real numbers. We need eleven $(mp) \times (mp)$ matrices ($m=d+1$), $\hat{\bm{\Sigma}}_y$, $\bm{\Omega}_{y \lambda}^{(r)}$, $\bm{\Omega}_{y \lambda}^{(r+1)}$, ${\bm U}^{(r)}$, ${\bm U}^{(r+1)}$, ${\bm V}^{(r)}$, ${\bm V}^{(r+1)}$, ${\bm P}$, $\hat{\bm \Sigma}_y - \rho^{(r)} \left( {\bm V}^{(r)} - {\bm U}^{(r)} \right)$, ${\bm \Omega}_{y \lambda}^{(r+1)} - {\bm V}^{(r+1)}$ and ${\bm V}^{(r+1)} - {\bm V}^{(r)}$, each requiring storage of $((d+1)p)^2$ reals. Ignoring storage of $m \times m$ matrices ${\bm A}^{(k \ell)}$'s which can be reused for different $k, \ell$ values, the memory needs are ${\cal O}(n(d+2)p + 11 ((d+1)p)^2)$.

\subsection{Model Selection} \label{modsel} 
For model selection (i.e., regularization parameter selection), we follow the BIC (Bayesian information criterion) as discussed in \cite[Sec.\ IV-A]{Tugnait2025b} (also \cite{Tugnait21a}) even though it applies only to i.i.d.\ data. Let $\hat{\bm{\Omega}}_{y \lambda}$ and $\hat{\bar{\cal E}}$ ($= \{ \{i,j\} \, : \, |[\hat{\bm{\Omega}}_{y \lambda}]_{ij} | > 0, \; i \ne j \in [mp]\}$) denote the estimated inverse covariance matrix of $\{{\bm y}(t)\}$ and estimated ``enlarged'' (element-level) edge-set, and let $| \hat{\bar{\cal E}} |$ denote the cardinality (\# of nonzero elements) of $\hat{\bar{\cal E}}$.  Noting that $\hat{\bm{\Omega}}_{y \lambda}$ is symmetric with nonzero diagonal elements, the number of free nonzero elements of $\hat{\bm{\Omega}}_{y \lambda}$ equal $\frac{1}{2}| \hat{\bar{\cal E}} |$+$p m $. The (pseudo-)BIC is then given by
\begin{equation}  \label{eqbic}
    \mbox{BIC}(\lambda , \alpha) = {\rm tr} ( \hat{\bm{\Sigma}}_y \hat{\bm{\Omega}}_{y \lambda}  ) 
		   - \ln(|\hat{\bm{\Omega}}_{y \lambda}|) 
		     +  \frac{\ln (\bar{n})}{\bar{n}} \, \Big( \frac{1}{2} | \hat{\bar{\cal E}} | \Big) 
\end{equation}
based on the optimized negative pseudo log-likelihood $- \ln  f_{{\bm Y}}({\bm Y}) \propto  \frac{\bar{n}}{2} \big( {\rm tr} ( \hat{\bm{\Sigma}}_y  \hat{\bm{\Omega}}_{y \lambda}  ) - \ln | \hat{\bm{\Omega}}_{y \lambda} | \big)$, where ${\bm Y} = \{ {\bm y}(t) \}_{t=d+1}^{n}$ and $f_{{\bm Y}}({\bm Y})$ is the joint pseudo probability density function (pdf) of ${\bm Y}$ under the assumption that $\{ {\bm y}(t) \}_{t=d+1}^{n}$ is i.i.d.  We do not know the exact pdf of ${\bm Y}$  as $\{ {\bm y}(t) \}_{t=d+1}^{n}$ is a correlated sequence. There are approaches in the literature \cite{Zhang2023} where one uses information criteria mimicking true BIC ($= - 2 \ln  f_{{\bm Y}}({\bm Y}) + \ln (\bar{n}) (\# \mbox{ of free nonzero parameters}))$ but replacing the true likelihood with a suitable loss function that is used for parameter estimation. We have chosen to use a pseudo pdf under the i.i.d.\ assumption. It is a heuristic that seems to work well in our numerical results presented later. 

Our objective is to select the pair $(\lambda,\alpha)$ to minimize BIC given by (\ref{eqbic}). To reduce computational burden, as a heuristic, instead of searching over a two-dimensional space $(\lambda,\alpha)$, we first search over a grid of $\lambda$ values with fixed $\alpha =0.05 \, (= \alpha_0)$, (somewhat arbitrary, $\alpha=0$ will make it group lasso). Then with $\lambda$ fixed at the selected value, we search over a grid of $\alpha$ values over $[0,1]$. 

We search over $\lambda$ values in the range $[\lambda_\ell , \lambda_u]$ selected via the following heuristic. For $\alpha = \alpha_0$, we first find the smallest $\lambda$, labeled $\lambda_{sm}$, for which we get a no-edge model (i.e., $| \hat{\cal E} | =0$). To this end, searching over a grid of $\lambda$ values, we find the smallest $\lambda$ for which the corresponding $| \hat{\cal E} | \approx \tau_{th} \approx 0$, i.e, for which we get close to a no-edge model; we take this value of $\lambda$ as $\lambda_{sm}$. We picked  $\tau_{th} = 0.005 \, p(p-1)/2 ~$ ($0.5\%$ of possible $p(p-1)/2$ edges are connected). Then we set $\lambda_u = \lambda_{sm}/2$ and $\lambda_\ell = \lambda_u/10$. The given choice of $\lambda_u$ precludes ``extremely'' sparse models while that of $\lambda_\ell$ precludes ``very'' dense models. We search over a grid of $\lambda$ values in the range $[\lambda_\ell , \lambda_u]$ to minimize BIC.

For the numerical results presented in the paper for synthetic data, we picked eight logarithmically spaced values in the interval $[\lambda_\ell , \lambda_u]$ for $\lambda$ selection and the grid $\{0.05, 0.1, 0.3, 0.5, 0.7, 0.9\}$ for $\alpha$ selection. For real data, we picked sixteen logarithmically spaced values in the interval $[\lambda_\ell , \lambda_u]$, with $\alpha$ grid as for synthetic data.

\subsubsection{Extended BIC}  In the high-dimensional case where both the sample size $n$ and the graph nodes $p$ grow with $n$, BIC for Gaussian graphical models may not be consistent and additional penalty may be warranted \cite{Foygel2010} (see also \cite{Zhang2023} for other cases, such as regression problems). In \cite{Foygel2010} an extended BIC (eBIC) is proposed for Gaussian graphical models with $p$  nodes, i.i.d.\ data $\{{\bm x}(t) \}_{t=1}^n =: {\bm X}$, joint pdf $f_{{\bm X}}({\bm X})$ and edge-set ${\cal E}$, as 
\begin{equation}  \label{eqebic1}
   \mbox{eBIC}_{\bar{\gamma}}({\cal E}) = - 2 \ln  f_{{\bm X}}({\bm X}) + \frac{|{\cal E}|}{2} \Big(\ln (n) 
	  + 4 \bar{\gamma} \ln (p) \Big)
\end{equation}
where $\bar{\gamma} \in [0,1]$, and $|{\cal E}|$ counts all edges in the graph with $\{i,j\}$ and $\{j,i\}$, $i \ne j$, counted as distinct edges (hence division by 2). In (\ref{eqebic1}) there is additional penalty for the number of nodes $p$, and $\bar{\gamma}$ is a user chosen parameter. If $\bar{\gamma}=0$, one has the classical BIC. Applying this concept of including the additional penalty, we propose an extended (pseudo-)BIC based on (\ref{eqbic}) as
\begin{align}  
    \mbox{eBIC}(\lambda , \alpha) = & {\rm tr} ( \hat{\bm{\Sigma}}_y \hat{\bm{\Omega}}_{y \lambda}  ) 
		   - \ln(|\hat{\bm{\Omega}}_{y \lambda}|) \nonumber \\
		   & \quad  +  \frac{\ln (\bar{n}) + 4 \bar{\gamma} \ln (mp)}{\bar{n}} \, \Big( \frac{1}{2} | \hat{\bar{\cal E}} | \Big) 
				     \label{eqebic}
\end{align}
where $\bar{\gamma} \in [0,1]$. Selection of $(\lambda,\alpha)$ to minimize $\mbox{eBIC}(\lambda , \alpha)$ is performed exactly as for $\mbox{BIC}(\lambda , \alpha)$.

\section{Theoretical Analysis}  \label{TA}
Here we analyze the properties of $\hat{\bm{\Omega}}_{y \lambda} = \arg\min_{\bm{\Omega}_y  \succ {\bm 0}}  \bar{\cal L}(\bm{\Omega}_y)$. Since the SCAD, MCP and log-sum penalties are non-convex, the objective function is  non-convex and in general, any optimization of the objective function will yield only a stationary point or a local minimum. The proofs of Theorems 1-3 in this section closely follow the corresponding results in \cite[Sec.\ V]{Tugnait2025b} dealing with multi-attribute graph estimation from i.i.d.\ data under lasso, SCAD and log-sum penalties; \cite{Tugnait2025b} does not discuss MCP. The proofs of all theoretical results are given in the two appendices. In this section we explicitly allow $p$ and $\lambda$ to be a functions of sample size $n$, denoted as $p_n$ and $\lambda_n$, respectively. 

Let ${\bm{\Sigma}}_y^\diamond$ and ${\bm{\Omega}}_y^\diamond$ denote the true value of the covariance ${\bm{\Sigma}}_y = E\{ {\bm y}(t) {\bm y}^\top(t) \}$ and its inverse ${\bm{\Omega}}_y ={\bm{\Sigma}}_y^{-1}$, respectively. In Lemma 2, proved in Appendix \ref{append1}, we bound the sample covariance estimation error $\hat{\bm{\Sigma}}_y - \bm{\Sigma}_{y}^\diamond$  with probability  $> 1- \frac{1}{p_n^{\tau -2}}$, $\tau >2$. The error bound depends upon the sample size and the decay parameter $\gamma$ in assumption (A1). 
Let $\hat{\bm{\Omega}}_{y \lambda} = \arg\min_{\bm{\Omega}_y  \succ {\bm 0}}  \bar{\cal L}(\bm{\Omega}_y)$. Using Lemma 2, and assumptions (A2) and (A3) stated later, in Theorem 1 we establish an upperbound on the Frobenius norm of the error $\hat{\bm{\Omega}}_{y \lambda} - {\bm{\Omega}}_y^\diamond$. The proof of Theorem 1 follows the method of proof of \cite[Theorem 1]{Tugnait2025b}, which, in turn, for the most part, follows the method of proof of \cite[Theorem 1]{Rothman2008} pertaining to lasso penalty. 

Let $\bm{\Delta} := {\bm \Omega}_y - {\bm \Omega}_y^\diamond$. The proof of Theorem 1 examines the cost difference $\bar{\cal L}(\bm{\Omega}_y)-\bar{\cal L}({\bm{\Omega}}_y^\diamond) = \sum_{i=1}^4 A_i$ where $A_1 = - \ln (|\bm{\Omega}_y^\diamond + \Delta|) + \ln (|\bm{\Omega}_y^\diamond|) + \mbox{tr} (\bm{\Sigma_y^\diamond  \Delta})$, $A_2 = \mbox{tr} (( \hat{\bm{\Sigma}_y}  - \bm{\Sigma}_y^\diamond ) \bm{\Delta})$, $A_3 = \alpha ({P}_e(\bm{\Omega}_y^\diamond + \Delta) -  {P}_e({\bm \Omega}_y))$ and $A_4 =  (1-\alpha)( {P}_g(\bm{\Omega}_y^\diamond + \Delta) - {P}_g({\bm \Omega}_y))$. Consider the compact set $\Psi = \{ \bm{\Delta} \, :\, \bm{\Delta} = \bm{\Delta}^\top, \; \|\bm{\Delta} \|_F \le B \}$ for some $B > 0$. The proof of Theorem 1 establishes that for some ${\bm \Delta} \in \Psi$ such that $\|{\bm \Delta}\|_F = B$, $\bar{\cal L}(\bm{\Omega}_y^\diamond + \Delta)-\bar{\cal L}({\bm{\Omega}}_y^\diamond) >0$. But the cost difference is zero when ${\bm \Delta} = {\bm 0}$. Hence, there exists a local minimum of $\bar{\cal L}(\bm{\Omega}_y)$ over the set $\Psi$ with Frobenius error norm bounded by $B$. In the proof, Lemma 2 is used to upperbound $A_2$ only. Assumptions (A2) and (A3) are used to upper/lower bound entities in $A_i$, $i \in [4]$, where the sparsity of $\bm{\Omega}_y^\diamond$ stated in assumption (A2) plays a critical role. For the lasso penalty, $\bar{\cal L}(\bm{\Omega}_y)$ is a convex function of $\bm{\Omega}_y$, therefore, the local minimizer $\hat{\bm{\Omega}}_{y \lambda}$ of Theorem 1 is a global  minimizer.

Lemma 3 characterizes the local convexity of $\bar{\cal L}(\bm{\Omega}_y)$ for non-convex penalties. Theorem 1 and Lemma 3 are then used in Theorem 2 to establish conditions under which the local minimizer $\hat{\bm{\Omega}}_{y \lambda}$ of Theorem 1 is unique. Theorem 3 addresses graph recovery using the estimator $\hat{\bm{\Omega}}_{y \lambda}$ of Theorem 1.

We now turn to the statement of Lemma 2. \\
{\it Lemma 2}. Under assumption (A1), the sample covariance $\hat{\bm{\Sigma}}_y$ satisfies the tail bound
\begin{equation}  \label{naeq58bx}
   P \left(  \max_{k,l} \Big| [ \hat{\bm{\Sigma}}_y - \bm{\Sigma}_{y}^\diamond ]_{kl} \Big|
	    > \delta_{*n} \right) \le \frac{1}{p_n^{\tau -2}} 
\end{equation}
for $\tau > 2$, under the following conditions on $\delta_{*n} > 0$ and sample size $\bar{n}=n-d \;$:
\begin{itemize}
\item[(a)] For $\gamma > \frac{3}{4}~$, 
\begin{equation}  \label{naeq58bx2}
   \delta_{*n} = c_a^2(p_n) \, \sqrt{ \ln(2 m^2 p_n^\tau ) / ( \bar{n} C_u ) } \, ,
\end{equation}
and sample size satisfies $\bar{n} \ge N_{\delta_*, \gamma}$, 
\begin{align*}
 \hspace*{-0.1in}  N_{\delta_*, \gamma} \ge & \left\{ \begin{array}{ll}
	     N_1 , & \gamma \in (1,\infty) \\
			 \arg \min \{ \bar{n} \, : \, \bar{n}/\ln^4(\bar{n}) \ge N_1 \}, & \gamma =1  \\
			\arg \min \{ \bar{n} \, : \, \bar{n}^{4 \gamma -3} \ge N_1 \}, & \gamma \in (\frac{3}{4},1)
			\end{array} \right.
\end{align*}
where $C_u \in (0, \infty)$ is a constant that depends only on $\gamma$ and 
\begin{equation}  \label{naeq58bx3}
    N_1 =   \ln(2 m^2 p_n^\tau )/ C_u  \, .
\end{equation}

\item[(b)] For $\gamma = \frac{3}{4}~$, 
\begin{equation}  \label{naeq58bx2}
   \delta_{*n} = c_a^2(p_n) \, \sqrt{ \ln(2 m^2 p_n^\tau ) \, \ln (\bar{n}) / ( \bar{n}  C_u ) } \, ,
\end{equation}
and sample size satisfies $\bar{n} \ge  \arg \min \{ \bar{n} \, : \,   \ln(\bar{n}) \, \ge N_1 \}$.

\item[(c)] For $\gamma \in (\frac{1}{2},\frac{3}{4})~$, 
\begin{equation}  \label{naeq58bx2}
   \delta_{*n} = \left\{ \begin{array}{ll}
	    c_a^2(p_n) \frac{\sqrt{ \ln(2 m^2 p_n^\tau ) /C_u} }{  \bar{n}^{2 \gamma -1}   } & \mbox{ if } \;  N_1 \le 1  \\
			c_a^2(p_n) \frac{ \ln(2 m^2 p_n^\tau ) }{ \bar{n}^{2 \gamma -1} C_u }  & \mbox{ if } \;   N_1 > 1 \; \quad \bullet  \end{array}
			  \right. 
\end{equation}
\end{itemize}
See also Remark 10 in Appendix \ref{append1} regarding the universal constant $C_u$.

Further assume
\begin{itemize}
\setlength{\itemindent}{0.1in}
\item[(A2)] Let $\bm{\Sigma}_{y}^\diamond = E\{ {\bm y}(t) {\bm y}^\top (t) \} \succ {\bm 0}$ denote the true covariance of ${\bm y}(t)$. Define ${\cal E}_{y}^\diamond = \{ \{i,j\} ~:~ ({\bm \Omega}_{y}^\diamond)^{(ij)} \ne {\bm 0}, ~i\ne j \}$ where $\bm{\Omega}_{y}^\diamond = (\bm{\Sigma}_{y}^\diamond)^{-1}$. Assume that card$({\cal E}_{y}^\diamond) =|{\cal E}_{y}^\diamond| \le s_{n}^\diamond$.

\item[(A3)] The minimum and maximum eigenvalues of $\bm{\Sigma}_{y}^\diamond$ satisfy 
\[
     0 < \beta_{\min} \le \phi_{\min}(\bm{\Sigma}_{y}^\diamond) \le \phi_{\max}(\bm{\Sigma}_{y}^\diamond) 
		      \le \beta_{\max} < \infty \, .
\]
Here $\beta_{\min}$ and $\beta_{\max}$ are not functions of $n$.
\end{itemize}
Assumption (A2) is more of a definition in that it specifies that the edge-set determined by ${\bm \Omega}_{y}^\diamond$ has  number of edges bounded by $s_n$. This bound influences the convergence rate in Theorem 1 as discussed later in Remark  6. Assumption (A2) is a sparsity condition. Assumption (A3) bounds the eigenvalues of the true covariance $\bm{\Sigma}_{y}^\diamond$ of $\{ {\bm y}(t) \}$ from below as well as above. Since ${\bm \Omega}_{y}^\diamond = (\bm{\Sigma}_{y}^\diamond)^{-1}$, the eigenvalues of ${\bm \Omega}_{y}^\diamond$ are also bounded from below and above. Assumptions (A2) and (A3) are technical conditions needed to prove Theorem 1 using the proof technique of \cite[Theorem 1]{Rothman2008}. These conditions are mild.

{\it Remark 4}. Assumption (A3) is stated in terms of the covariance $\bm{\Sigma}_{y}^\diamond$ of augmented ${\bm y}(t)$. Now we relate it to the properties of $\{ {\bm x}(t) \}$. Define
\begin{equation}  \label{neweq10070}
   \tilde{\bm y}(t) := [{\bm x}^\top(t) \; {\bm x}^\top(t-1) \; \cdots \; {\bm x}^\top(t-d)]^\top \, .
\end{equation}
Then $\tilde{\bm y}(t) = {\bm P} {\bm y}(t)$ for some permutation matrix ${\bm P}$ (${\bm P} {\bm P}^\top = {\bm P}^\top {\bm P} = {\bm I}_{mp}$, $m=d+1$) and $\bm{\Sigma}_{\tilde{y}}^\diamond = E\{ \tilde{\bm y}(t) \tilde{\bm y}^\top(t)\} = {\bm P} \bm{\Sigma}_{y}^\diamond {\bm P}^\top$. By \cite[Prop.\ 2.3]{Basu15}, for ARMA models discussed in Sec.\ \ref{AR}, for any $d \ge 0$ we have
\begin{align}  
  \min_f \, & \phi_{\min}({\bm S}_x(f)) \le  \phi_{\min}(\bm{\Sigma}_{\tilde{y}}^\diamond) 
	  = \phi_{\min}(\bm{\Sigma}_{y}^\diamond)  \nonumber \\
	& \le \phi_{\max}(\bm{\Sigma}_{\tilde{y}}^\diamond) = \phi_{\max}(\bm{\Sigma}_{y}^\diamond)
	  \le \max_f \phi_{\max}({\bm S}_x(f))   \label{neweq10072}
\end{align}
where we have used the fact that ${\bm P}$ is orthogonal, hence minimum and maximum eigenvalues of $\bm{\Sigma}_{\tilde{y}}^\diamond$ and $\bm{\Sigma}_{y}^\diamond$ are identical. One can take $\beta_{\min} = \min_f \phi_{\min}({\bm S}_x(f))$ and $\beta_{\max} = \max_f \phi_{\max}({\bm S}_x(f))$ for all values of $d$ for ARMA models with $0 < \beta_{\min} \le \beta_{\max} < \infty$. This is also true for $\{ {\bm x}(t) \}$ satisfying assumption (A1) with $\gamma > 1$ \cite[Sec.\ 2]{Basu15} so long as ${\bm S}_x(f) \succ {\bm 0}$.  Since $\beta_{\min} >0$, we also have  $\bm{\Sigma}_{y}^\diamond  \succ {\bm 0}$ for every $d \ge 0$.
 $\quad \Box$

{\it Remark 5}. The results stated in Remark 4 for ARMA and related linear processes do not apply to ARFIMA models discussed in Sec.\ \ref{ARF}. By (\ref{neweq10037}), $\lim_{f \rightarrow 0} {\bm S}_x(f) = \infty$. By \cite[Cor.\ 4.5.11]{Horn_Johnson} of the Ostrowski's theorem \cite[Cor.\ 4.5.9]{Horn_Johnson}, for the ARFIMA process $\{ {\bm x}(t) \}$ discussed in Sec.\ \ref{ARF}, we have
\begin{align}  
 & |1-e^{-\iota 2 \pi f}|^{-2 \, \underbar{d}} ~ \phi_{\min}({\bm S}_u(f)) \le 
      \phi_{\min}({\bm S}_x(f))  \nonumber \\
	& \;\;  \le \phi_{\max}({\bm S}_x(f)) 
			  \le |1-e^{-\iota 2 \pi f}|^{-2 \, \bar{d}} ~ \phi_{\max}({\bm S}_u(f)) \label{neweq10075}
\end{align}
where $\underbar{d} = \min\{d_i, \, i \in [p]\} \in (0,0.5)$ and $\bar{d} = \max\{d_i, \, i \in [p]\} \in (0,0.5)$.
Since $\min_f \,1/ |1-e^{-\iota 2 \pi f}|^{2 \, \underbar{d}} = 1/2^{2 \, \underbar{d}}$, we have
\begin{align}  
 \min_f \, & \phi_{\min}({\bm S}_x(f)) \ge \big( \min_f \, \phi_{\min}({\bm S}_u(f) \big) /4^{\underbar{d}}
      > 0
     \label{neweq10077}
\end{align}
and since $\max_f \,1/ |1-e^{-\iota 2 \pi f}|^{2 \, \underbar{d}} = \infty$, we have
\begin{align}  
 \max_f \, & \phi_{\max}({\bm S}_x(f)) \le \infty \, .
     \label{neweq10078}
\end{align} 
Thus, we may take $\beta_{\min}$ to be the lower bound in (\ref{neweq10077}) applicable for any $d \ge 0$ whereas the upper bound in  (\ref{neweq10078}) is trivial and  of no use. Qualitatively, the ARMA part of ARFIMA captures the short-range dependence (''smaller'' lags in the auto-correlation function) while the fractional integration part captures the long-range dependence (much larger lags in the auto-correlation function, as $\tau \rightarrow \infty$) \cite{Hosking1981, Kechagias2015, Pipiras}). As noted in Remark 3, the conditional independence structure of the ARFIMA process $\{{\bm x}(t)\}$ coincides with that of the corresponding ARMA process $\{{\bm u}(t)\}$. By Remark 1, we would like $d$ to be ``large,'' but not too large so that only short range dependence is captured. For any finite $d$ (maximum delay in delay embedding), $\phi_{\max}(\bm{\Sigma}_{y}^\diamond)$ will be finite. Consider a simple example with ${\bm u}(t) = \tilde{\bm B}_0 {\bm e}(t)$, $q=0$ and $r=0$ in (\ref{neweq10030}) and let ${\bm \Sigma}_u = \tilde{\bm B}_0  \tilde{\bm B}^\top_0  \succ {\bm 0}$. Let $\tilde{\bm B}_0$ be such that $[{\bm \Sigma}_u]_{ii} =1$ for $i \in [p]$.  Suppose $d_i = \tilde{d}$, $i \in [p]$ in (\ref{neweq10034}). Then ${\bm R}_x(\tau) = r(\tau) \otimes {\bm \Sigma}_u$ where $r(\tau)$ is the inverse discrete-time Fourier transform of $|1-e^{-\iota 2 \pi f}|^{-2 \, \tilde{d}}$, i.e., 
\[
        r(\tau) = \int_{-1/2}^{1/2} \, 
				          |1-e^{-\iota 2 \pi f}|^{-2  \, \tilde{d}} \, e^{\iota 2 \pi f \tau} \, df \, ,
\] 
$\tau \in {\cal Z}$. Then we have
\begin{align}  
 \bm{\Sigma}_{\tilde{y}}^\diamond & = \bm{\Sigma}_r \otimes {\bm \Sigma}_u \, , \label{neweq10080} \\
 \bm{\Sigma}_r & = \begin{bmatrix}  r(0) & r(1) & \cdots & r(d) \\
        r(-1) & r(0) & \cdots & r(d-1) \\
					\vdots & \vdots & \ddots & \vdots \\
					r(-d) & r(-d+1) & \cdots & r(0) \end{bmatrix} \, , 
     \label{neweq10082} \\
	\phi_{\max}(\bm{\Sigma}_{y}^\diamond) & = \phi_{\max}(\bm{\Sigma}_{\tilde{y}}^\diamond) =
	  \phi_{\max}(\bm{\Sigma}_r) ~ \phi_{\max}(\bm{\Sigma}_u )\, . \label{neweq10083}
\end{align}
By \cite[pp.\ 37-38]{Pipiras} (note that $[{\bm \Sigma}_u]_{ii} =1$ for $i \in [p]$)
\begin{align}  
 r(\tau) &= \alpha_\tau r(0) \, , \quad r(0) = \frac{\Gamma(1-2 \tilde{d})}
            {(\Gamma(1- \tilde{d}))^2} \, ,
     \label{neweq10084} \\
	\alpha_\tau & = \prod_{\ell =1}^\tau \frac{\ell -1 + \tilde{d}}{\ell - \tilde{d}} \, .  \label{neweq10085}
\end{align}
For large $\tau$, $r(\tau) = {\cal O}(\tau^{2 \tilde{d} -1})$ \cite[p.\ 37]{Pipiras}.
For $\tilde{d} = 0.01, 0.35, 0.49$, we obtain $\phi_{\max}(\bm{\Sigma}_r) = 1.03, 5.04, 93.90$, respectively for $d=5$, $\phi_{\max}(\bm{\Sigma}_r) = 1.04, 7.68, 170.03$, respectively for $d=10$, and $\phi_{\max}(\bm{\Sigma}_r) = 1.06, 12.06, 320.39$, respectively for $d=20$.  Thus, with $\phi_{\max}(\bm{\Sigma}_u )$ uniformly bounded in delay $d$, $\phi_{\max}(\bm{\Sigma}_{y}^\diamond)$ is finite but increases with increasing $d$ (as well as increasing $\tilde{d}$). For $\tilde{d} = 0.01, 0.35, 0.49$, we obtain $\phi_{\min}(\bm{\Sigma}_r) = 0.987, 0.629, 0.523$, respectively for $d=5$, $\phi_{\min}(\bm{\Sigma}_r) = 0.986, 0.620, 0.512$, respectively for $d=10$, and $\phi_{\min}(\bm{\Sigma}_r) = 0.986, 0.617, 0.508$, respectively for $d=20$.  Note that $1/4^{\tilde{d}} = 0.986, 0.616, 0.507$ for $\tilde{d} = 0.01, 0.35, 0.49$, respectively. Thus, $\phi_{\min}(\bm{\Sigma}_r)$ values for various delays agree with the lower bound in (\ref{neweq10077}) quite well.
 $\quad \Box$

Let $\hat{\bm{\Omega}}_{y \lambda} = \arg\min_{\bm{\Omega}_y  \succ {\bm 0}}  \bar{\cal L}(\bm{\Omega}_y)$.
Theorem 1  establishes local consistency of $\hat{\bm{\Omega}}_{y \lambda}$. Its proof is given in Appendix \ref{append2}. \\
{\it Theorem 1 (Local Consistency)}. For $\tau > 2$, $m=d+1$, $\bar{n}=n-d$, and $\delta_{*n}$, $N_1$ and $N_{\delta_*, \gamma}$ as in Lemma 2, define
\begin{align}  
    R = & 8(1+m) / \beta_{\min}^2 \, , \label{neq15ab0} \\
    r_n = & \delta_{*n} \sqrt{ m p_n+ m^2 s_{n}^\diamond}  = o(1)\, , \label{neq15ab1} \\
		N_2 = & \arg \min \left\{ \bar{n} \, : \, r_n \le 
		    \frac{ 0.1}{ R \beta_{\min} } \right\} \, , \label{neq15ab3} \\
    N_3 = &  \arg \min \Big\{\bar{n} \, : \, r_n \le \frac{\epsilon}{R} \Big\} \, ,  \label{neq15ab4}  \\
		N_4 = &  \arg \min \Big\{\bar{n} \, : \, \lambda_n \le 
		       \frac{\min_{(i,j): \, [{\bm \Omega}_{y}^\diamond]_{ij} \ne 0} |[{\bm \Omega}_{y}^\diamond]_{ij}|}{a+1} \Big\}  ,  \label{neq15ab41}  \\
		N_5 = &  \arg \min \Big\{\bar{n} \, : \, \lambda_n \le 
		       \frac{\min_{(i,j): \, [{\bm \Omega}_{y}^\diamond]_{ij} \ne 0} |[{\bm \Omega}_{y}^\diamond]_{ij}|}{b+1} \Big\} 
					\, ,  \label{neq15ab42}  \\
	\lambda_{n \ell} = & 2 \delta_{*n} \, ,  \label{neq15ab5} \\
		\lambda_{n u1} = &  \frac{m+1}{m \sqrt{s_{n}^\diamond}} r_n 
		  = (m+1) \, \sqrt{1+ p_n/(m^2 s_{n}^\diamond)} \, \delta_{*n} \, , \label{neq15ab6} \\
		\lambda_{n u2} = & \max \left( R r_n , \lambda_{n u1} \right) \, .  \label{neq15ab7} 
\end{align}
Under assumptions (A1)-(A3), there exists a local minimizer $\hat{\bm{\Omega}}_{y \lambda}$ of $\bar{\cal L}(\bm{\Omega}_y)$  satisfying
\begin{equation}  \label{neq15}
  \| \hat{\bm{\Omega}}_{y\lambda} - \bm{\Omega}_{y}^\diamond \|_F  \le R  r_n
\end{equation}
with probability greater than $1-1/ p_n^{\tau-2}$ if
\begin{itemize}
\item[(i)] for the lasso penalty $\rho_\lambda(t) = \lambda |t|$,  sample size $\bar{n} >  \max \{ N_{\delta_*, \gamma}, N_2 \}$ and the regularization parameter satisfies $\lambda_{n \ell} \le \lambda_n \le \lambda_{nu1}$,
\item[(ii)] for the SCAD penalty $\rho_\lambda(t)$,  sample size $\bar{n} >  \max \{ N_{\delta_*, \gamma}, N_2, N_4 \}$ and the regularization parameter satisfies $\lambda_n = \lambda_{nu2}$,
\item[(iii)] for the log-sum penalty $\rho_\lambda(t)$, sample size $\bar{n} >  \max \{ N_{\delta_*, \gamma}, N_2, N_3 \}$ and the regularization parameter satisfies $\lambda_{n \ell} \le \lambda_n \le \lambda_{nu1}$,
\item[(iv)] for the MCP penalty $\rho_\lambda(t)$,  sample size $\bar{n} >  \max \{ N_{\delta_*, \gamma}, N_2, N_5 \}$ and the regularization parameter satisfies $\lambda_n = \lambda_{nu2}$.
\end{itemize}
For the lasso penalty, $\hat{\bm{\Omega}}_{y\lambda}$ is a global minimizer whereas for the other three penalties, it is a local minimizer. $\quad \bullet$

{\it Remark 6}. {\it Convergence Rate}. Theorem 1 helps in choosing $m=d+1$ so that for given $ p_n $, $\lim_{n \rightarrow \infty} \\| \hat{\bm{\Omega}}_{y\lambda} - \bm{\Omega}_{y}^\diamond \|_F = 0$. In terms of the rate of convergence,  
  $\| \hat{\bm{\Omega}}_{y\lambda} - \bm{\Omega}_{y}^\diamond \|_F = {\cal O}_P \left( m r_n/\beta_{\min}^2 \right) = {\cal O}_P ( e_n /\beta_{\min}^2)$ where $e_n := m \delta_{*,n} \sqrt{m p_n + m^2 s_{n}^\diamond}$ $~$($mp_n + m^2 s_{n}^\diamond$ is the maximum number of nonzero elements in $\bm{\Omega}_y^\diamond$) and $\beta_{\min}$ is defined in assumption (A3). Therefore, for $\| \hat{\bm{\Omega}}_{y\lambda} - \bm{\Omega}_{y}^\diamond \|_F \rightarrow 0$ as $n \rightarrow \infty$, we must have $e_n \rightarrow 0$, assuming $\beta_{\min}$ is unchanged with $n$. For $\gamma > 3/4$, by Lemma 2(a), $e_n^2 = c_a^4(p_n)m^4 (p_n m^{-1}+ s_{n}^\diamond)  \ln(m^2 p_n^\tau)/n$.  Suppose that for some positive constants $b_1, b_2, \theta_1$ and $ \theta_2$, we have $m = b_1 n^{\theta_1}$ and $p_n + s_{n}^\diamond = b_2 n^{\theta_2}$. Ignoring $c_a(p_n)$, we have $\lim_{n \rightarrow \infty} e_n =0$ if $1-4\theta_1 -\theta_2 > 0$. If $p_n$ and $s_{n}^\diamond$ do not change with $n$ (i.e., $\theta_2 =0$), then $\theta_1 < 1/4$, i.e., $d=m-1 = {\cal O}(n^{1/4})$. If $\theta_2 \le 1/3$ so that $p_n + s_{n}^\diamond = {\cal O}(n^{1/3})$, then $\theta_1 < 1/6$, i.e., $d$ should increase much more slowly with sample size $n$ if the graph size $p_n$ also increases with $n$. 
	$\quad \Box$
	
{\it Remark 7}. {\it Long Range Dependence}. Here we examine Theorem 1 in terms of differences between SRD and LRD time series. By Theorem 1,  the Frobenius error norm $\| \hat{\bm{\Omega}}_{y\lambda} - \bm{\Omega}_{y}^\diamond \|_F = {\cal O}_P (\delta_{* n}/\beta_{\min}^2)$ where $\delta_{* n} \propto 1/n^{0.5}$ (cf.\ Lemma 2(a)) for SRD sequences and for LRD sequences with $\gamma > 0.75$, and $\delta_{* n} \propto 1/n^{2 \gamma -1}$ (cf.\ Lemma 2(c)) for LRD sequences when $\gamma \in (0.5, 0.75)$, with all other parameters remaining the same in $R$ and $r_n$ in (\ref{neq15ab0}) and (\ref{neq15ab1}), respectively.  As discussed in Remark 5, $\beta_{\min}$ decreases with increasing $\tilde{d}$ (recall that $\gamma = 1- \tilde{d}$; see Sec.\ \ref{ARF}). Consider three series with identical CIGs: a vector AR(3) series as in Sec.\ \ref{SRDM1}, and ARFIMA(3,0.15,0) and ARFIMA(3,0.35,0) series as in Sec.\ \ref{LRDM} which are generated via the aforementioned AR(3) series. The AR(3) series is SRD, the ARFIMA(3,0.15,0) series is LRD with $\gamma = 0.85$ and the ARFIMA(3,0.35,0) series is LRD with $\gamma = 0.65$. Let $\beta_{\min} = \beta_0, \beta_{0.15}$ and $\beta_{0.35}$ for the AR(3), ARFIMA(3,0.15,0) and ARFIMA(3,0.35,0) series, respectively. Then $\beta_0 >\beta_{0.15} > \beta_{0.35}$. The Frobenius error norms $e_0, e_{0.15}$ and $e_{0.35}$ for the AR(3), ARFIMA(3,0.15,0) and ARFIMA(3,0.35,0) series, respectively, are ${\cal O}_P (1/(\beta_0 n^{0.5}))$, ${\cal O}_P (1/(\beta_{0.15} n^{0.5}))$ and ${\cal O}_P (1/(\beta_{0.35} n^{0.3}))$, respectively, implying $e_0 < e_{0.15} <e_{0.35}$.  We revisit this discussion later in Sec.\ \ref{LRDM}. 
	$\quad \Box$

We follow the proof technique of \cite[Lemma 6]{Loh2017} in establishing  Lemma 3, as in \cite[Sec.\ V]{Tugnait2025b}. \\
{\it Lemma 3 (Local Convexity)}. The optimization problem
\begin{align}  
    \hat{\bm{\Omega}}_{y\lambda} = & \arg\min_{ \bm{\Omega}_{y} \in {\cal B} }  
		\bar{\cal L}(\bm{\Omega}_{y}) \, , \label{neq4000} \\
		{\cal B} = & \{ \bm{\Omega}_y \,:\, \bm{\Omega}_y \succ {\bm 0} , \; 
		   \| \bm{\Omega}_y \| \le 0.99 \, \theta \} \, , \\
		\theta = & \left\{ \begin{array}{ll}
		   \infty & : \;\; \mbox{Lasso} \\
			  \sqrt{(a-1)/m} & : \;\; \mbox{SCAD} \\
				\sqrt{ \epsilon /(m \lambda_n) } & : \;\; \mbox{log-sum} \\
			  \sqrt{b/m} & : \;\; \mbox{MCP} \, , \end{array} \right. \label{neq4010} 
\end{align}
consists of a strictly convex objective function over a convex constraint set, for all four penalties, where $\lambda_n$ is as defined in Theorem 1. $\quad \bullet$

Lemma 3 and Theorem 1 lead to Theorem 2. \\
{\it Theorem 2}. Assume the conditions of Theorem 1. Then $\hat{\bm{\Omega}}_{y\lambda}$ as defined in Theorem 1 is unique, satisfying $\| \hat{\bm{\Omega}}_{y\lambda} - \bm{\Omega}_{y}^\diamond \|_F  \le R  r_n$ with probability greater than $1-1/p_n^{\tau-2}$ if $R r_n + 1/\beta_{\min} \le 0.99 \, \theta$ where $\theta$ is as in Lemma 3.  $\quad \bullet$

{\it Remark 8}. We see from Theorem 1 that as $n \rightarrow \infty$, $\lambda_n \rightarrow 0$ (since $r_n = o(1)$), therefore, we eventually have ``global'' convexity for log-sum penalty by (\ref{neq4010}) for any $\bm{\Omega}_y^\diamond$. But such is not the case for SCAD or MCP where one may need $a$ (or $b$, respectively) to become large in which case they both would behave more like lasso. $\quad \Box$

We now turn to graph recovery. Define
\begin{align}  
    \hat{\cal E} = & \left\{ \{q,\ell\} \, : \, \|\hat{\bm \Omega}_{y \lambda}^{(q \ell)} \|_F > \gamma_n > 0,
		    q \ne \ell \right\} \, , \label{neq4020} \\
		{\cal E}^\diamond = &   \left\{ \{q,\ell\} \, : \, \| ({\bm \Omega}_{y}^\diamond)^{(q \ell)} \|_F >  0,
		    q \ne \ell \right\} \, , \label{neq4022} \\
       \bar{\sigma}_n = &   R r_n   \, , \label{neq4022} \\
			\nu_t  = &  \min_{\{q,\ell\} \in {\cal E}^\diamond} \,  
			    \|({\bm \Omega}_{y}^\diamond)^{(q \ell)} \|_F  \, , \label{neq4024} \\
			N_6 = & \arg \min \Big\{ n \, : \, \bar{\sigma}_n \le 0.4 \nu_t \Big\} \, , \label{neq4025}
\end{align}
where $R$ and $r_n$ are as in (\ref{neq15ab0}) and (\ref{neq15ab1}), respectively. \\
{\it Theorem 3}. For $\gamma_n = 0.5 \nu_t$ and $n \ge N_6$, $\hat{\cal E} = {\cal E}^\diamond$ with probability greater than $1-1/p_n^{\tau-2}$ under the conditions of Theorem 1.  $\quad \bullet$

\section{Synthetic Data Examples} \label{NE}
We now present numerical results using synthetic data to illustrate the proposed approaches and compare them with several existing approaches. In synthetic data examples the ground truth is known which allows for assessment of the efficacy of various approaches to graph learning.

\subsection{Short-Range Dependence Models} \label{SRDM}
\subsubsection{Graphs With 120 Nodes} \label{SRDM1}
We consider two models for data generation  with $p=120$. \\
{\it MA model}. The time series data $\{ {\bm x}(t) \}$ is generated using a vector moving average (MA) model of order 3 (MA(3)): 
\begin{equation}  \label{VMA1}
    {\bm x}(t) = 0.5 {\bm I}_p {\bm w}(t) + \sum_{i=1}^3 ({\bm B}_i/i) {\bm w}(t-i) \, , \quad {\bm x}(t) \in \mathbb{R}^{p} \, ,
\end{equation} 
where ${\bm w}(t)$ is i.i.d.\ zero-mean Gaussian with precision matrix ${\bm \Omega}$ and ${\bm B}_i$ is block-diagonal with $15 \times 15$ sub-blocks ${\bm B}^{(q)}_i$, $q \in [8]$. The diagonal entries of ${\bm \Omega}$ are set to 0.5, and the off-diagonal entries follow an Erd\"{o}s-R\`{e}nyi (ER) graph with connection probability $p_{er} = 0.001$: if nodes $j$ and $k$ are not connected in the ER graph, we have  $[\bm{\Omega}]_{jk} = 0$, and if they are connected, then $[\bm{\Omega}]_{jk}$ is uniformly distributed over $[-0.4,-0.1] \cup [0.1,0.4]$.  Only 20\% of entries of ${\bm B}^{(q)}_i$'s are nonzero  (randomly picked) and the nonzero elements are independently and uniformly distributed over $[-0.4,-0.2] \cup [0.2,0.4]$. 

{\it AR model}. Here the time series data $\{ {\bm x}(t) \}$ is generated using a vector autoregressive (AR) model of order 3 (VAR(3)): 
\begin{equation}  \label{VAR1}
    {\bm x}(t) = \sum_{i=1}^3 {\bm A}_i {\bm x}(t-i) + {\bm w}(t) \, , \quad {\bm x}(t) \in \mathbb{R}^{p} \, ,
\end{equation} 
where ${\bm w}(t)$ is i.i.d.\ zero-mean Gaussian with precision matrix ${\bm \Omega} = 0.5 {\bm I}_p$ and ${\bm A}_i$ is block-diagonal with $15 \times 15$ sub-blocks ${\bm A}^{(q)}_i$, $q \in [8]$. Only 10\% of entries of ${\bm A}^{(q)}_i$'s are nonzero (randomly picked) and the nonzero elements are independently and uniformly distributed over $[-0.8,-0.3] \cup [0.3,0.8]$. We then check if the VAR(3) model is stable with all eigenvalues of the companion matrix $\le 0.95$ in magnitude; if not, then we scale ${\bm A}_i$'s to fulfill this condition (see \cite[Sec.\ VI.A]{Tugnait2025c}).  
 
For both models, the first 100 samples are discarded to eliminate transients, and  we generate $n$ observations if ${\bm x}(t)$ with $n \in \{256, 1024, 2048\}$. In each run, we calculate the true PSD ${\bm S}_x^\diamond(f)$ for $f \in [0,0.5]$ at intervals of 0.01, and then take $\{i,j \} \in {\cal E}^\diamond$ (the true edge-set) if $g_{ij} > \bar{\tau} g_{\max}$, else $\{i,j \} \not\in {\cal E}^\diamond$, where $g_{\max} = \max_{i,j \in [p]} (1/F) \sum_f |[({\bm S}_x^\diamond(f))^{-1}]_{ij}|$, $g_{ij} = (1/F) \sum_f |[({\bm S}_x^\diamond(f))^{-1}]_{ij}|$, $F$ is number of frequencies points in $[0,0.5]$ at intervals of 0.01, and the threshold $\bar{\tau} = 0.01$ for both MA and AR models. To avoid very ``peaky'' inverse PSDs, if $g_{\max} > 5,000$ we redraw the samples till this condition is satisfied: it is needed for the MA models where we do not control the roots of $\mbox{det}({\cal B}(z)) = 0$ (${\cal B}(z)$ is defined in Sec.\ \ref{AR}). We need $\mbox{det}({\cal B}(z)) \ne 0$ for any $|z|=1$ for ${\bm S}_x^{-1}(f)$ to exist for every $f$. In practical terms, we should have $\mbox{det}({\cal B}(z)) \ne 0$ for $\{ z ~:~ 1-\delta \le |z| \le1+\delta, ~ \delta > 0 \}$.  This set-up leads to connected edges in the true graphs in the range [4.75\%,12.44\%] and [8.53\%,9.94\%] for MA and AR models, respectively, based on 100 runs.

Simulation results based on 100 runs are shown in Table \ref{table1} where the performance measures are $F_1$-score and Hamming distance (between estimated and true edge-sets $\hat{\cal E}$ and ${\cal E}^\diamond$, respectively) for efficacy in edge detection, and timing per run as a surrogate for computational complexity.  The $F_1$-score is defined as $F_1 = 2 \times \mbox{precision} \times \mbox{recall}/(\mbox{precision} + \mbox{recall})$ where $\mbox{precision} = | \hat{\cal E} \cap {\cal E}^\diamond|/ |\hat{\cal E}|$, $\mbox{recall} = |\hat{\cal E} \cap {\cal E}^\diamond|/ |{\cal E}^\diamond|$, and the Hamming distance is between $\hat{\cal E} $ and ${\cal E}^\diamond$, scaled by 0.5 to count only distinct edges. The ideal $F_1$ score is 1 and the ideal Hamming distance is 0. All algorithms were run on a Window 11 Enterprise operating system with processor Intel(R) Core(TM) i7-10700 CPU @2.90 GHz with 32 GB RAM, using MATLAB R2023a. We compare eight approaches:
\begin{itemize}
\item[(i)] ``IID:lasso'': a time-domain approach assuming the data is i.i.d.\ and therefore exploits only the sample covariance $\hat{\bm \Sigma}_x = (1/n) \sum_{t=1}^n {\bm x}(t) {\bm x}^\top(t)$ and the resulting graph is based on the precision matrix ${\bm \Omega}_x$ of $\{ {\bm x}(t) \}$. We minimize $- \ln (|\bm{\Omega}_x|)  + {\rm tr} \left( \hat{\bm \Sigma}_x \bm{\Omega}_x \right)$ w.r.t.\ $\bm{\Omega}_x$ with the lasso penalty, using the ADMM approach of  \cite[Sec.\ 6.5]{Boyd2010}. If the resulting estimate is denoted by $\hat{\bm \Omega}_x$, the estimated edge-set is 
$\hat{\cal E} = \{ \{k, \ell \} \, : \, |[\hat{\bm{\Omega}}_x]_{k \ell}| > 0, ~ k, \ell \in [p] \}$ (cf.\ (\ref{edgesel})).
\item[(ii)] ``IID:log-sum'': a time-domain approach assuming the data is i.i.d.\ and it is as for IID:lasso except the lasso penalty is replaced with the log-sum penalty ($\epsilon = 0.0001$) and we use LLA with ADMM for optimization. 
\item[(iii)] ``DE:lasso'': proposed delay embedding (DE) approach with lasso penalty using ADMM.
\item[(iv)] ``DE:log-sum'': proposed DE approach with log-sum penalty ($\epsilon = 0.0001$) using LLA with ADMM.
\item[(v)] ``DE:SCAD'': proposed DE approach with SCAD penalty ($a = 3.7$) using LLA with ADMM.
\item[(vi)] ``DE:MCP'': proposed DE approach with MCP penalty ($b = 3.7$) using LLA with ADMM.
\item[(vii)] ``FD:lasso'': the frequency-domain (FD) approach of \cite{Tugnait2022a} with lasso penalty using ADMM.
\item[(viii)] ``FD:log-sum'': the FD approach of \cite{Tugnait2025c} as applied to single-attribute models, with log-sum penalty ($\epsilon = 0.0001$) using LLA with ADMM.
\end{itemize} 

\begin{table*}[htbp]
\caption{\small{\it $F_1$ scores, Hamming distances and timings (s) for the synthetic data examples ($p=120$, MA model (\ref{VMA1}), AR model (\ref{VAR1})), averaged over 100 runs (standard deviation $\sigma$ in parentheses). ``DE:lasso and ``DE:log-sum'' are proposed delay embedding approaches with lasso and log-sum penalties, respectively, ``FD:lasso'' and ``FD:log-sum'' are the frequency-domain approaches with lasso \cite{Tugnait2022a} and log-sum \cite{Tugnait2025c} penalties, respectively, and ``IID:lasso'' and ``IID:log-sum'' are time-domain approaches based on i.i.d.\ modeling of data (see also Remark 9).}} \label{table1} 
\begin{center}
\begin{tabular}{c|ccc|ccc}   \hline\hline
 $n$ &  256 &  1024  & 2048  &  256 &  1024  & 2048  \\  \hline\hline
\multicolumn{7}{c}{ $\lambda$'s and $\alpha$'s picked to maximize $F_1$ score } \\
\multicolumn{4}{c|}{MA model: $F_1$ score ($\sigma$) } 
 & \multicolumn{3}{c}{AR model: $F_1$ score ($\sigma$)}\\ \hline
IID:lasso  &  0.326 (0.036)  &  0.482 (0.044)  & 0.555 (0.053)   
  & 0.264 (0.035)  &  0.400 (0.055)  & 0.483 (0.055) \\
IID:log-sum  &  0.331 (0.035)  &  0.511 (0.045)  & 0.593 (0.054)   
  & 0.256 (0.034)  &  0.406 (0.060)  & 0.501 (0.062) \\ \hline
DE:lasso  &  0.492 (0.048)  &  0.730 (0.067)  & 0.791 (0.077)   
  & 0.517 (0.042) & 0.695 (0.032) & 0.763 (0.032) \\ 
DE:log-sum  &  0.587 (0.050)  &  0.861 (0.083)  & 0.903 (0.093)   
  & 0.621 (0.038) & 0.808 (0.028) & 0.873 (0.021) \\
DE:SCAD  & 0.489 (0.046)  & 0.744 (0.069) &  0.811 (0.079) 
  & 0.495 (0.046) & 0.687 (0.025) & 0.762 (0.032) \\
DE:MCP  & 0.498 (0.049)  & 0.751 (0.068) &  0.815 (0.081)   
  & 0.495 (0.037) & 0.694 (0.022) & 0.770 (0.030) \\ \hline
FD:lasso \cite{Tugnait2022a}  &  0.463 (0.044)  &  0.697 (0.059)    & 0.752 (0.070)   
  & 0.447 (0.032) & 0.639 (0.028) & 0.710 (0.023) \\ 
FD:log-sum \cite{Tugnait2025c}  &  0.515 (0.044)  &  0.824 (0.077)  & 0.883 (0.086)   
  & 0.482 (0.030) & 0.716 (0.039) & 0.782 (0.021) \\  \hline\hline
	\multicolumn{4}{c|}{MA model: Hamming distance ($\sigma$) } 
 & \multicolumn{3}{c}{AR model: Hamming distance ($\sigma$)}\\ \hline
IID:lasso   &  905 (160)  &  618 (84)  & 542 (41)   
  & 989 (192)  &  681 (113)  & 592 (86) \\
IID:log-sum   &  910 (198)  &  619 (66)  & 505 (30)   
  & 1027 (223)  &  680 (89)  & 562 (61) \\ \hline
DE:lasso   &  785 (169)  &  387 (49)  & 310 (76)   
  & 646 (108)  &  396 (56)  & 318 (51) \\ 
DE:log-sum   &  549 (78)  &  205 (80)  & 140 (113)   
  & 466 (44)  &  245 (35)  & 172 (26) \\
DE:SCAD  & 810 (142) & 370 (54)  & 284 (75)   
  & 675 (107) & 409 (51)  & 324 (52) \\
DE:MCP  & 740 (107) & 355 (50)  & 280 (84)   
  & 695 (107) & 401 (50) & 313 (48) \\ \hline
FD:lasso   &  715 (108)  &  409 (39)  & 354 (67)   
  & 652 (106) & 417 (41) & 345 (33) \\ 
FD:log-sum   &  716 (145)  &  263 (77)  & 166 (98)   
  & 601 (62) & 312 (28) & 256 (29) \\ \hline\hline
	\multicolumn{4}{c|}{MA model: Timing (s) ($\sigma$) } 
 & \multicolumn{3}{c}{AR model: Timing (s)  ($\sigma$)}\\ \hline
IID:lasso  &  0.069 (0.008)  &  0.068 (0.008)  & 0.078 (0.010)   
  & 0.088 (0.015)  &  0.079 (0.011)  & 0.078 (0.013) \\
IID:log-sum  &  0.297 (0.029)  &  0.231 (0.021)  & 0.225 (0.019)   
  & 0.244 (0.030)  &  0.166 (0.012)  & 0.148 (0.014) \\ \hline
DE:lasso   &  3.65 (0.51)  &  4.46 (0.37)    & 4.86 (0.79)   
  & 4.76 (0.79) & 4.62 (0.76) & 4.69 (0.82) \\ 
DE:log-sum   &  13.23 (1.28)  &  15.12 (1.30)  & 16.69 (2.51)   
  & 12.11 (1.59) & 12.83 (1.38) & 13.11 (1.60) \\ 
DE:SCAD  & 14.48 (2.16)  & 12.58 (0.83) & 14.05 (1.48)   
  & 12.98 (2.11)  & 11.98 (2.24) & 13.27 (2.24) \\
DE:MCP  & 13.28 (1.21)  & 12.34 (1.04) & 14.27 (1.39\texttt{})   
  & 13.48 (2.05)  & 12.21 (2.19) & 13.38 (2.35) \\ \hline
FD:lasso   &  2.99 (1.12)  &  3.39 (0.48)    & 3.84 (0.64)   
  & 4.13 (1.09) & 5.43 (1.14) & 6.35 (1.59) \\ 
FD:log-sum   &  8.89 (1.24) & 8.38 (0.56) & 7.82 (0.56)   
  & 7.79 (0.92) & 9.66 (2.24) & 10.26 (2.77) \\  \hline\hline\hline
	\multicolumn{7}{c}{ DE:log-sum, $\lambda$'s and $\alpha$'s picked to minimize an information criterion } \\
\multicolumn{4}{c|}{MA model: $F_1$ score ($\sigma$) } 
 & \multicolumn{3}{c}{AR model: $F_1$ score ($\sigma$)}\\ \hline
BIC  &  0.535 (0.039) & 0.853 (0.083) & 0.892 (0.092)   
  & 0.643 (0.057) & 0.824 (0.035) & 0.876 (0.019) \\ 
eBIC, $\bar{\gamma}$=0.5  &  0.532 (0.041) & 0.555 (0.047) & 0.819 (0.089)   
  & 0.529 (0.031) & 0.737 (0.059) & 0.838 (0.043) \\ 
eBIC, $\bar{\gamma}$=1  &  0.519 (0.043) & 0.548 (0.026) & 0.638 (0.071)   
  & 0.522 (0.041) & 0.637 (0.074) & 0.787 (0.046) \\ \hline\hline
	\multicolumn{4}{c|}{MA model: Hamming distance ($\sigma$) } 
 & \multicolumn{3}{c}{AR model: Hamming distance ($\sigma$)}\\ \hline
BIC   &  513 (40) & 224 (93) & 163 (116)   
  & 372 (36) & 219 (30) & 169 (21)  \\ 
eBIC, $\bar{\gamma}$=0.5   &  514 (41) & 383 (43) & 240 (86)   
  & 430 (24) & 284 (48) & 197 (39)  \\
eBIC, $\bar{\gamma}$=1   &  520 (46) & 489 (39) & 413 (50)   
  & 434 (28) & 358 (52) & 243 (40)  \\   \hline\hline
	\multicolumn{4}{c|}{MA model: Timing (s) ($\sigma$) } 
 & \multicolumn{3}{c}{AR model: Timing (s) ($\sigma$)}\\ \hline
BIC   &  11.73 (1.23) & 15.79 (1.48) & 15.58 (1.45)   
  & 13.28 (2.14) & 12.07 (1.59) & 13.43 (1.33)  \\  
eBIC, $\bar{\gamma}$=0.5   &  12.32 (1.41) & 15.75 (1.68) & 15.98 (1.52)   
  & 12.58 (2.04) & 12.52 (2.15) & 13.73 (1.43)  \\  
eBIC, $\bar{\gamma}$=1.0   &  13.26 (1.84) & 14.68 (1.74) & 14.54 (2.15)   
  & 12.97 (2.24) & 13.17 (1.89) & 13.29 (1.60)  \\  \hline\hline
\end{tabular} 
\end{center}
\end{table*} 
 
For the DE approaches we need to pick delay $d$ which was $d=3$ for both MA and AR models and for all sample sizes. For the FD approaches we need to pick number of frequency test points $M$ in the interval $(0,0.5)$ which were $M=2,3,4$ for both MA and AR models, for sample sizes $n=256, 1024, 2048$, respectively.  All non-convex penalized approaches were based on LLA, initialized with the lasso results, therefore, the timing for non-convex penalties includes the timing for lasso as well. In approaches (iii)-(viii) we use sparse group penalties (cf.\ (\ref{eqth2_20})) requiring choice of the pair $(\lambda,\alpha)$. In Table \ref{table1}, $\lambda$'s and $\alpha$'s were picked from a grid of values to maximize the $F_1$ score in each Monte Carlo run (ground truth is known in synthetic examples) -- this establishes how well a method will perform if $\lambda$'s and $\alpha$'s are judiciously picked. Instead of searching over a two-dimensional space $(\lambda,\alpha)$, we first search over a grid of $\lambda$ values with fixed $\alpha =0.05 \, (= \alpha_0)$. Then with $\lambda$ fixed at the selected value, we search over a grid of $\alpha$ values over $[0,1]$. We used the grid $\{0.05, 0.1, 0.3, 0.5, 0.7, 0.9\}$ for $\alpha$ selection. We picked ten logarithmically spaced values in the interval $[\underline{\lambda} , \overline{\lambda}]$ for $\lambda$ selection with $(\underline{\lambda} , \overline{\lambda}) = (0.01, 0.5)$ for lasso, SCAD and MCP penalties, and $(\epsilon \underline{\lambda} , \epsilon \overline{\lambda}) = (0.001, 0.1)$ for the log-sum penalty.

In Table \ref{table1} we also show results for log-sum penalty  (DE:log-sum) when $\lambda$'s and $\alpha$'s are picked based on data alone, using the BIC or eBIC method discussed in Sec.\ \ref{modsel} (BIC is based on \cite[Sec.\ IV-A]{Tugnait2025b} and \cite[Sec.\ III-B]{Tugnait21a}, and eBIC is based on \cite{Foygel2010}). For eBIC we picked $\bar{\gamma}$ in (\ref{eqebic1}) as either 0.5 or 1  (recommended values in \cite{Foygel2010}).

It is seen from Table \ref{table1} that log-sum-penalized approaches (FD or DE) significantly outperform lasso based DE and FD approaches, respectively, which in turn, significantly outperform  the i.i.d.\ modeling based methods that ignore time-dependence, yielding higher $F_1$ scores and lower Hamming distances. The performance of the DE approaches is slightly to significantly better than the FD approaches when comparing DE:lasso with FD:lasso, and DE:log-sum with FD:log-sum, with $F_1$ scores and Hamming distances as performance metrics. For instance, for $n=256$, for the proposed DE:log-sum  we have an $F_1$ score of  0.587 and Hamming distance of 549 for the MA model, whereas FD:log-sum yields an $F_1$ score of  0.515 and Hamming distance of 716. The corresponding  $F_1$ score and Hamming distance for the AR model for $n=256$ are 0.621 and 466, respectively, for DE:log-sum, and 0.482 and 601, respectively, for FD:log-sum. It is also seen that the proposed DE:SCAD and DE:MCP approaches do not offer as much performance improvements over DE:lasso as offered by DE:log-sum. Finally, it is also seen from Table \ref{table1} that for $(\lambda,\alpha)$ selection based on data alone, the performance of the eBIC method is worse than that of the BIC method.

{\it Remark 9}. Given the sample covariance $\hat{\bm \Sigma}_x = (1/n) \sum_{t=1}^n {\bm x}(t) {\bm x}^\top(t)$, the IID:lasso approach considered in Table \ref{table1} minimizes
\begin{equation} \label{iidlasso}
   - \ln (|\bm{\Omega}_x|)  + {\rm tr} \left( \hat{\bm \Sigma}_x \bm{\Omega}_x \right) + \sum_{i \ne j} |[\bm{\Omega}_x]_{ij}|
\end{equation}
w.r.t.\ $\bm{\Omega}_x$ using the ADMM approach of \cite[Sec.\ 6.5]{Boyd2010}, which is also the approach in the Algorithm \ref{alg0} with $d=0$ and $\alpha=1$ (requires modifications to step 6 of Algorithm \ref{alg0}). The resulting graph is based on the estimated $\bm{\Omega}_x$. This cost function has been used in \cite{Ravikumar2011, Rothman2008, Friedman2008}, but optimized using different approaches. In \cite{Cai2011} a constrained $\ell_1$-minimization (CLIME) approach is proposed to estimate $\bm{\Omega}_x$ under the assumption of i.i.d.\ data, and the resulting graph is based on the estimated $\bm{\Omega}_x$. A software package called ``huge'' is available in \cite{Zhao2012} for precision matrix based graph estimation where several approaches including graphical lasso of \cite{Friedman2008}, have been implemented for i.i.d.\ data. The CLIME approach \cite{Cai2011} exhibited ``numerical instability'' in the simulation examples considered in \cite{Zhao2012}. In \cite{Shu19} the approaches of \cite{Rothman2008} and \cite{Cai2011} to estimate $\bm{\Omega}_x$ are analyzed and compared via simulations when the data $\{ {\bm x}(t) \}$ is time-dependent. It is concluded in \cite{Shu19} that the performances of the approaches of \cite{Rothman2008} and \cite{Cai2011} are similar in terms of the matrix estimation error norms in estimating $\bm{\Omega}_x$. Note that $\bm{\Omega}_x$ does not yield the underlying time-series graph for dependent data. The lasso penalty in (\ref{iidlasso}) is replaced with non-convex penalties such as SCAD or LSP, in \cite{Lam2009, Loh2017, Tugnait2025b}, and one minimizes
\begin{equation} \label{iidlsp}
   - \ln (|\bm{\Omega}_x|)  + {\rm tr} \left( \hat{\bm \Sigma}_x \bm{\Omega}_x \right) 
	           + \sum_{i \ne j} \rho_\lambda \left(|[\bm{\Omega}_x]_{ij}| \right)
\end{equation}
where $\rho_\lambda(u)$, $u \in \mathcal{R}$, is as in Sec.\ \ref{PNLL1}. 
Note that \cite{Tugnait2025b} considers the multi-attribute case with i.i.d.\ data, and it applies here after setting the number of attributes to one. The IID:log-sum approach considered in Table \ref{table1} is the approach in the Algorithm \ref{alg1} with $d=0$ and $\alpha=1$ (requires modifications to step 6 of Algorithm \ref{alg0}). We note that non-convex penalties are not considered in \cite{Ravikumar2011, Rothman2008, Friedman2008, Cai2011}. $\quad \Box$

\begin{table}[htbp]
\caption{\small{\it MA model (\ref{VMA1}) and AR model (\ref{VAR1}), $p=120$: $F_1$ scores, Hamming distances and timings using DE:log-sum approach, averaged over 100 runs, for varying time-delays $d$.}} \label{table2} 
\begin{center}
\begin{tabular}{cccc}   \hline\hline
 $n$ &  256 &  1024  & 2048  \\  \hline\hline
\multicolumn{4}{c}{MA model (\ref{VMA1}) } \\ \hline
\multicolumn{4}{c}{$F_1$ score ($\sigma$): $\lambda$'s and $\alpha$'s picked to maximize $F_1$ } \\ \hline
$d$=1  &  0.551 (0.051)  &  0.788 (0.071)  & 0.846 (0.081)     \\  
$d$=2   &  0.586 (0.052)  &  0.848 (0.082)  & 0.893 (0.090)     \\ 
$d$=3   &   0.587 (0.050)  &  0.861 (0.083)  & 0.903 (0.093)    \\ 
$d$=4   &  0.589 (0.050)  &  0.863 (0.081)   & 0.903 (0.093)     \\ 
$d$=5   &  0.581 (0.049)  &  0.860 (0.080)   & 0.900 (0.092)     \\ \hline\hline
\multicolumn{4}{c}{Hamming distance ($\sigma$): $\lambda$'s and $\alpha$'s picked to maximize $F_1$ } \\ \hline
$d$=1    &  636 (60)  &  301 (57)  & 219 (85)     \\  
$d$=2   &  596 (79)  &  224 (79)  & 155 (108)     \\ 
$d$=3   &  549 (78)  &  205 (80)  & 140 (113)    \\
$d$=4   &  553 (70)  &  200 (78)  & 139 (112)    \\ 
$d$=5   &  543 (53)  &  202 (77)  & 143 (110)    \\ \hline\hline
\multicolumn{4}{c}{Timing (s) ($\sigma$): $\lambda$'s and $\alpha$'s picked to maximize $F_1$ } \\ \hline
$d$=1    &  4.75 (0.34)  &  4.19 (0.43)  & 4.43 (0.46)     \\  
$d$=2   &  7.94 (0.54)  &  8.13 (0.62)  & 8.57 (0.62)     \\ 
$d$=3   &  13.23 (1.28)  &  15.12 (1.30)  & 16.69 (2.51)     \\
$d$=4   &  21.29 (1.46)  &  23.50 (1.81)  & 24.32 (3.27)    \\ 
$d$=5   &  27.89 (3.32)  &  31.34 (1.75)  & 33.16 (2.85)    \\ \hline\hline
\multicolumn{4}{c}{AR model (\ref{VAR1}) } \\ \hline
\multicolumn{4}{c}{$F_1$ score ($\sigma$): $\lambda$'s and $\alpha$'s picked to maximize $F_1$ } \\ \hline
$d$=1  &  0.497 (0.040)  &  0.666 (0.051)  & 0.741 (0.041)     \\  
$d$=2   &  0.581 (0.031)  &  0.759 (0.034)  & 0.824 (0.023)     \\ 
$d$=3   &   0.621 (0.038) & 0.808 (0.028) & 0.873 (0.021)    \\ 
$d$=4   &  0.631 (0.049)  &  0.817 (0.037)   & 0.877 (0.022)     \\ 
$d$=5   &  0.634 (0.052)  &  0.818 (0.038)   & 0.877 (0.023)     \\ \hline\hline
\multicolumn{4}{c}{Hamming distance ($\sigma$): $\lambda$'s and $\alpha$'s picked to maximize $F_1$ } \\ \hline
$d$=1    &  574 (37)  &  392 (38)  & 320 (34)     \\  
$d$=2   &  502 (31)  &  307 (28)  & 236 (23)     \\ 
$d$=3   &  466 (44)  &  245 (35)  & 172 (26)     \\
$d$=4   &  437 (66)  &  231 (45)  & 164 (26)    \\ 
$d$=5   &  420 (73)  &  223 (43)  & 160 (26)    \\ \hline\hline
\multicolumn{4}{c}{Timing (s) ($\sigma$): $\lambda$'s and $\alpha$'s picked to maximize $F_1$ } \\ \hline
$d$=1    &  3.65 (0.40)  &  3.45 (0.19)  & 3.31 (0.45)     \\  
$d$=2   &  7.13 (0.93)  &  6.97 (0.46)  & 6.89 (0.79)     \\ 
$d$=3   &  12.11 (1.59) & 12.83 (1.38) & 13.11 (1.60)     \\
$d$=4   &  19.48 (3.03)  &  19.44 (2.78)  & 19.87 (2.98)    \\ 
$d$=5   &  27.18 (4.37)  &  28.02 (5.10)  & 28.12 (4.41)    \\ \hline\hline
\end{tabular} 
\end{center}
\end{table}

\subsubsection{Varying Embedding Time-Delays, 120 Nodes} \label{SRDM2}
Table \ref{table2} shows the performance results for the MA model (\ref{VMA1}) and the AR model (\ref{VAR1}) using the proposed DE:log-sum approach with varying embedding time-delay $d \in \{1,2,3,4,5 \}$. The number of unknown parameters of ${\bm \Omega}_y$ being estimated are ${\cal O}(((d+1)p)^2)$. As noted in Remark 1, we are using ${\bm \Omega}_y^{(ij)} = {\bm 0}$, ${\bm y}(t) \in \mathbb{R}^{(d+1)p}$, as surrogate for $[{\bm S}_x^{-1}(f)]_{ij} = 0 \; \forall f \in [0,1]$ with the approximation improving with increasing $d$. The simulation results are shown in Table \ref{table2}. It is seen from Table \ref{table2} that as $d$ increases, the performance metrics the $F_1$ score and Hamming distance improve significantly at first, but then they plateau out with increasing $d$ (compare $d=3$ results with $d=4,5$ results). Note that the number of unknown parameters increase as ${\cal O}((d+1)^2)$, which also accounts for significant increase in computational time with increasing $d$. 

As noted earlier in Remark 2 in Sec.\ \ref{MAMTS}, this observation suggests the heuristic that the user should try several values of $d$ and pick the smallest value, say $d_0$, beyond which (e.g., for $d=d_0+1$) the estimated graph is only marginally different from the estimated graph for $d=d_0$. In practice, since the true graph is unknown, one cannot compute the $F_1$ score or the Hamming distance. For real data, what constitutes marginal difference would be problem and user dependent. We revisit this later in the context of financial time series based real data results presented in Sec.\ \ref{NEreal}.

\begin{table}[htbp]
\caption{\small{\it Varying AR model order: VAR($L$) as in (\ref{VARL}), $L \in \{1,2,3,4 \}$, $p=120$. $F_1$ scores, Hamming distances and timings  using DE:log-sum with $d=3$, averaged over 100 runs.}} \label{table3} 
\begin{center}
\begin{tabular}{cccc}   \hline\hline
 $n$ &  256 &  1024  & 2048  \\  \hline\hline
\multicolumn{4}{c}{$F_1$ score ($\sigma$): $\lambda$'s and $\alpha$'s picked to maximize $F_1$ } \\ \hline
$L$=1  &  0.764 (0.028)  &  0.866 (0.017)  & 0.892 (0.014)     \\  
$L$=2   &  0.676 (0.048)  &  0.823 (0.022)  & 0.852 (0.019)     \\ 
$L$=3   &   0.621 (0.038) & 0.808 (0.028) & 0.873 (0.021)     \\ 
$L$=4   &  0.589 (0.040)  &  0.796 (0.034)  & 0.865 (0.030)     \\ \hline\hline
\multicolumn{4}{c}{Hamming distance ($\sigma$): $\lambda$'s and $\alpha$'s picked to maximize $F_1$ } \\ \hline
$L$=1    &  104 (16)  &  64 (10)  & 56 (10)    \\  
$L$=2   &  286 (51)  &  170 (29)  & 148 (15)    \\ 
$L$=3   &  466 (44)  &  245 (35)  & 172 (26)     \\
$L$=4   &  559 (29)  &  291 (33)  & 198 (36)    \\ \hline\hline
\multicolumn{4}{c}{Timing (s) ($\sigma$): $\lambda$'s and $\alpha$'s picked to maximize $F_1$ } \\ \hline
$L$=1    &  16.09 (2.24)  &  14.92 (2.63)  & 12.86 (3.05)    \\  
$L$=2   &  14.50 (2.58)  &  14.22 (2.68)  & 13.34 (3.06)    \\ 
$L$=3   &  12.11 (1.59) & 12.83 (1.38) & 13.11 (1.60)     \\
$L$=4   &  12.28 (1.53) & 12.57 (0.98) & 12.78 (1.15)    \\ \hline\hline
\end{tabular} 
\end{center}
\end{table}

\subsubsection{Varying Order and Sparsity of AR models, 120 Nodes} \label{SRDM3}

{\it Varying Order}.  In Table \ref{table3} we display some numerical ablation results by varying the AR model order. We use a VAR($L$) model with $L \in \{1,2,3,4 \}$, given by 
\begin{equation}  \label{VARL}
    {\bm x}(t) = \sum_{i=1}^L {\bm A}_i {\bm x}(t-i) + {\bm w}(t) \, , \quad {\bm x}(t) \in \mathbb{R}^{p} \, ,
\end{equation}  
where ${\bm w}(t)$ is i.i.d.\ zero-mean Gaussian with precision matrix ${\bm \Omega} = 0.5 {\bm I}_p$ and ${\bm A}_i$ is block-diagonal with $15 \times 15$ sub-blocks ${\bm A}^{(q)}_i$, $q \in [8]$, as for (\ref{VAR1}), and $p=120$. Only 10\% of entries of ${\bm A}^{(q)}_i$'s are nonzero (randomly picked) and the nonzero elements are independently and uniformly distributed over $[-0.8,-0.3] \cup [0.3,0.8]$. The graph sparsity varies with varying $L$ even though sparsity of ${\bm A}^{(q)}_i$'s remains unchanged. Based on 100 Monte Carlo runs, the number of edges in the resulting true graphs are in the ranges $[2.90\%, 4.47\%]$, $[5.94\%, 8.01\%]$, $[8.53\%, 9.94\%]$ and $[9.43\%, 11.04\%]$ for $L=1$, 2, 3 and 4, respectively. We used the log-sum penalty and $d=3$. The simulation results are shown in Table \ref{table3}.  It is seen that the results are consistent across model orders, both in terms of the $F_1$ scores and the Hamming distances. 

{\it Varying Sparsity, AR model (\ref{VAR1})}. Here we use VAR(3) model as in (\ref{VAR1}) (or in (\ref{VARL}) with $L=3$), and vary the percentage of nonzero entries in ${\bm A}^{(q)}_i$'s. By varying these percentages as 2\%, 5\% and 10\%, we obtain the number of edges in the resulting true graphs in the ranges $[1.26\%, 2.46\%]$, $[4.19\%,6.29\%]$ and $[8.53\%, 9.94\%]$, respectively (based on 100 Monte Carlo runs). The simulation results are shown in Table \ref{table4}. It is seen that the results are consistent across various sparsity levels in terms of the $F_1$ scores. Higher sparsity levels implies more edges in the graph, and the larger Hamming distances for higher sparsity reflects more errors in graph recovery for the same relative errors measured by the $F_1$ score. By Theorem 1, the estimation error norm $\| \hat{\bm{\Omega}}_{y\lambda} - \bm{\Omega}_{y}^\diamond \|_F = {\mathcal O}(\sqrt{mp_n + m^2 s_{n}^\diamond})$ where $s_{n}^\diamond$ is the number of (off-diagonal) edges in the graph and $m=d+1$.

\begin{table}[htbp]
\caption{\small{\it Varying sparsity of AR(3) model (\ref{VAR1}), $p$=120: $F_1$ scores, Hamming distances and timings  using DE:log-sum with $d=3$, averaged over 100 runs.}} \label{table4} 
\begin{center}
\begin{tabular}{cccc}   \hline\hline
 $n$ &  256 &  1024  & 2048  \\  \hline\hline
$\%$ edges & \multicolumn{3}{c}{$F_1$ score ($\sigma$): $\lambda$'s, $\alpha$'s maximize $F_1$ } \\ \hline
$[1.26\%,2.46\%]$  &  0.779 (0.030)  &  0.852 (0.021)  & 0.861 (0.021)     \\  
$[4.19\%,6.29\%]$   &  0.650 (0.052)  &  0.793 (0.018)  & 0.824 (0.015)     \\ 
$[8.53\%, 9.94\%]$   &   0.621 (0.038) & 0.808 (0.028) & 0.873 (0.021)     \\  \hline\hline
$\%$ edges & \multicolumn{3}{c}{Hamming distance ($\sigma$): $\lambda$'s, $\alpha$'s maximize $F_1$ } \\ \hline
$[1.26\%,2.46\%]$    &  53 (11)  &  40 (9)  & 41 (10)    \\  
$[4.19\%,6.29\%]$   &  220 (50)  &  148 (18)  & 142 (15)    \\ 
$[8.53\%, 9.94\%]$    &  466 (44)  &  245 (35)  & 172 (26)     \\ \hline\hline
$\%$ edges & \multicolumn{3}{c}{Timing (s) ($\sigma$): $\lambda$'s, $\alpha$'s maximize $F_1$ } \\ \hline
$[1.26\%,2.46\%]$    &  11.02 (1.92)  &  9.04 (1.46)  & 7.98 (1.43)    \\  
$[4.19\%,6.29\%]$   &  11.68 (1.94)  &  12.40 (2.53)  & 13.34 (1.53)    \\ 
$[8.53\%, 9.94\%]$    &  12.11 (1.59) & 12.83 (1.38) & 13.11 (1.60)     \\ \hline\hline
\end{tabular} 
\end{center}
\end{table}

\begin{table}[htbp]
\caption{\small{\it $F_1$ scores, Hamming distances and timings (s) for AR(3) models, varying number of nodes p=60, 120 or 240,  using DE:log-sum with $d=3$, averaged over 100 runs (standard deviation $\sigma$ in parentheses). }} \label{table5} 
\begin{center}
\begin{tabular}{c|ccc}   \hline\hline
 $n$ &  256 &  1024  & 2048    \\  \hline\hline
\multicolumn{4}{c}{ $F_1$ score ($\sigma$): $\lambda$'s and $\alpha$'s picked to maximize $F_1$ } \\ \hline
p=60  &   0.713 (0.047) & 0.845 (0.029) & 0.880 (0.024) \\ 
p=120  &   0.621 (0.038) & 0.808 (0.028) & 0.873 (0.021) \\ 
p=240  &   0.524 (0.034) & 0.767 (0.030) & 0.856 (0.018) \\ \hline\hline
	\multicolumn{4}{c}{Hamming distance ($\sigma$): $\lambda$'s and $\alpha$'s picked to maximize $F_1$}\\ \hline
p=60   & 101 (18)  &  58 (19)  & 47 (8) \\  
p=120   & 466 (44)  &  245 (35)  & 172 (26) \\  
p=240   & 2622 (67)  &  1388 (106)  & 889 (81) \\ \hline\hline
	\multicolumn{4}{c}{Timing (s) ($\sigma$): $\lambda$'s and $\alpha$'s picked to maximize $F_1$ }\\ \hline
p=60   &  2.90 (0.31) & 2.85 (0.40) & 2.96 (0.41) \\ 
p=120  &  12.11 (1.59) & 12.83 (1.38) & 13.11 (1.60) \\ 
p=240  &  58.93 (6.55) & 58.09 (8.36) & 60.75 (7.56) \\  \hline\hline
\end{tabular} 
\end{center}
\end{table} 

\begin{table}[htbp]
\caption{\small{\it $F_1$ scores, Hamming distances and timings (s) for non-Gaussian AR(3) model (\ref{VAR1}) ($p=120$), averaged over 100 runs (standard deviation $\sigma$ in parentheses). ``DE: lasso and ``DE: log-sum'' are proposed delay embedding approaches with lasso and log-sum penalties, respectively, using $d=3$.}} \label{table6} 
\begin{center}
\begin{tabular}{c|ccc}   \hline\hline
 $n$ &  256 &  1024  & 2048   \\  \hline\hline
\multicolumn{4}{c}{Laplace ${\bm w}(t)$; $\lambda$'s and $\alpha$'s picked to maximize $F_1$ score } \\ \hline
\multicolumn{4}{c}{$F_1$ score ($\sigma$) } \\ \hline
DE:lasso  &  0.515 (0.052) & 0.698 (0.035) & 0.758 (0.038) \\ 
DE:log-sum  &   0.621 (0.046) & 0.811 (0.027) & 0.868 (0.023) \\  \hline\hline
	\multicolumn{4}{c}{Hamming distance ($\sigma$) } \\ \hline
DE:lasso   &   658 (131)  &  392 (60)  & 324 (56) \\ 
DE:log-sum   &  468 (47)  &  243 (34)  & 178 (24) \\ \hline\hline
	\multicolumn{4}{c}{Timing (s) ($\sigma$) } \\ \hline
DE:lasso   &  5.36 (1.03) & 5.38 (1.06) & 5.13 (0.72) \\ 
DE:log-sum   &   13.94 (2.14) & 15.07 (2.46) & 15.14 (2.59) \\  \hline\hline
\multicolumn{4}{c}{Exponential ${\bm w}(t)$; $\lambda$'s and $\alpha$'s picked to maximize $F_1$ score } \\ \hline
\multicolumn{4}{c}{$F_1$ score ($\sigma$) } \\ \hline
DE:lasso  &  0.514 (0.053) & 0.697 (0.033) & 0.760 (0.037) \\ 
DE:log-sum  &   0.619 (0.048) & 0.812 (0.027) & 0.867 (0.023) \\  \hline\hline
	\multicolumn{4}{c}{Hamming distance ($\sigma$) } \\ \hline
DE:lasso   &   657 (125)  &  392 (56)  & 322 (54) \\ 
DE:log-sum   &  466 (52)  &  243 (33)  & 180 (25) \\ \hline\hline
	\multicolumn{4}{c}{Timing (s) ($\sigma$) } \\ \hline
DE:lasso   &  5.57 (0.94) & 5.66 (1.13) & 5.18 (0.75) \\ 
DE:log-sum   &   14.26 (1.77) & 15.67 (1.82) & 14.73 (1.74) \\  \hline\hline
\end{tabular} 
\end{center}
\end{table} 

\subsubsection{AR models, Varying Number Of Nodes} \label{SRDM4}
We now consider AR(3) models for time-dependent data generation with varying number of graph nodes $p \in \{60, 120, 240 \}$. The objective is to empirically study the performance stability/scalability of the proposed solutions with varying model dimensions. The AR(3) model follows (\ref{VAR1}) where ${\bm A}_i \in \mathbb{R}^{p \times p}$, $i \in [3]$, is block-diagonal with 
\begin{itemize}
\item[(i)] six $10 \times 10$ sub-blocks ${\bm A}^{(q)}_i$, $q \in [6]$, when $p=60$,
\item[(ii)] eight $15 \times 15$ sub-blocks ${\bm A}^{(q)}_i$, $q \in [8]$, when $p=120$ (as in Sec.\ \ref{SRDM1}),
\item[(iii)] eight $30 \times 30$ sub-blocks ${\bm A}^{(q)}_i$, $q \in [8]$, when $p=240$.
\end{itemize}
All other details regarding generation of ${\bm \Omega}$, ${\bm A}^{(q)}_i$'s and $\{{\bm x}(t)\}$ are exactly as before in Sec.\ \ref{SRDM1}. For the given choices, the percentage of distinct connected edges in the true graphs do not change much with changing $p$: in 100 Monte Carlo runs, the number of edges in the resulting true graphs are in the ranges $[9.04\%, 12.77\%]$, $[8.53\%, 9.94\%]$ and $[10.22\%, 11.35\%]$ for $p=60$, 120 and 240, respectively. Since the total number of edges in a $p$-node graph would be $p(p-1)/2$, $10\%$ edges result in 123, 714 and 2868 edges in graphs with $p=$60, 120 and 240 nodes, respectively.

We used the log-sum penalty and $d=3$. The simulation results are shown in Table \ref{table5}. Recall that the computational complexity is ${\cal O}((d+1)^3 p^3)$, the number of unknowns in ${\bm \Omega}_y$ is ${\cal O}((d+1)^2 p^2)$ and the estimation error norm $\| \hat{\bm{\Omega}}_{y\lambda} - \bm{\Omega}_{y}^\diamond \|_F = {\mathcal O}(\sqrt{(d+1) p_n + (d+1)^2 s_{n}^\diamond})$ where $s_{n}^\diamond$ is the number of edges in the graph. As $p$ increases, so does $s_{n}^\diamond$ in our models, increasing the estimation error norm. This explains the degradation in the performance metrics with increasing $p$, but the results improve with increasing sample size, achieving comparable $F_1$ scores of 0.880, 0.873 and 0.856 for $p=$60, 120 and 240, respectively, when $n=$2048.

\begin{table}[htbp]
\caption{\small{\it AR(3), ARFIMA(3,0.15,0) and ARFIMA(3,0.35,0) Models, $p=120$: $F_1$ scores, Hamming distances and timings (s), averaged over 100 runs. For fractional integration $d_i = \tilde{d} = 0.15$ or $0.35$ $\forall i \in [120]$. ``DE:lasso and ``DE:log-sum'' are proposed delay embedding approaches with lasso and log-sum penalties, respectively ($d=3$), ``FD:lasso'' and ``FD:log-sum'' are the frequency-domain approaches with lasso \cite{Tugnait2022a} and log-sum \cite{Tugnait2025c} penalties, respectively}} \label{table7} 
\begin{center}
\begin{tabular}{cccc}   \hline\hline
 $n$ &  256 &  1024  & 2048  \\  \hline\hline
model & \multicolumn{3}{c}{$\lambda$'s and $\alpha$'s picked to maximize $F_1$ } \\ \hline
 \multicolumn{4}{c}{DE:lasso $F_1$ score ($\sigma$)} \\ \hline
AR(3)    &  0.517 (0.042) & 0.695 (0.032) & 0.763 (0.032)     \\  
ARFIMA(3,0.15,0) &  0.470 (0.043)  &  0.664 (0.044)  & 0.725 (0.040)   \\  
ARFIMA(3,0.35,0) &  0.370 (0.037)  &  0.495 (0.042)  & 0.558 (0.052)  \\ \hline
 \multicolumn{4}{c}{DE:log-sum $F_1$ score ($\sigma$)} \\ \hline
AR(3)    &  0.621 (0.038) & 0.808 (0.028) & 0.873 (0.021)     \\  
ARFIMA(3,0.15,0) &  0.597 (0.046)  &  0.801 (0.038)  & 0.865 (0.032)   \\  
ARFIMA(3,0.35,0) &  0.551 (0.051)  &  0.732 (0.049)  & 0.802 (0.047)  \\ \hline\hline
 \multicolumn{4}{c}{DE:lasso Hamming distance ($\sigma$)} \\ \hline
AR(3)    &  646 (108)  &  396 (56)  & 318 (51)     \\  
ARFIMA(3,0.15,0) &  777 (123)  &  454 (77)  & 376 (63)   \\  
ARFIMA(3,0.35,0) &  1267 (199)  &  900 (138)  & 714 (138)  \\ \hline
 \multicolumn{4}{c}{DE:log-sum Hamming distance ($\sigma$)} \\ \hline
AR(3)    &  466 (44)  &  245 (35)  & 172 (26)      \\  
ARFIMA(3,0.15,0) &  499 (78)  &  243 (45)  & 178 (36)   \\  
ARFIMA(3,0.35,0) &  549 (87)  &  343 (63)  & 246 (50)  \\ \hline\hline
 \multicolumn{4}{c}{DE:lasso Timing (s) ($\sigma$)} \\ \hline
AR(3)    &  4.76 (0.79) & 4.62 (0.76) & 4.69 (0.82)     \\  
ARFIMA(3,0.15,0) &  5.06 (0.80) & 5.36 (0.98) & 5.14 (1.13)   \\  
ARFIMA(3,0.35,0) &  7.19 (1.69) & 8.45 (1.51) & 8.38 (1.58)   \\ \hline
 \multicolumn{4}{c}{DE:log-sum Timing (s) ($\sigma$)} \\ \hline
AR(3)    &  12.11 (1.59) & 12.83 (1.38) & 13.11 (1.60)     \\  
ARFIMA(3,0.15,0) &  13.51 (3.01) & 14.37 (3.71) & 13.60 (2.50)   \\  
ARFIMA(3,0.35,0) &  29.06 (5.28) & 25.51 (3.98) & 24.82 (4.23)  \\ \hline
 \multicolumn{4}{c}{FD:lasso $F_1$ score ($\sigma$)} \\ \hline
AR(3)    &  0.447 (0.032) & 0.639 (0.028) & 0.710 (0.023)  \\      
ARFIMA(3,0.15,0) &  0.412 (0.036)  &  0.592 (0.052)  & 0.670 (0.040)  \\      
ARFIMA(3,0.35,0) &  0.326 (0.032)  &  0.430 (0.041)  & 0.412 (0.028) \\ \hline
 \multicolumn{4}{c}{FD:log-sum $F_1$ score ($\sigma$)} \\ \hline
AR(3)    &  0.482 (0.030) & 0.716 (0.039) & 0.782 (0.021)  \\  
ARFIMA(3,0.15,0) &  0.460 (0.037)  &  0.704 (0.022)  & 0.767 (0.038)    \\  
ARFIMA(3,0.35,0) &  0.426 (0.041)  &  0.636 (0.040)  & 0.667 (0.032)  \\ \hline\hline
 \multicolumn{4}{c}{FD:lasso Hamming distance ($\sigma$)} \\ \hline
AR(3)    &  652 (106) & 417 (41) & 345 (33)    \\  
ARFIMA(3,0.15,0) &  749 (135)  &  472 (52)  & 394 (59)   \\  
ARFIMA(3,0.35,0) &  1119 (193)  &  826 (140)  & 1258 (108)  \\ \hline
 \multicolumn{4}{c}{FD:log-sum Hamming distance ($\sigma$)} \\ \hline
AR(3)    &  601 (62) & 312 (28) & 256 (29)    \\  
ARFIMA(3,0.15,0) &  637 (98)  &  346 (24)  & 264 (48)    \\  
ARFIMA(3,0.35,0) &  605 (81)  &  408 (32)  & 460 (39)  \\ \hline\hline
 \multicolumn{4}{c}{FD:lasso Timing (s) ($\sigma$)} \\ \hline
AR(3)    &  4.13 (1.09) & 5.43 (1.14) & 6.35 (1.59)      \\  
ARFIMA(3,0.15,0) &  4.75 (1.44) & 7.05 (1.85) & 7.54 (1.55)   \\  
ARFIMA(3,0.35,0) &  8.31 (1.90) & 12.08 (3.17) & 10.85 (1.65)   \\ \hline
 \multicolumn{4}{c}{FD:log-sum Timing (s) ($\sigma$)} \\ \hline
AR(3)    &  7.79 (0.92) & 9.66 (2.24) & 10.26 (2.77)        \\  
ARFIMA(3,0.15,0) &  8.75 (1.52) & 12.47 (2.16) & 12.09 (2.05)   \\  
ARFIMA(3,0.35,0) &  13.46 (2.73) & 19.07 (2.16) & 19.44 (3.10)   \\ \hline\hline
\end{tabular} 
\vspace*{-0.1in}
\end{center}
\end{table}

\subsubsection{AR models With Non-Gaussian Data} \label{SRDM5}
Here we use the AR model (\ref{VAR1}) as in Sec.\ \ref{SRDM1}, except that now $\{ {\bm w}(t) \}$ is i.i.d., zero-mean and non-Gaussian; other details remain unchanged. We have the components of ${\bm w}(t)$ as mutually independent and identically distributed, zero-mean, unit variance either exponential or Laplace random variables, whereas recall that in  Sec.\ \ref{SRDM1}, they are Gaussian. As discussed in Sec.\ \ref{STSM}, the resulting graphs are no longer CIGs, but are partial correlation graphs reflecting uncorrelation instead of independence. However, since the PSDs are identical for all three cases (Gaussian, exponential or Laplace $\{ {\bm w}(t) \}$), the corresponding graphs based on ${\bm S}_x^{-1}(f)$ are identical. The simulation results are shown in Table \ref{table6} using the proposed approaches, DE:lasso and DE:log-sum, with $d=3$. The probability density function (pdf) of Laplacian ${\bm w}(t)$ has heavier (longer) tails than Gaussian pdf of  ${\bm w}(t)$, and the pdf of exponential ${\bm w}(t)$ is skewed. The simulation results based on 100 runs are shown in Table \ref{table6}. Compared to the Gaussian AR(3) model results shown in Table \ref{table1}, the results shown in Table \ref{table6} are quite similar. For instance, for $n=2048$, for the proposed DE:log-sum approach, we have an $F_1$ score of  0.873, 0.868 and 0.867, and Hamming distance of 172, 178 and 180, for the Gaussian, Laplace and exponential series, respectively. For $n=256$, for the proposed DE:log-sum approach, we have an $F_1$ score of  0.621, 0.621 and 0.619, and Hamming distance of 466, 468 and 466, for the Gaussian, Laplace and exponential series, respectively.

\subsection{Long-Range Dependence Model} \label{LRDM}
We now consider an ARFIMA($q,\tilde{d},0$) model with ${\bm u}(t)$ in (\ref{neweq10030}) being a VAR(3) models exactly as in Sec.\ \ref{SRDM1} with $p=120$, and the fractionally integrated part is as described in Sec.\ \ref{ARF} with $d_i = \tilde{d} = 0.15$ or 0.35, $~ \forall i \in [p]$. This leads to $\gamma = 1-\tilde{d} = 0.85$ or $0.65$ in Assumption (A1), and for large $\tau$, the correlation function ${\bm R}_x (\tau)$ of $\{ {\bm x}(t) \}$ follows $|[{\bm R}(\tau)]_{ij}| = {\cal O}(|\tau|^{-0.7})$ or ${\cal O}(|\tau|^{-0.3})$ $\forall i,j \in [p]$ for $\gamma = 0.85$ and $0.65$, respectively (see Sec.\ \ref{ARF}). To generate $x_i(t)$ as in (\ref{neweq10031}), we used a moving-average approximation $x_i(t)  = \sum_{\ell =0}^{T-1} c_{i \ell} u_i(t-\ell)$ with $T=2000$ \cite{Taqqu2003}. We generate the AR part as in Sec.\ \ref{SRDM1} (with stability check etc.). As discussed in Remark 3 (Sec.\ \ref{ARF}), the underlying AR(3) sequence and the two ARFIMA sequences all have identical CIGs.

Simulation results based on 100 runs and using the proposed DE:lasso and DE:log-sum approaches, as well as the FD:lasso and FD:log-sum approaches, (all four approaches as in Sec.\ \ref{SRDM1} and Table \ref{table1}), are shown in Table \ref{table7}. We used delay $d=3$ for all sample sizes for DE:lasso and DE:log-sum, and for FD:lasso and FD:log-sum, the number of frequency test points $M$ in the interval $(0,0.5)$ were $M=2,3,4$ for sample sizes $n=256, 1024, 2048$, respectively, as in Sec.\ \ref{SRDM1} and Table \ref{table1}). The CIGs of the ARFIMA(3,0.15,0) and ARFIMA(3,0.35,0) processes are identical, and are exactly as the CIG of the underlying AR(3) model which was also used in Sec.\ \ref{SRDM1}. In Table \ref{table7} we show the results for these three processes where the results for the AR(3) series are exactly as in Table \ref{table1}. We see that the $F_1$ scores for ARFIMA(3,0.35,0) in Table \ref{table7} are lower than those for the ARFIMA(3,0.15,0) model, which, in turn, are lower than those for the AR(3) model, for the corresponding approaches, reflecting the bounds on $\| \hat{\bm{\Omega}}_{y\lambda} - \bm{\Omega}_{y}^\diamond \|_F$ given in Theorem 1. As noted in Remark 7, we see from Theorem 1 that $\| \hat{\bm{\Omega}}_{y\lambda} - \bm{\Omega}_{y}^\diamond \|_F = {\cal O}_P (\delta_{* n}/\beta_{\min}^2)$ where $\beta_{\min}$ is defined in assumption (A3), and $\delta_{* n} \propto 1/n^{0.5}$ (cf.\ Lemma 2(a)) for SRD sequences and for LRD sequences with $\gamma > 0.75$, and $\delta_{* n} \propto 1/n^{0.3}$ (cf.\ Lemma 2(c)) for LRD sequences when $\gamma = 0.65$, with all other parameters remaining the same in $R$ and $r_n$ in (\ref{neq15ab0}) and (\ref{neq15ab1}), respectively.  As discussed in Remark 5, $\beta_{\min}$ decreases with increasing $\tilde{d}$, with $\beta_{\min}$ being the largest for the AR(3) series and the smallest for  ARFIMA(3,0.35,0) series, among the three series under discussion. The performance in Table \ref{table7} improves with sample size in terms of both the $F_1$ score and Hamming distance. 

Note that the analysis of the frequency-domain approaches of \cite{Jung2015a, Tugnait2022a, Tugnait2022d, Tugnait2025c} is restricted to SRD processes although they (or any other approach) can be applied to both SRD and LRD series. It is seen from Table \ref{table7} that the proposed delay embedding approaches DE:lasso and DE:log-sum significantly outperform the corresponding frequency-domain approaches FD:lasso and FD:log-sum. For instance, for $n=2048$, for the proposed DE:log-sum approach, we have an $F_1$ score of  0.802 and Hamming distance of 246 for the ARFIMA(3,0.35,0) series, whereas FD:log-sum yields an $F_1$ score of  0.667 and Hamming distance of 460.

\begin{figure*}[htbp]
\begin{subfigure}[b]{0.5\textwidth}
\begin{center}
\includegraphics[width=0.75\linewidth]{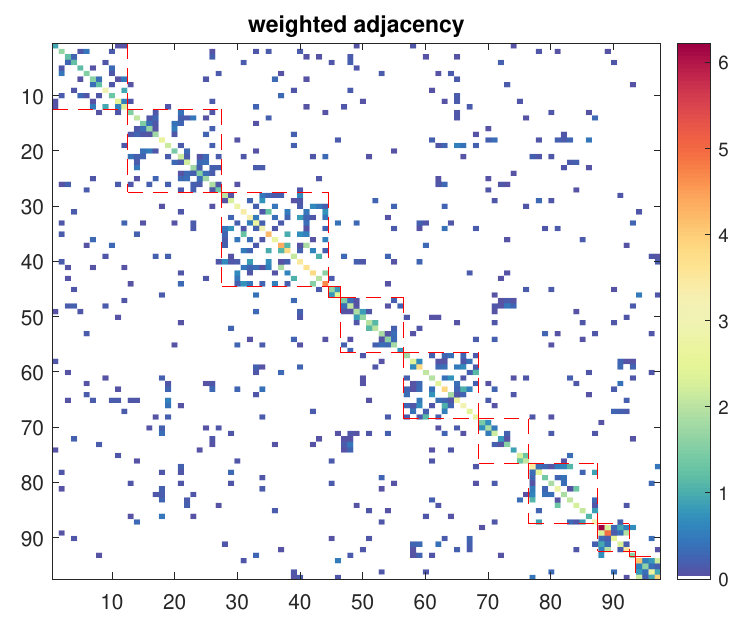} 
\caption{Heatmap, IID:log-sum, 508 edges}
\end{center}
\end{subfigure}%
\begin{subfigure}[b]{0.5\textwidth}
\begin{center}
\includegraphics[width=0.9\linewidth]{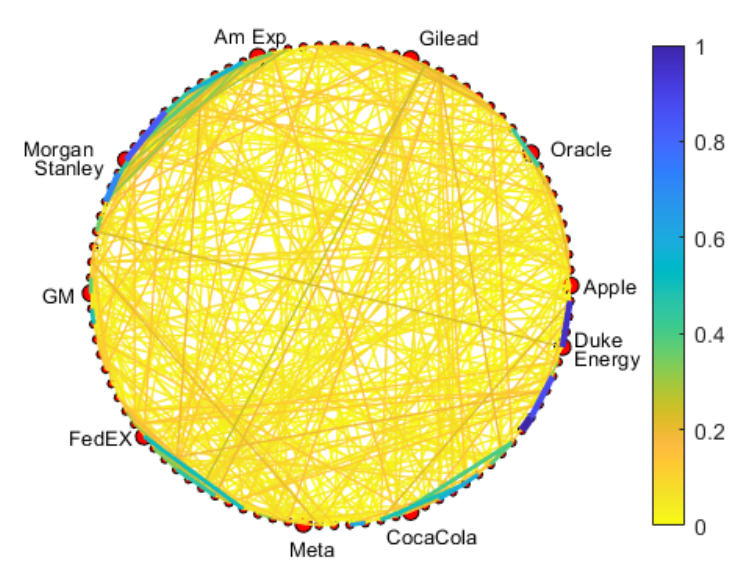}
\caption{Graph, IID:log-sum, 508 edges}
\end{center}
\end{subfigure}%
\newline
\begin{subfigure}[b]{0.5\textwidth}
\begin{center}
\includegraphics[width=0.75\linewidth]{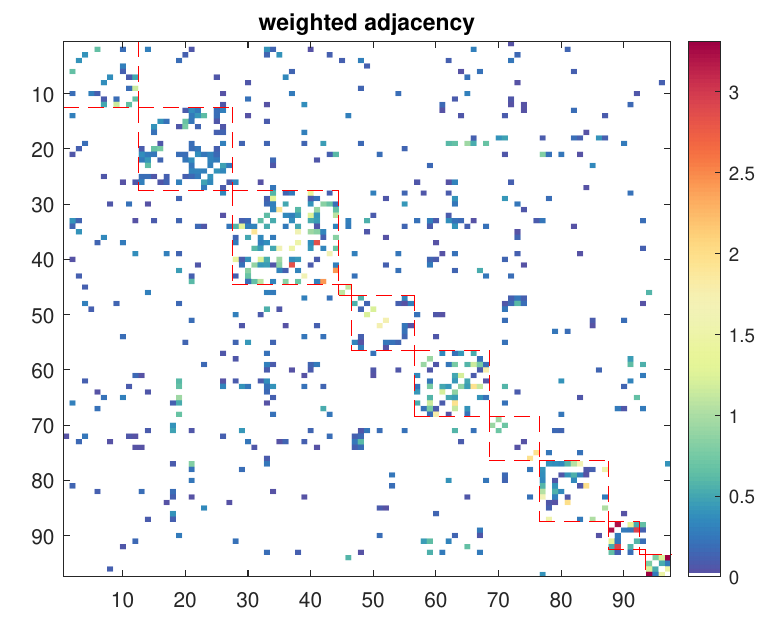}
\caption{Heatmap, FD:log-sum, 414 edges}
\end{center}
\end{subfigure}%
\begin{subfigure}[b]{0.5\textwidth}
\begin{center}
\includegraphics[width=0.9\linewidth]{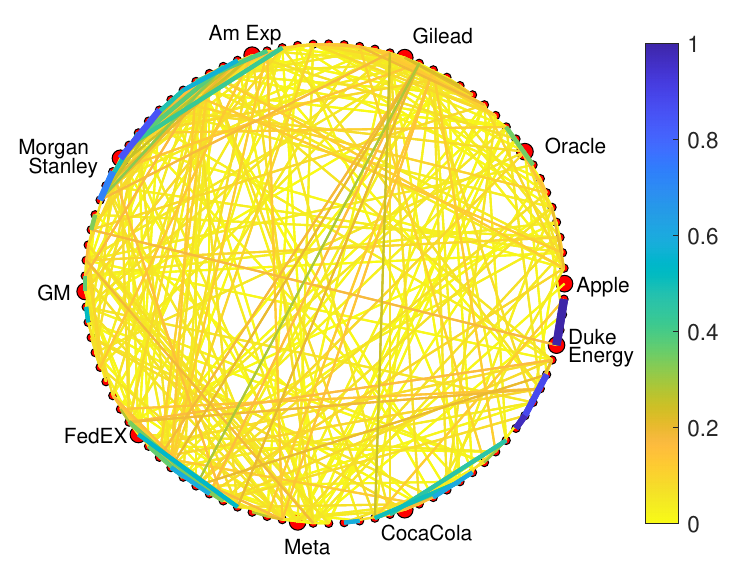}
\caption{Graph, FD:log-sum, 414 edges}
\end{center}
\end{subfigure}%
\newline
\begin{subfigure}[b]{0.5\textwidth}
\begin{center}
\includegraphics[width=0.75\linewidth]{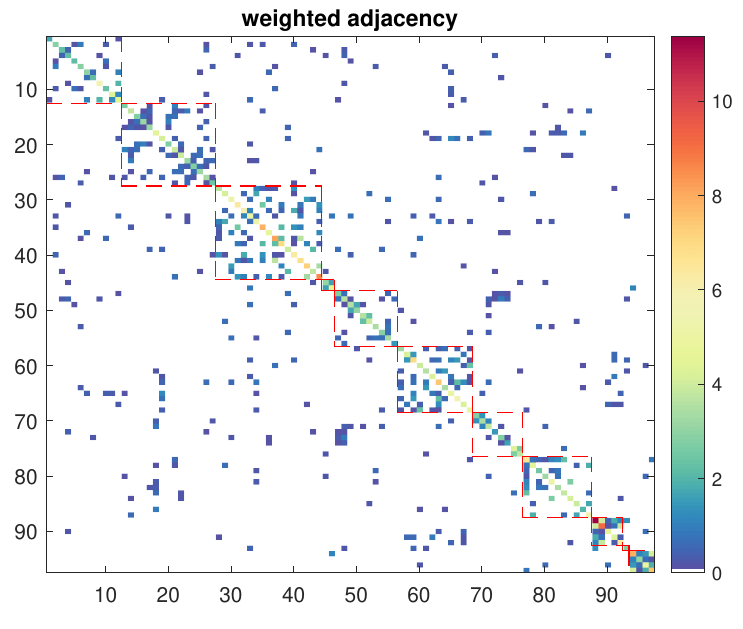}
\caption{Heatmap, DE:log-sum, 347 edges}
\end{center}
\end{subfigure}%
\begin{subfigure}[b]{0.5\textwidth}
\begin{center}
\includegraphics[width=0.9\linewidth]{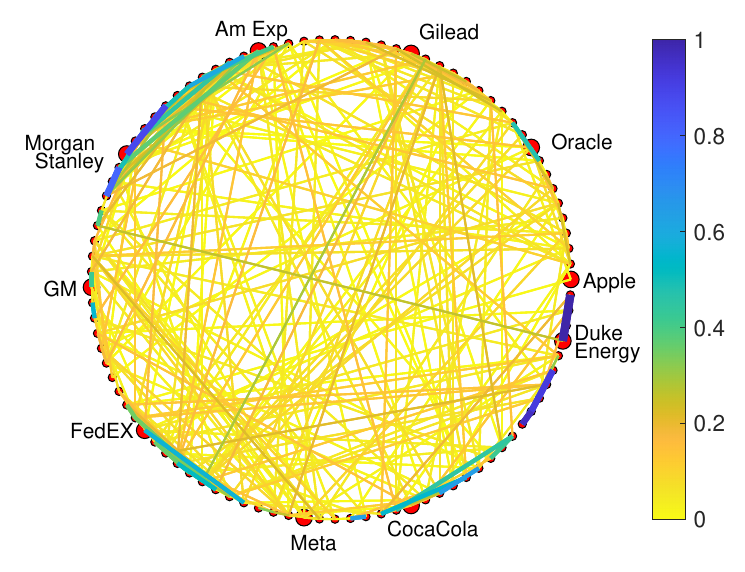}
\caption{Graph, DE:log-sum, 347 edges} \label{figrealf}
\end{center}
\end{subfigure}%
\caption{\small{Weighted adjacency heatmaps and conditional independence graphs for financial time series (S\&P 97 stocks share prices) over period Jan.\ 2, 2013 to Jan.\ 14, 2015 (512 samples): (a)-(b) time-domain IID modeling with log-sum penalty, (c)-(d) freq-domain approach with log-sum penalty, (e)-(f) proposed delay embedding approach with log-sum penalty (delay $d$=3). The heatmaps show $\| \hat{\bm \Omega}_{y \lambda}^{(k \ell)} \|_F$, $k, \ell \in [97]$ and the graphs show color-coded normalized $\| \hat{\bm \Omega}_{y \lambda}^{(k \ell)} \|_F/(\max_{k, \ell: ~k \ne \ell} \| \hat{\bm \Omega}_{y \lambda}^{(k \ell)} \|_F)$. In Figs.\ (a), (c) and (e), the red squares (in dashed lines) show the 11 GICS sectors – they are not part of the edge weights. The nodes are labeled as numbers 1 through 97 in heatmaps, and in the graphs, one counts counter-clockwise starting with the node labeled Apple, which is node 1. Nodes 10, 20, 30, 41, 50, 60, 72, 79 and 94 are labeled Oracle, Gilead, Am Exp, Morgan Stanley, GM, FedEx, Meta, CocaCola and Duke Energy, respectively.}} \label{figreal}
\end{figure*}

\begin{figure}[htbp]
\begin{subfigure}[b]{0.5\textwidth}
\begin{center}
\includegraphics[width=0.9\linewidth]{realLspDelayAlpha_graph2.pdf}
\caption{DE:log-sum with delay=3, 347 edges}
\end{center}
\end{subfigure}%
\newline
\begin{subfigure}[b]{0.5\textwidth}
\begin{center}
\includegraphics[width=0.9\linewidth]{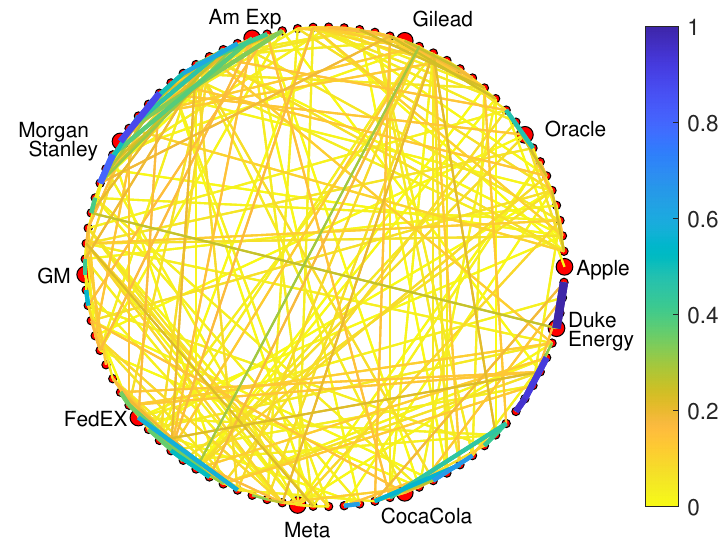}
\caption{DE:log-sum with delay=4, 307 edges}
\end{center}
\end{subfigure}%
\newline
\begin{subfigure}[b]{0.5\textwidth}
\begin{center}
\includegraphics[width=0.9\linewidth]{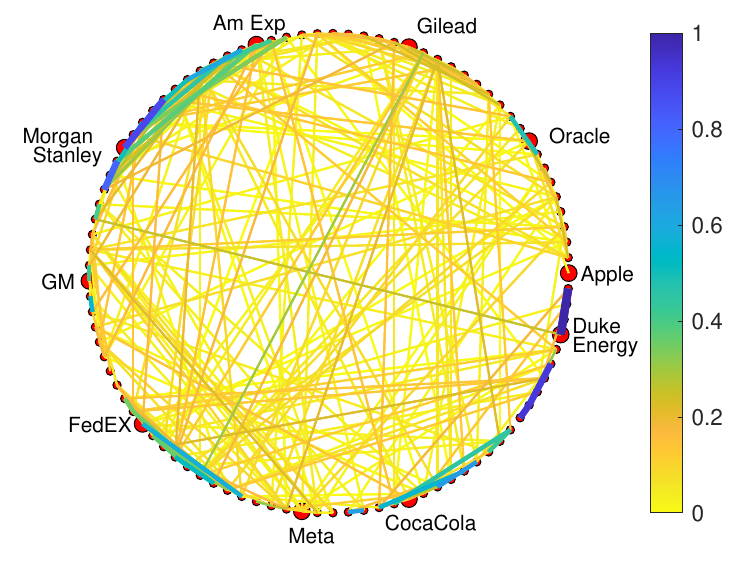}
\caption{DE:log-sum with delay=5, 310 edges}
\end{center}
\end{subfigure}%
\vspace*{-0.05in}
\caption{\small{Conditional independence graphs for financial time series (S\&P 97 stocks share prices) over period Jan.\ 2, 2013 to Jan.\ 14, 2015 (512 samples). All graphs are based on the proposed delay embedding approach with log-sum penalty: (a) delay $d$=3, (b) delay $d$=4, (c) delay $d$=5. Other details are as in Fig.\ \ref{figreal}.}} \label{figreal2}
\end{figure}

\section{Real Data: Financial Time Series} \label{NEreal}
Here our aim is to visualize and explore the conditional dependency structure underlying a financial time series. We consider daily share prices (at close of the day) of 97 stocks in the S\&P 100 index from Jan.\ 1, 2013 through Jan.\ 14, 2015. This data was gathered from the Yahoo Finance website.  Since the ground truth is not available, the main objective is to illustrate how our proposed approaches compare with some existing approaches, and how the real data results relate to the synthetic data results. 

If $z_m(t)$ is share price of $m$th stock on day $t$, we consider (as is conventional in such studies) $x_m(t) = \ln (z_m(t)/z_m(t-1))$ as the time series to analyze. Such transformations are common in the analysis of financial time series. For instance, such pre-processing is used in \cite[Sec.\ 5.2]{Songsiri2010} for topology selection for graphical models for international stock market data. We have $x_m(t) = \ln(z_m(t)) - \ln(z_m(t-1))$ which implies that we first perform log($\cdot$) transformation (generally believed to make data ``more Gaussian''), followed by lag one differencing to make the data close to univariate uncorrelated and stationary. 

These 97 stocks are classified into 11 sectors (according to the Global Industry Classification Standard (GICS)) and we order the nodes to group them as information technology (nodes 1-12), health care (nodes 13-27), financials (nodes 28-44), real estate (nodes 45-46), consumer discretionary (nodes 47-56), industrials (nodes 57-68), communication services (nodes 69-76), consumer staples (nodes 77-87), energy (nodes 88-92), materials (node 93), and utilities (nodes 94-97). For each $m$, $x_m(t)$ was centered and normalized to unit variance. The pre-processed data from Jan.\ 2, 2013 to Jan.\ 14, 2015 was taken as the time series $\{ {\bm x}(t) \}$ with 512 samples, $p=97$. The resulting time series weighted adjacency heatmaps and  the corresponding graphs, using log-sum penalty ($\epsilon = 0.0001$), are shown in Fig.\ \ref{figreal} for IID, freq.-domain (number of frequency test points $M=3$) and the proposed delay embedding (delay $d=3$) approaches, resulting in 508, 414 and 347, edges, respectively. The tuning parameter pair $(\lambda, \alpha)$ was selected via BIC (pseudo-BIC for DE as for the synthetic data examples in Table \ref{table1}); for FD:log-sum we used the BIC method detailed in \cite{Tugnait2022a, Tugnait2025c}, and for IID:log-sum we used the classical BIC method \cite{Foygel2010} to select $\lambda$. The strength of the connections in the graphs and the weights of the adjacency matrices in Fig.\ \ref{figreal} are based on the metrics used to determine presence/absence of graph edges: $\|\hat{\bm{\Omega}}_{y \lambda}^{(k \ell)} \|_F$ (see (\ref{edgesel})) for the proposed DE approach, and $|[\hat{\bm{\Omega}}_x]_{k \ell}|$ for the IID approach for heatmaps, and normalized $\| \hat{\bm \Omega}_{y \lambda}^{(k \ell)} \|_F/(\max_{k, \ell: ~k \ne \ell} \| \hat{\bm \Omega}_{y \lambda}^{(k \ell)} \|_F)$ for graph edge weights which are color-coded. In the heatmaps shown in Figs.\ \ref{figreal}(a), \ref{figreal}(c) and \ref{figreal}(e), the nodes are labeled by integers 1-97 and they have been ordered in terms of the 11 GICS sectors, as stated earlier. In the graphs shown in Figs.\ \ref{figreal}(b), \ref{figreal}(d) and \ref{figreal}(f), some of the nodes (larger red circles) are labeled by names, with node 1 labeled as Apple. Starting from Apple as node 1, one counts counterclock-wise. Thus, the node labeled Oracle is node 10, and nodes labeled Gilead, Am Exp, Morgan Stanley, GM. FedEX, Meta, CocaCola and Duke Energy are nodes 20, 30, 41, 50, 60, 72, 79 and 94, respectively.

The IID model based graph in Fig.\ \ref{figreal}(a) is relatively dense with 508 edges. The FD:log-sum and the proposed DE:log-sum approaches yield sparser graphs with 414 and 347 edges, respectively. These observations are consistent with the synthetic data results shown in Table \ref{table1} where in terms of the $F_1$ score and the Hamming distance, DE:log-sum outperforms FD:log-sum, which, in turn, outperforms the IID:log-sum approach. All three graphs appear to agree on stronger edges but differ in low-strength edges, which very likely to be false in the graph resulting from the IID:log-sum approach (cf.\ synthetic data results). We also see that in general, the stocks in a GICS sector are more strongly connected to each other compared to the stocks  in a different sector. For instance, the stocks in the financial sector are nodes 28-44 (including American Express (Am Exp) at node 30  and Morgan Stanley at node 41) and they are strongly connected to each other in all three graphs. Similarly, the stocks in the industrials sector are nodes 57-68 (including Federal Express (FedEX) at node 60) and they too are strongly connected to each other in all three graphs. 

The degree of a node in an undirected graph is defined as the number of edges incident on the node. The maximum degrees of the estimated graphs using IID:log-sum, FD:log-sum and DE:log-sum approaches are 17, 24 and 17, respectively, and the corresponding mean degrees ($\pm$ standard deviation) are 10.47 ($\pm$ 2.70),  8.54 ($\pm$ 4.14) and 7.15 ($\pm$ 2.86), respectively. Some of the highest degree nodes are:
\begin{itemize} 
\item {\it IID:log-sum graph}  \\ Johnson and Johnson (node 21, degree=17), \\ Blackrock (node 33, degree=17), 
\\ Berkshire Hathaway (node 34, degree=17), \\ Metlife (node 40, degree=16), 
\\ Emerson Electric (node 59, degree=16), \\ 3M Co.\ (node 65, degree=16).
\item {\it FD:log-sum graph} \\ Blackrock (node 33, degree=24), \\ Berkshire Hathaway (node 34, degree=21), 
\\ Honeywell (node 63, degree=19), \\ Johnson and Johnson (node 21, degree=18), 
\\ Danaher Corp.\ (node 19, degree=17), \\ Emerson Electric (node 59, degree=16).
\item {\it DE:log-sum graph} \\ Blackrock (node 33, degree=17), \\ 3M Co.\ (node 65, degree=15), 
\\ Metlife (node 40, degree=13), \\ Honeywell (node 63, degree=13), \\ Johnson and Johnson (node 21, degree=12), 
\\ Thermo Fisher Scientific (node 26, degree=12), \\ Berkshire Hathaway (node 34, degree=12),  
\\ Booking Holdings (node 48, degree=12), \\ Walt Disney (node 71, degree=12).
\end{itemize}

The strongest six edges as measured by the edge weights, are (in the order of decreasing weights):
\begin{itemize} 
\item {\it IID:log-sum graph} \\ $\{ 88, 89 \}$=\{Conoco Phillips, Chevron\}, 
\\ $\{ 94, 97 \}$=\{Duke Energy, Southern Co.\}, 
\\ $\{ 89,92 \}$=\{Chevron, Exxon Mobil\}, 
\\ $\{ 37, 41 \}$=\{Goldman Sachs, Morgan Stanley\},\\ 
 $\{ 42, 44 \}$=\{US Bancorp, Wells Fargo\}, \\ $\{ 75, 76 \}$=\{AT\&T, Verizon\}.
\item {\it FD:log-sum graph} \\ $\{ 94, 97 \}$=\{Duke Energy, Southern Co.\},  \\ $\{ 88, 89 \}$=\{Conoco Phillips, Chevron\},
\\  $\{ 89, 92 \}$=\{Chevron, Exxon Mobil\}, \\ $\{ 37, 41 \}$=\{Goldman Sachs, Morgan Stanley\}, 
\\ $\{ 42, 44 \}$=\{US Bancorp, Wells Fargo\}, \\ $\{ 81, 84 \}$=\{Altria Group, Phillip Morris\}.
\item {\it DE:log-sum graph} \\ $\{ 94, 97 \}$=\{Duke Energy, Southern Co.\}, \\ $\{ 88, 89 \}$=\{Conoco Phillips, Chevron\}, 
\\ $\{ 88, 92 \}$=\{Conoco Phillips, Exxon Mobil\},\\ $\{ 37, 41 \}$=\{Goldman Sachs, Morgan Stanley\}), 
\\ $\{ 42, 44 \}$=\{US Bancorp, Wells Fargo\}, \\ $\{ 81, 84 \}$=\{Altria Group, Phillip Morris\}.
\end{itemize}
Notice that the strongest edges detailed above are between node pairs that are in the same GICS sector.

An in-depth analysis would require domain expertise and is not within the scope of this paper. Our objective here has been to illustrate how our proposed approaches compare with some existing approaches.

In Fig.\ \ref{figreal2} we turn to the issue raised/addressed in Remark 2 (Sec.\ \ref{MAMTS}) and Table \ref{table2} (Sec.\ \ref{SRDM2}: how to pick time delay $d$? It was argued in Remark 2 to follow the heuristic of trying several values of $d$ and picking the smallest value $d=d_0$ beyond which (e.g., for $d=d_0+1$) the estimated graph is only marginally different from the estimated graph for $d=d_0$.  The synthetic data results of Sec.\ \ref{SRDM2} and Table \ref{table2} support this heuristic. In practice, since the true graph is unknown, one cannot compute the $F_1$ score or the Hamming distance to quantify marginal difference. It was stated in Remark 2 that what constitutes marginal difference would be problem and user dependent. Using the DE:log-sum approach (coupled with the BIC method for $(\lambda,\alpha)$ selection) for delay values of $d=$3, 4 and 5, we show the estimated graphs in Fig.\ \ref{figreal2}. Fig.\ \ref{figreal2}(a) is the same as Fig.\ \ref{figreal}(f). The estimated graphs have 347, 307 and 310 edges for $d=3$, 4 and 5, respectively. The graphs for $d=3$ and $d=4$ have 305 edges in common and they disagree on 44 edges, while the graphs for $d=4$ and $d=5$ have 304 edges in common and they disagree on 9 edges. A visual examination also shows very minor differences in the three graphs. This suggests that one may take Fig.\ \ref{figreal2}(a) (i.e., Fig.\ \ref{figreal}(f)) as the estimated graph. 

\section{CONCLUSION}
Estimation of the CIG of a high-dimensional stationary multivariate Gaussian time series was investigated using delay embedding and time-domain methods which allowed the series to be possibly LRD, unlike past work on this topic where frequency-domain methods whose analysis is restricted to SRD series, abound. The problem was formulated as one of multi-attribute graph estimation for random vectors where a vector is associated with each node of the graph. For an embedding dimension $d$, at each node the associated $(d+1)$-dimensional random vector consists of a time series component and its $d$ delayed copies, with integer delays up to $d \ge 1$.  A penalized negative pseudo log-likelihood function with convex lasso or non-convex regularizations was proposed to estimate the CIG. An existing ADMM approach was used to optimize the objective function. We established  sufficient conditions in a high-dimensional setting for consistency, local convexity when using non-convex penalties, and graph recovery, for both SRD and LRD series. 

We illustrated our approach with numerical examples using both synthetic and real (financial time series) data where comparisons with some competing approaches were provided. Compared to i.i.d.\ modeling based approaches, our proposed approaches yield much higher $F_1$ scores and lower Hamming distances between estimated and true edge-sets of the time series graph. Compared to the existing frequency-domain approaches, our proposed approaches are slightly to significantly better. For a $p$-dimensional time series, the computational complexity of the ADMM algorithm is ${\cal O}((d+1)^3 p^3)$. Based on the numerical examples, a heuristic was suggested and illustrated for selection of $d$ which, in theory, should tend to infinity to infer the time series graph.

In our theoretical analysis we exploited some results from \cite{Tugnait2025b} whose analysis is limited to i.i.d.\ data.  For analysis, in \cite{Tugnait2025b} two different sets of sufficient conditions have been presented: with and without some irrepresentability conditions. In this paper we did not consider any irrepresentability condition because while results of \cite[Sec.\ V-B]{Tugnait2025b} are applicable here, they are stated in terms of the covariance of the multi-attribute process $\{ {\bm y}(t)\}$ whereas in this paper we would like to formulate them in terms of the original process $\{ {\bm x}(t) \}$ also, which we are unable to do, unlike as in Remarks 4 and 5 of this paper. 

The assumptions that $\{ {\bm x}(t) \}$ is stationary and linear (i.e., generated via (\ref{neweq100})) are critical to the results of this paper. If $\{ {\bm x}(t) \}$  is not necessarily Gaussian, then the zeros of ${\bm S}_x^{-1}(f)$ encode a partial correlation graph \cite{Dahlhaus2000} which is not necessarily a CIG. For non-Gaussian sequences, our time-delay embedding approach yields an approximation to the underlying time series partial correlation graph.

\appendices
\section{Proof of Lemma 2} \label{append1}
Here we prove Lemma 2 for which we need \cite[Lemma VI.2, supplementary]{Chen16}, restated here as Lemma 4 ($\gamma$ is called $\beta$ in \cite{Chen16}) as applied to $\hat{\bm{\Sigma}}_y = \frac{1}{\bar{n}} \sum_{t=d+1}^{n} {\bm y}(t) {\bm y}^\top(t)$, $\bar{n}=n-d$, $\bm{\Sigma}_{y}^\diamond = E \{ {\bm y}(t) {\bm y}^\top(t) \}$. (The cases $\gamma=1$ and $\gamma=\frac{3}{4}$ are not in \cite[Lemma VI.2, supplementary]{Chen16}, but can be inferred from \cite[Lemma VI.1, supplementary]{Chen16}.) It is the basis of our Lemma 2, which in turn, is central to our Theorem 1. \\
{\it Lemma 4} \cite{Chen16}. Under assumption (A1), the elements of the sample covariance $\hat{\bm{\Sigma}}_y$ satisfy the tail bound
\begin{align}  
   P & \left(   \Big| [ \hat{\bm{\Sigma}}_y - \bm{\Sigma}_{y}^\diamond ]_{kl} \Big| \ge \delta \right) \nonumber \\
	   & \quad  \le 2 \, \exp \left(-C_u \min \left( \frac{\delta^2}{c_a^4(p) L_{n,\gamma}}, 
		         \frac{\delta}{c_a^2(p) J_{n,\gamma}} \right) \right)
			  \label{naeq58bx}
\end{align}
for every $\delta > 0$ and every $k,l$, where $C_u \in (0, \infty)$ is a constant that depends only on $\gamma$, and 
\begin{align*} 
(L_{n ,\gamma}^{-1}, J_{n, \gamma}^{-1}) & = \left\{
\begin{array}{ll}
(\bar{n},\bar{n}) , & \gamma > 1 \\
 (\bar{n},\bar{n}/ \ln^2(\bar{n})) , & \gamma =1 \\
   (\bar{n},\bar{n}^{2 \gamma -1}) , & 1 > \gamma > \frac{3}{4} \\
	 (\bar{n}/ \ln(\bar{n}),\bar{n}^{2 \gamma -1}), & \gamma =\frac{3}{4} \\
	 (\bar{n}^{4 \gamma -2},\bar{n}^{2 \gamma -1}) , & \frac{3}{4} > \gamma > \frac{1}{2} \, . \quad \bullet
	\end{array} \right.
\end{align*}
{\it Remark 10}. The universal constant $C_u$ results from the application of the Hanson-Wright inequality \cite{Rudelson13}. In \cite{Chen16}, the constant $c_a(p)$ (see assumption (A1)) is absorbed in $C_u$. We keep it separate to highlight its possible dependence on $p$.  $\quad \Box$

{\it Proof of Lemma 2}. For $\gamma \in (1,\infty)$ and $\bar{n} \ge N_1$, using Lemma 4 it is easy to verify that $\delta_{*n}^2/(c_a^4(p)L_{n,\gamma}) \le \delta_{*n}/(c_a^2(p)J_{n,\gamma})$ for the specified $\delta_{*n}$. Applying the union bound over all $(mp_n)^2$ entries of $\hat{\bm{\Sigma}}_y - \bm{\Sigma}_{y}^\diamond$ in Lemma 4, we have
\begin{align}
   P & \left(  \max_{k,l} \Big| [ \hat{\bm{\Sigma}}_y - \bm{\Sigma}_{y}^\diamond ]_{kl} \Big|
	    > \delta_{*n} \right)  \le P_{tb}  \nonumber \\
		& 
			 = 2 (mp_n)^2 \, \exp  \left( - C_u \, \frac{\ln(2 m^2 p_n^\tau)}{ \bar{n} C_u} \, 
			  \, \bar{n} \right)  \nonumber \\
		&  = 2 (mp_n)^2 \, \exp  \left( \ln(2 m^2 p_n^\tau )^{-1} \right)
					= \frac{1}{p_n^{\tau -2}} \, . \label{naeq58102}
\end{align}
For $\gamma \in (\frac{3}{4},1]$ and stated conditions on $\bar{n}$, we again have $\delta_{*n}^2/(c_a^4(p)L_{n,\gamma}) \le \delta_{*n}/(c_a^2(p)J_{n,\gamma})$ for the specified $\delta_{*n}$, and therefore, the desired result follows in a manner similar to (\ref{naeq58102}). For part (b), when $\gamma = \frac{3}{4}$, for the specified $\delta_{*n}$ and stated condition on $\bar{n}$, we have $\delta_{*n}^2/(c_a^4(p)L_{n,\gamma}) \le \delta_{*n}/(c_a^2(p)J_{n,\gamma})$ for the specified $\delta_{*n}$, and therefore, the desired result follows as before. For $\gamma \in (\frac{1}{2},\frac{3}{4})$, for the specified $\delta_{*n}$, it is easy to verify that we have $\delta_{*n}^2/(c_a^4(p)L_{n,\gamma}) \le \delta_{*n}/(c_a^2(p)J_{n,\gamma})$ if $N_1 \le 1$, else we have $\delta_{*n}^2/(c_a^4(p)L_{n,\gamma}) > \delta_{*n}/(c_a^2(p)J_{n,\gamma})$ if $N_1 > 1$. For $N_1 \le 1$, the desired results follows in a manner similar to (\ref{naeq58102}). If $N_1 >1$, $\gamma \in (\frac{1}{2},\frac{3}{4})$ and $\delta_{*n} = c_a^2(p) \ln(2 m^2 p_n^\tau ) / ( \bar{n}^{2 \gamma -1} C_u )$, then applying the union bound to left-side of (\ref{naeq58bx}), we have
\begin{align}
   P & \left(  \max_{k,l} \Big| [ \hat{\bm{\Sigma}}_y - \bm{\Sigma}_{y}^\diamond ]_{kl} \Big|
	    > \delta_{*n} \right)  \le P_{tb}  \nonumber \\
		& 
			 = 2 (mp_n)^2 \, \exp  \left( - C_u \, \frac{\ln(2 m^2 p_n^\tau)}{ ( \bar{n}^{2 \gamma -1} C_u )} \, 
			  \, \bar{n}^{2 \gamma -1} \right)  \nonumber \\
		&  = 2 (mp_n)^2 \, \exp  \left( \ln(2 m^2 p_n^\tau )^{-1} \right)
					= \frac{1}{p_n^{\tau -2}} \, . \label{naeq58102b}
\end{align}
This completes the proof. $\quad \blacksquare$

\section{Proofs of Theorems 1-3 and Lemma 3} \label{append2}
First some notation. Given ${\bm A} \in \mathbb{R}^{p \times p}$, ${\bm A}^+ = \mbox{diag}({\bm A})$ is a diagonal matrix with the same diagonal as ${\bm A}$, and  ${\bm A}^- = {\bm A} - {\bm A}^+$ is ${\bm A}$ with all its diagonal elements set to zero. \\
{\em Proof of Theorem 1}. 
To lighten notation, in this proof, we denote $\bm{\Omega}_y$, $\bm{\Omega}_{y}^\diamond$, $\bm{\Sigma}_{y}^\diamond$, ${\cal E}_{y}^\diamond$  and $\hat{\bm{\Omega}}_{\lambda}$ as $\bm{\Omega}$, $\bm{\Omega}^\diamond$, $\bm{\Sigma}^\diamond$, ${\cal E}^\diamond$ and $\hat{\bm{\Omega}}$, respectively. 
Let $\bm{\Omega} = \bm{\Omega}^\diamond + \bm{\Delta}$ with both $\bm{\Omega}, \, \bm{\Omega}^\diamond \succ {\bm 0}$, and 
\begin{equation}
  Q(\bm{\Omega}) := \bar{\cal L} ({\bm \Omega}) - \bar{\cal L} ({\bm \Omega}^\diamond) \, .
\end{equation}
The estimate $\hat{\bm{\Omega}}$ minimizes $Q(\bm{\Omega})$, or equivalently, $\hat{\bm{\Delta}} = \hat{\bm{\Omega}} - \bm{\Omega}^\diamond$ minimizes $G(\bm{\Delta}) := Q(\bm{\Omega}^\diamond + \bm{\Delta})$.
We will follow the method of proof of \cite[Theorem 1]{Tugnait2025b}, which, in turn, for the most part, follows the method of proof of \cite[Theorem 1]{Rothman2008} pertaining to lasso penalty. Consider the set
\begin{equation}  \label{naeq1001}
  \Theta_n(R) :=  \left\{ \bm{\Delta} \, :\, 
	 \bm{\Delta} = \bm{\Delta}^\top, \; \|\bm{\Delta} \|_F = R r_n \right\}
\end{equation}
where $R$ and $r_n$ are as in (\ref{neq15ab0}) and (\ref{neq15ab1}), respectively.  Since $G(\hat{\bm{\Delta}}) \le G(\bm{0}) = 0 $, if we can show that $\inf_{\bm{\Delta}}  \{ G(\bm{\Delta}) \, :\, \bm{\Delta} \in \Theta_n(R) \} > 0$, then a minimizer $\hat{\bm{\Delta}}$ must be inside $\Theta_n(R)$, and hence $\| \hat{\bm{\Delta}} \|_F \le   R r_n$.
It is shown in \cite[(9)]{Rothman2008} that
\begin{equation} \label{nneq100}
     \ln (|\bm{\Omega}^\diamond + \Delta|) - \ln (|\bm{\Omega}^\diamond|) = \mbox{tr} (\bm{\Sigma^\diamond  \Delta}) - A_1
\end{equation}
where, with $\bm{H}(\bm{\Omega}^\diamond, \bm{\Delta}, v ) = (\bm{\Omega}^\diamond+v \bm{\Delta})^{-1} \otimes (\bm{\Omega}^\diamond+v \bm{\Delta})^{-1}$ and $v$ denoting a scalar,
\begin{align} \label{naeq1110}
     A_1 = & \mbox{vec}(\bm{\Delta})^\top \left( \int_0^1 (1-v) 
			  \bm{H}(\bm{\Omega}^\diamond, \bm{\Delta}, v ) \, dv \right)  \mbox{vec}(\bm{\Delta}) \, .
\end{align}
Noting that $\bm{\Omega}^{-1} = \bm{\Sigma}$, we can rewrite $G({\bm{\Delta}})$ as 
\begin{equation}
     G({\bm{\Delta}}) = A_1 + A_2 + A_3 +A_4\, ,  \label{mainG}
\end{equation}
where
\begin{align} 
     A_2 = & \mbox{tr} \left( (\hat{\bm{\Sigma}} 
		 - \bm{\Sigma}^\diamond ) \bm{\Delta}  \right) \, ,  \label{naeq1100} \\
     A_3 = & \alpha  \sum_{i,j=1; i \ne j}^{mp_n} \left( \rho_\lambda (| \Omega_{ij}^\diamond +\Delta_{ij} |)
		    - \rho_\lambda (| \Omega_{ij}^\diamond  |) \right) \, , \label{naeq1120} \\
		 A_4 = & (1-\alpha) m \sum_{q,\ell=1; q \ne \ell}^{p_n} 
		  \big( \rho_\lambda ( \|({\bm \Omega}^\diamond)^{(q \ell)} + {\bm \Delta}^{(q \ell)}\|_F  )  \nonumber \\
		        & \;\;  - \rho_\lambda ( \|({\bm \Omega}^\diamond)^{(q \ell)} \|_F ) \big) \, . \label{naeq1121}
\end{align}
Following \cite[p.\ 502]{Rothman2008}, we have
\begin{equation}
     A_1  \ge \frac{ \| \bm{\Delta} \|_F^2 }{  2 (\| \bm{\Omega}^\diamond \| +  \| \bm{\Delta} \|)^2 }
		  \ge \frac{ \| \bm{\Delta} \|_F^2 }{  2 \left( \beta_{\min}^{-1} + R r_n \right)^2 } \label{boundA1}
\end{equation}
where we have used the fact that $\| \bm{\Omega}^\diamond \| = \| (\bm{\Sigma}^\diamond)^{-1} \| = \phi_{\max }((\bm{\Sigma}^\diamond)^{-1}) = (\phi_{\min }(\bm{\Sigma}^\diamond))^{-1} \le \beta_{\min}^{-1} $ and $\| \bm{\Delta} \| \le \| \bm{\Delta} \|_F = R r_n$. We now consider $A_2$ in (\ref{naeq1100}). We have
\begin{equation}
     A_2  =  \underbrace{\sum_{i,j=1; i\ne j}^{mp_n} [\hat{\bm{\Sigma}} - \bm{\Sigma}^\diamond]_{ij} \Delta_{ji}}_{L_1} + 
		\underbrace{\sum_{i=1}^{m p_n} [\hat{\bm{\Sigma}} - \bm{\Sigma}^\diamond]_{ii} \Delta_{ii} }_{L_2}
\end{equation}
To bound $L_1$, using Lemma 2, with probability $> 1- 1/p_n^{\tau-2}$,
\begin{align}
    | L_1 | & \le  \| \bm{\Delta}^-\|_1 \, \max_{i,j} \big|  [\hat{\bm{\Sigma}} - \bm{\Sigma}^\diamond]_{ij}  \big| 
		    \le   \| \bm{\Delta}^-\|_1 \, \delta_{*n}  \, . \label{naeq1205}
\end{align}
Similarly, by Cauchy-Schwarz inequality, \cite[Lemma 2]{Tugnait21a} and (\ref{neq15ab1}), 
\begin{align}
	| L_2 | & \le  \| \bm{\Delta}^+\|_1 \, \delta_{*n} \nonumber \\
	   &  \le  \delta_{*n} \, \sqrt{ m p_n} \,  \| \bm{\Delta}^+ \|_F 
				\le \| \bm{\Delta}^+ \|_F r_n  \, .
\end{align}
Therefore, with probability $> 1- 1/p_n^{\tau-2}$, 
\begin{align}
     | A_2 | & \le \| \bm{\Delta}^-\|_1 \, \delta_{*n} +
			  \| \bm{\Delta}^+ \|_F r_n  \, . \label{naeq1330} 
\end{align}
We now derive a different bound on $A_2$. Define $\tilde{\bm{\Delta}} \in \mathbb{R}^{p_n \times p_n}$ with $(i,j)$th element $\tilde{\Delta}_{ij} = \| \bm{\Delta}^{(ij)} \|_F$, where $\bm{\Delta}^{(ij)} $ is defined from ${\bm \Delta}$ similar to (\ref{neweq21}). By the Cauchy-Schwarz inequality,
\begin{align}
   \| \bm{\Delta}^-\|_1 = &  \sum_{i,j=1; i\ne j}^{mp_n} |\Delta_{ij}| \le m \| \tilde{\bm{\Delta}}^-\|_1  \nonumber \\
	  & \;  + \underbrace{\Big( \sum_{k=1}^{p_n} \|\bm{\Delta}^{(kk)}\|_1 - \| \bm{\Delta}^+\|_1 \Big)}_{=: B} \, .
\end{align}
Then using $\sum_k \|\bm{\Delta}^{(kk)}\|_1 \le m \sum_k \tilde{\Delta}_{kk} \le m \sqrt{p_n} \, \| \tilde{\bm{\Delta}}^+ \|_F$, we have
\begin{align*}
     |L_2| + \delta_{*n} \, B & 
		 \le \delta_{*n} \, (\sum_{k=1}^{p_n} \|\bm{\Delta}^{(kk)}\|_1) \nonumber \\
		& \;\;  \le \| \tilde{\bm{\Delta}}^+ \|_F \sqrt{m}\, r_n \, .
\end{align*}
Therefore, an alternative bound is
\begin{align}
     | A_2 | & 
				 \le m \| \tilde{\bm{\Delta}}^-\|_1 \, \delta_{*n} + \sqrt{m} \,
			  \| \tilde{\bm{\Delta}}^+ \|_F  r_n \, . \label{naeq1330a}
\end{align}

For the rest of the proof we have two different approaches, one for lasso and log-sum and the other for SCAD and MCP penalties. The following applies to lasso and log-sum penalties. \\
{\it For Lasso and Log-Sum Penalties}: We now bound $A_3$ in (\ref{naeq1120}).  Let $\bar{\cal E}^\diamond$ denote the true enlarged edge-set corresponding to ${\cal E}^\diamond$  (referred to as ${\cal E}_{y}^\diamond$ in assumption (A2)) when one interprets multi-attribute model as a single-attribute model with $mp_n$ nodes. Let $(\bar{\cal E}^\diamond)^c$ denote its complement. Using the mean-value theorem, we have ($\rho_\lambda^\prime(u) = \frac{d \rho_\lambda (u)}{du}$)
\begin{align}
  \rho_\lambda (| \Omega_{ij}^\diamond +\Delta_{ij} |) = & \rho_\lambda (| \Omega_{ij}^\diamond |) \nonumber \\
	& +  
	  \rho_\lambda^\prime (| \tilde{\Omega}_{ij} |)  (| \Omega_{ij}^\diamond +\Delta_{ij} | - | \Omega_{ij}^\diamond  |) \label{expan}
\end{align}
where $|\tilde{\Omega}_{ij}| = |\Omega_{ij}^\diamond| + \nu (| \Omega_{ij}^\diamond +\Delta_{ij} | - | \Omega_{ij}^\diamond  |)$ for some $\nu \in [0,1]$. We have   
\begin{align}
   A_3 & =  \alpha  \sum_{(i,j) \in \bar{\cal E}^\diamond} \rho_\lambda^\prime 
	      (| \tilde{\Omega}_{ij} |)  (| \Omega_{ij}^\diamond +\Delta_{ij} | - | \Omega_{ij}^\diamond  |) \nonumber \\
			& \;\; 	+ \alpha  \sum_{(i,j) \in (\bar{\cal E}^\diamond)^c} \rho_\lambda (| \Delta_{ij} |) \\
	  & \ge - \alpha  \sum_{(i,j) \in \bar{\cal E}^\diamond} \rho_\lambda^\prime 
	      (| \tilde{\Omega}_{ij} |)  | \Delta_{ij} |
				+ \alpha  \sum_{(i,j) \in (\bar{\cal E}^\diamond)^c} C_\lambda | \Delta_{ij} | \nonumber \\
				& \quad\quad \;\; \mbox{for } \;
				  | \Delta_{ij} | \le \delta_\lambda ,  \label{eqnn2}
\end{align}
where in the last step above, we used the triangle inequality and (\ref{prop7}) . Now use property (ii) of the penalty functions and $C_\lambda = \lambda/2$ to conclude that 
\begin{align}				
A_3 & \ge  - \alpha \lambda_n \sum_{(i,j) \in \bar{\cal E}^\diamond} | \Delta_{ij} | 
          + \alpha (\lambda_n/2) \sum_{(i,j) \in (\bar{\cal E}^\diamond)^c}  | \Delta_{ij} | \, . \label{naeq1220}
\end{align}
Next we bound $A_4$ in (\ref{naeq1121}). Considering the true edge-set ${\cal E}^\diamond$ for the multi-attribute graph, let $({\cal E}^\diamond)^c$ denote its complement. If the edge $\{i,j\} \in ({\cal E}^\diamond)^c$, then $(\bm{\Omega}^\diamond)^{(ij)} = {\bm 0}$, therefore, $\|({\bm \Omega}^\diamond)^{(ij)} + {\bm \Delta}^{(ij)}\|_F - \|({\bm \Omega}^\diamond)^{(ij)} \|_F = \|{\bm \Delta}^{(ij)}\|_F $. For $\{i,j\} \in {\cal E}^\diamond$, by the triangle inequality,
   $\|({\bm \Omega}^\diamond)^{(ij)} + {\bm \Delta}^{(ij)}\|_F - \|({\bm \Omega}^\diamond)^{(ij)} \|_F 
	  \ge  - \| {\bm \Delta}^{(ij)}\|_F$.
Mimicking the steps for bounding $A_3$, we have
\begin{align}
     A_4 & \ge  - (1-\alpha)m \lambda_n \sum_{(i,j) \in {\cal E}^\diamond} \| {\bm \Delta}^{(ij)} \|_F \nonumber \\
      & \;\;     + (1-\alpha)m (\lambda_n/2) \sum_{(i,j) \in ({\cal E}^\diamond)^c}  \| {\bm \Delta}^{(ij)} \|_F \, .\label{naeq1220b}
\end{align}

Split $A_2$ as $A_2 = \alpha A_2 + (1-\alpha) A_2$, apply bound (\ref{naeq1330}) to $\alpha A_2$ and (\ref{naeq1330a}) to $(1-\alpha) A_2$, use $\| \bm{\Delta}^- \|_1 = \| \bm{\Delta}^-_{\bar{\cal E}^\diamond} \|_1 + \|\bm{\Delta}^-_{(\bar{\cal E}^\diamond)^c} \|_1$ and $\| \tilde{\bm{\Delta}}^- \|_1 = \| \tilde{\bm{\Delta}}^-_{{\cal E}^\diamond} \|_1 + \|\tilde{\bm{\Delta}}^-_{({\cal E}^\diamond)^c} \|_1$. We have 
\begin{align}
  \alpha A_2+ & A_3 \ge  - \alpha |A_2| 
	 + \alpha \lambda_n ( 0.5 \|\bm{\Delta}^-_{(\bar{\cal E}^\diamond)^c} \|_1 - \| \bm{\Delta}^-_{\bar{\cal E}^\diamond} \|_1 ) 
	    \nonumber \\
	\ge & \alpha ( 0.5 \lambda_n - \delta_{*n}) \sum_{(i,j) \in (\bar{\cal E}^\diamond)^c}  | \Delta_{ij} |  \nonumber \\
	&  - \alpha (  \lambda_n + \delta_{*n}) \sum_{(i,j) \in \bar{\cal E}^\diamond}  | \Delta_{ij} |
	 - \alpha r_n \| \bm{\Delta}^+ \|_F \, .
\end{align}
Since we pick $\lambda_n \ge \lambda_{n\ell}$ in Theorem 1, $0.5 \lambda_n - \delta_{*n} \ge 0$ and therefore, the first term above can be neglected. Now $\sum_{(i,j) \in \bar{\cal E}^\diamond}  | \Delta_{ij} | \le \sqrt{m^2 s_{n}^\diamond} \|{\bm \Delta}\|_F$, by the Cauchy-Schwarz inequality, and $\| \bm{\Delta}^+ \|_F \le \| \bm{\Delta} \|_F$. We then have
\begin{align}	
	\alpha A_2+ & A_3 \ge  - \alpha \left( (  \lambda_n + \delta_{*n}) \sqrt{m^2 s_{n}^\diamond} 
	           + r_n \right) \| \bm{\Delta} \|_F \, .
	\label{naeq5200}
\end{align}
Similarly, we have
\begin{align}	
	(1-\alpha) A_2+ & A_4 \ge  -(1- \alpha)m \nonumber \\
	  & \;\; \times \left( (  \lambda_n + \delta_{*n}) \sqrt{ s_{n}^\diamond} 
	           +  r_n \right) \| \bm{\Delta} \|_F \, .
	\label{naeq5210}
\end{align}
From (\ref{naeq5200}) and (\ref{naeq5210}) we have
\begin{align}
  A_2+ & A_3+A_4 \ge  - \| \bm{\Delta} \|_F  \left( \lambda_n m \sqrt{s_{n}^\diamond} +\delta_{*n} m \sqrt{s_{n}^\diamond}
	      + m r_n \right) \nonumber \\
	  & \ge - \| \bm{\Delta} \|_F  \left( \lambda_n m \sqrt{s_{n}^\diamond} + (1+ m)  r_n \right)	\nonumber \\
	  &
		 \ge - 2   (1+ m)  r_n \| \bm{\Delta} \|_F \label{naeq5330}
\end{align}
where we used the fact that since $\lambda_n \le \lambda_{nu1}$, $ \lambda_n m \sqrt{s_{n}^\diamond} \le  (1+m) r_n$.
Using (\ref{mainG}), the bound (\ref{boundA1}) on $A_1$, bound (\ref{naeq5330}) on $A_2+  A_3+A_4$, and $\| \bm{\Delta} \|_F = R r_n$, we have with probability $> 1- 1/p_n^{\tau-2}$, 
\begin{align}
     G({\bm{\Delta}}) \ge &  \; \| \bm{\Delta} \|_F^2 \left[ \frac{1}{2 (\beta_{\min}^{-1} + R r_n )^2}  
		     - \frac{2(1+m)}{R} \right]  \, .\label{naeq1370} 
\end{align}
For the given choice of $N_2$, $R r_n \le R r_{N_2} \le 0.1/\beta_{\min}$ for $\bar{n} \ge N_2$. Also, $2(1+m)/R = \beta_{\min}^2/4$ by (\ref{neq15ab0}). Then for $\bar{n} \ge N_2$,
\begin{align*}
    \frac{1}{2 (\beta_{\min}^{-1} + R r_n )^2}- \frac{2(1+m)}{R}
		\ge \beta_{\min}^2 \left( \frac{1}{2.42} - \frac{1}{4} \right) > 0 \, ,
\end{align*}
implying $G({\bm{\Delta}})  > 0$. This proves (\ref{neq15}).  The choice of $N_3$ for log-sum penalty ensures that $| \Delta_{ij} | \le \delta_\lambda = \epsilon$  needed in (\ref{eqnn2}) is satisfied with high probability (w.h.p.): if $R r_n \le  \epsilon$, then  $| \Delta_{ij} | \le  \|{\bm \Delta}\|_F \le R r_n \le \epsilon$. 

The following applies to SCAD and MCP penalties. \\
{\it For SCAD and MCP  Penalties}: Here we address (\ref{expan}) differently. Using the triangle inequality, we have 
\begin{align}
  |\tilde{\Omega}_{ij}| & \ge |\Omega_{ij}^\diamond| + \nu \big( | \Omega_{ij}^\diamond  | - | \Delta_{ij} | - | \Omega_{ij}^\diamond  | \big) \nonumber \\
	  & \ge |\Omega_{ij}^\diamond| - | \Delta_{ij} | \, . \label{expan2}
\end{align}
Since $| \Delta_{ij} | \le \| \bm{\Delta} \|_F \le R r_n$, the choice $\lambda_n = \lambda_{nu2}$ implies that $\lambda_n \ge Rr_n$, satisfying $| \Delta_{ij} | \le  \lambda_n$. Therefore, $|\tilde{\Omega}_{ij}| \ge |\Omega_{ij}^\diamond| - \lambda_n$. 
For $\bar{n} \ge N_4$ and $\bar{n} \ge N_5$ for SCAD and MCP, respectively, $\rho_\lambda^\prime (|\tilde{\Omega}_{ij}|)=0$ (see (\ref{neq15ab41}) and (\ref{neq15ab42})) if $\{i,j\} \in \bar{\cal E}^\diamond$, i.e, $| \Omega_{ij}^\diamond  | \ne 0$, since in this case $|\tilde{\Omega}_{ij}| \ge (a+1) \lambda_n  - \lambda_n = a \lambda_n$ for SCAD and $|\tilde{\Omega}_{ij}| \ge (b+1) \lambda_n  - \lambda_n = b \lambda_n$ for MCP. Therefore, 
\begin{align}
   A_3 & =   \alpha  \sum_{(i,j) \in (\bar{\cal E}^\diamond)^c} \rho_\lambda (| \Delta_{ij} |) \nonumber \\
	  &  \ge 
				 \alpha  \sum_{(i,j) \in (\bar{\cal E}^\diamond)^c} C_\lambda | \Delta_{ij} |  \;\; \mbox{for } \;
				  | \Delta_{ij} | \le \delta_\lambda  \nonumber \\
		& = \alpha (\lambda_n/2) \sum_{(i,j) \in (\bar{\cal E}^\diamond)^c}  | \Delta_{ij} | \, . \label{eqnn22}
\end{align} 
Mimicking the steps for bounding $A_3$ above and under same conditions, we have
\begin{align}
     A_4 & \ge   (1-\alpha)m (\lambda_n/2) \sum_{(i,j) \in ({\cal E}^\diamond)^c}  \| {\bm \Delta}^{(ij)} \|_F \, .\label{naeq1220b2}
\end{align} 
Thus, 
\begin{align}
  \alpha A_2+ & A_3 \ge  - \alpha |A_2| 
	 + 0.5 \, \alpha \lambda_n  \|\bm{\Delta}^-_{(\bar{\cal E}^\diamond)^c} \|_1  
	    \nonumber \\
	\ge & \alpha ( 0.5 \lambda_n - \delta_{*n}) \sum_{(i,j) \in (\bar{\cal E}^\diamond)^c}  | \Delta_{ij} |  \nonumber \\
	&  - \alpha \delta_{*n} \sum_{(i,j) \in \bar{\cal E}^\diamond}  | \Delta_{ij} |
	 - \alpha r_n \| \bm{\Delta}^+ \|_F \, .
\end{align}
Since we pick $\lambda_n = \max \left( R r_n , \lambda_{n u1} \right)$ in Theorem 1, $0.5 \lambda_n - \delta_{*n} \ge 0$ and therefore, the first term above can be neglected. Now $\sum_{(i,j) \in \bar{\cal E}^\diamond}  | \Delta_{ij} | \le \sqrt{m^2 s_{n}^\diamond} \|{\bm \Delta}\|_F$, by the Cauchy-Schwarz inequality, and $\| \bm{\Delta}^+ \|_F \le \| \bm{\Delta} \|_F$. We then have
\begin{align}	
	\alpha A_2+ & A_3 \ge  - \alpha \left( \delta_{*n} \sqrt{m^2 s_{n}^\diamond} 
	           + r_n \right) \| \bm{\Delta} \|_F \, .
	\label{naeq52002}
\end{align}
By very similar arguments we also have
\begin{align}	
	(1-\alpha) A_2+ & A_4 \ge  -(1- \alpha)m \nonumber \\
	  & \;\; \times \left( \delta_{*n} \sqrt{ s_{n}^\diamond} 
	           +  r_n \right) \| \bm{\Delta} \|_F \, .
	\label{naeq52102}
\end{align}
From (\ref{naeq52002}) and (\ref{naeq52102}) we have
\begin{align}
  A_2+ & A_3+A_4 \ge  - \| \bm{\Delta} \|_F  \left( \delta_{*n} m \sqrt{s_{n}^\diamond}
	      + m r_n \right) \nonumber \\
	  &
		 \ge -    (1+ m)  r_n \| \bm{\Delta} \|_F \label{naeq52330}
\end{align}
where we used the fact that $\delta_{*n} m \sqrt{s_{n}^\diamond} \le  r_n$.
Mimicking (\ref{naeq1370}), we have with probability $> 1- 1/p_n^{\tau-2}$, we have
\begin{align}
     G({\bm{\Delta}}) \ge &  \; \| \bm{\Delta} \|_F^2 \left[ \frac{1}{2 (\beta_{\min}^{-1} + R r_n )^2}  
		     - \frac{(1+m)}{R} \right]  \nonumber \\
				\ge& \beta_{\min}^2 \left( \frac{1}{2.42} - \frac{1}{8} \right) > 0 \, , \label{naeq13702} 
\end{align}
implying $G({\bm{\Delta}})  > 0$. This proves (\ref{neq15}). For the SCAD penalty, we  need $| \Delta_{ij} | \le \delta_\lambda = \lambda_n$ in (\ref{eqnn22}). Since $| \Delta_{ij} | \le \| \bm{\Delta} \|_F \le R r_n$, the choice $\lambda_n = \lambda_{nu2}$ implies that $\lambda_n \ge Rr_n$, satisfying $| \Delta_{ij} | \le  \lambda_n$. For the MCP we need $| \Delta_{ij} | \le \delta_\lambda = b \lambda_n$ in (\ref{eqnn22}), which follows from $\lambda_n = \lambda_{nu2}$ and $| \Delta_{ij} | \le  \lambda_n < b \lambda_n$ for $b >1$. This completes the proof.   $\quad \blacksquare$

{\it Proof of Lemma 3}.   Consider $h({\bm \Omega}_y) = {\cal L}({\bm \Omega}_y)- \frac{\mu}{2} \| {\bm \Omega}_y \|_F^2$ for some $\mu \ge 0$. The Hessian of ${\cal L}({\bm \Omega}_y)$ w.r.t.\ $\mbox{vec}({\bm \Omega}_y)$ is $\nabla^2 {\cal L}({\bm \Omega}_y) = {\bm \Omega}_y^{-1} \otimes {\bm \Omega}_y^{-1}$ 
with 
\begin{align}
     \phi_{\min}(\nabla^2 {\cal L}({\bm \Omega}_y)) = &   \phi_{\min}^2( {\bm \Omega}_y^{-1})
		  =  \frac{1}{ \phi_{\max}^2( {\bm \Omega}_y)}   = \frac{1}{ \| {\bm \Omega}_y \|^2 } \, .
			 \label{naeq8372} 
\end{align}
Since $\nabla^2 h({\bm \Omega}_y) = {\bm \Omega}_y^{-1} \otimes {\bm \Omega}_y^{-1} - \mu {\bm I}_{(mp)^2}$, it follows that $h({\bm \Omega}_y)$  is positive semi-definite, hence convex, if  
\begin{align}
      \| {\bm \Omega}_y \| & \le \sqrt{\frac{1}{\mu}} \, .
			 \label{naeq8375} 
\end{align}
One of the properties of the penalty functions $\rho_\lambda(u)$ is that $g(u):=\rho_\lambda(u) +\frac{\mu}{2} u^2$ is convex for some $\mu \ge 0$, and another property is that $\rho_\lambda(u)$ is non-decreasing on $\mathbb{R}_+$. Therefore, by the composition rules \cite[Sec.\ 3.2.4]{Boyd2004}, $g(| [ {\bm{\Omega}}_y ]_{ij}|)$ and $g(\| {\bm \Omega}_y^{(q \ell )} \|_F)$ are convex. Hence,
\begin{align}
     &  P_e({\bm \Omega}_y) + \frac{\mu_e}{2} \| {\bm \Omega}_y \|_F^2  
		 =    \sum_{i \ne j}^{mp} 
		 \Big( \rho_\lambda(\big| [ {\bm{\Omega}}_y ]_{ij} \big|) +\frac{\mu_e}{2} \big| [ {\bm{\Omega}}_y ]_{ij} \big|^2 \Big)	 
		  \label{naeq8378}
\end{align}
is convex for $\mu_e = \mu \ge 0$, and similarly,
\begin{align}
  &   P_g({\bm \Omega}_y) + \frac{\mu_g}{2} \| {\bm \Omega}_y \|_F^2  
		 =  m   \; \sum_{ q \ne \ell}^p \; 
		   \Big( \rho_\lambda ( \| {\bm \Omega}_y^{(q \ell )} \|_F )  \nonumber \\
		& \quad\quad	+\frac{\mu_g}{2 m } \| {\bm \Omega}_y^{(q \ell )} \|_F^2 \Big)	
				  \label{naeq8379}
\end{align}
is convex for $\mu_g = m  \, \mu$, where $\mu$ is the value that renders $\rho_\lambda(u) +\frac{\mu}{2} u^2$ convex. Now express $\bar{\cal L}({\bm \Omega}_y)$ as
\begin{align}
   \bar{\cal L}({\bm \Omega}_y) = & \alpha \bar{\cal L}_e({\bm \Omega}_y)  
	 + (1-\alpha) \bar{\cal L}_g({\bm \Omega}_y) \, , \label{naeq8381} \\
	\bar{\cal L}_e({\bm \Omega}_y) = & 	{\cal L}({\bm \Omega}_y) - \frac{\mu}{2} \| {\bm \Omega}_y \|_F^2
	  + P_e({\bm \Omega}_y) + \frac{\mu}{2} \| {\bm \Omega}_y \|_F^2  \, , \label{naeq8381a} \\
		\bar{\cal L}_g({\bm \Omega}_y) = & {\cal L}({\bm \Omega}_y) - \frac{\mu}{2} \| {\bm \Omega}_y \|_F^2
	  + P_g({\bm \Omega}_y) + \frac{\mu}{2} \| {\bm \Omega}_y \|_F^2   \, .
				  \label{naeq8381b}
\end{align}
By (\ref{naeq8375}), (\ref{naeq8378}) and (\ref{naeq8381a}), $\bar{\cal L}_e({\bm \Omega}_y)$ is convex function of ${\bm \Omega}_y$ if $\| {\bm \Omega}_y \|  \le \sqrt{\frac{1}{\mu}}$,  and by (\ref{naeq8375}), (\ref{naeq8379}) and (\ref{naeq8381b}), $\bar{\cal L}_g({\bm \Omega}_y)$ is convex in ${\bm \Omega}_y$ if $\| {\bm \Omega}_y \|  \le \sqrt{\frac{1}{\mu_g}} = \sqrt{\frac{1}{m \mu }}$. Thus, for $\bar{\cal L}({\bm \Omega}_y)$ to be strictly convex, using the (minimum) values of $\mu$ to make $\rho_\lambda(u) +\frac{\mu}{2} u^2$ convex, we require
\begin{align}  
		   \| \bm{\Omega}_y \| < & \sqrt{\frac{1}{m \mu }}  \nonumber \\
		 = & \left\{ \begin{array}{ll}
		   \infty & : \;\; \mbox{Lasso} \\
			  \sqrt{\frac{a-1}{m }} & : \;\; \mbox{SCAD} \\
				\sqrt{ \frac{ \epsilon}{m \lambda_n} } & : \;\; \mbox{log-sum} \\
			  \sqrt{\frac{b}{m }} & : \;\; \mbox{MCP}. \end{array} \right. \label{naeq8390} 
\end{align}
The choice $\| \bm{\Omega}_y \| <  \sqrt{\frac{1}{m \mu }}$ makes ${\cal L}({\bm \Omega}_y)- \frac{\mu}{2} \| {\bm \Omega}_y \|_F^2$ positive definite, hence strictly convex. We take $\| \bm{\Omega}_y \| = 0.99 \,  \sqrt{\frac{1}{m \mu }}$, completing the proof.  $\quad \blacksquare$

{\em Proof of Theorem 2}. If $1/\beta_{\min} \le  0.99 \, \theta$, then $\bm{\Omega}_{y}^\diamond \in {\cal B}$ since $\| \bm{\Omega}_{y}^\diamond \| \le 1/\beta_{\min}$ by assumption (A3). To establish that  $\hat{\bm{\Omega}}_{y \lambda}  \in {\cal B}$, consider (${\bm \Delta}$ is as in the proof of Theorem 1)
\begin{align}  
		   \| \hat{\bm{\Omega}}_{y \lambda} \| \le & \| \hat{\bm{\Omega}}_{y \lambda} - \bm{\Omega}_{y}^\diamond\|
			  + \| \bm{\Omega}_{y}^\diamond \|  \nonumber \\
		 \le & \| {\bm \Delta} \| + 1/\beta_{\min} \le \| {\bm \Delta} \|_F + 1/\beta_{\min} \nonumber \\
		 \le & R r_n + 1/\beta_{\min} \, .
		\label{naeq8394} 
\end{align}
Therefore, $\hat{\bm{\Omega}}_{y \lambda}  \in {\cal B}$. Thus, both  $\hat{\bm{\Omega}}_{y \lambda}$ and $\bm{\Omega}_{y}^\diamond$ are feasible. The desired result then follows from Theorem 1 and (local) strict convexity of $\bar{\cal L}({\bm \Omega}_y)$ over ${\cal B}$ implied by Lemma 3. $\quad \blacksquare$

{\em Proof of Theorem 3}. We have $\| \hat{\bm \Omega}_{y \lambda}^{(q \ell)} - ({\bm \Omega}_{y}^\diamond)^{(q \ell)} \|_F \le \| \hat{\bm \Omega}_{y \lambda} - {\bm \Omega}_y^\diamond \|_F \le \bar{\sigma}_n$ w.h.p. For the edge $\{ q,\ell \} \in {\cal E}^\diamond$, we have
\begin{align}  
		   \| \hat{\bm \Omega}_{y \lambda}^{(q \ell)} \|_F = & 
			    \| ({\bm \Omega}_{y}^\diamond)^{(q \ell )} +
					 \hat{\bm \Omega}_{y \lambda}^{(q \ell )} - ({\bm \Omega}_{y}^\diamond)^{(q \ell)} \|_F  \nonumber \\
		 \ge & \| ({\bm \Omega}_{y}^\diamond)^{(q \ell )} \|_F       
		             - \| \hat{\bm \Omega}_{y \lambda}^{(q \ell )} - ({\bm \Omega}_{y}^\diamond)^{(q \ell )} \|_F \nonumber \\
		 \ge & \nu_t - \bar{\sigma}_n \ge 0.6 \, \nu_t \;\; \mbox{ for } \;\; n \ge N_6 \nonumber \\
		 > & \gamma_n \, .
		\label{naeq8397} 
\end{align}
Thus, ${\cal E}^\diamond \subseteq \hat{\cal E}$. Now consider the set complements $({\cal E}^\diamond)^c$ and $\hat{\cal E}^c$. For the edge $\{ q,\ell \} \in ({\cal E}^\diamond)^c$, $\| ({\bm \Omega}_{y}^\diamond)^{(q \ell )} \|_F = 0$. For $n \ge N_6$, w.h.p.\ we have
\begin{align}  
		   \| \hat{\bm \Omega}{y \lambda}^{(q \ell)} \|_F = & 
			    \| ({\bm \Omega}_{y}^\diamond)^{(q \ell )} +
					 \hat{\bm \Omega}_{y \lambda}^{(q \ell )} - ({\bm \Omega}_{y}^\diamond)^{(q \ell)} \|_F  \nonumber \\
		 \le & 0 + \bar{\sigma}_n \le 0.4 \, \nu_t <  \gamma_n \, ,
		\label{naeq8399} 
\end{align}
implying that $\{ q,\ell \} \in (\hat{\cal E}^\diamond)^c$. Thus, $({\cal E}^\diamond)^c \subseteq \hat{\cal E}^c$, hence $\hat{\cal E} \subseteq {\cal E}^\diamond$, establishing $\hat{\cal E} = {\cal E}^\diamond$.
 $\quad \blacksquare$

\bibliographystyle{unsrt} %

\end{document}